\documentclass{article}
\usepackage{fullpage}
\usepackage[utf8]{inputenc}
\usepackage[T1]{fontenc} 
\usepackage{csquotes}         
\usepackage[english]{babel}
\usepackage{textgreek} 

\usepackage[toc,page, title, titletoc]{appendix}
\usepackage{wrapfig}
\usepackage{authblk} 

\usepackage[backend=biber, natbib=true,style=apa, sorting=ynt, sortcites=true,uniquename=false]{biblatex}

\usepackage{amssymb}
\usepackage{pifont} 
\usepackage{amsmath,bm}
\usepackage{bbm}
\usepackage{amsthm}
\usepackage{thmtools, thm-restate}
\usepackage[hidelinks]{hyperref}
\usepackage{aligned-overset}
\usepackage[noabbrev,capitalize]{cleveref}

\newtheorem{theorem}{Theorem}[subsection]

\usepackage[table, dvipsnames]{xcolor}
\usepackage[column=0]{cellspace}
\def \vec #1{\texorpdfstring{\bm{#1}}{#1}}
\def \mat #1{\texorpdfstring{\bm{#1}}{#1}}

\usepackage{pdflscape}
\usepackage{array}     
\usepackage{booktabs}  
\usepackage{longtable} 
\usepackage{makecell} 
\usepackage{afterpage}

\usepackage{caption}
\usepackage{subcaption}

\usepackage{alphalph}

\usepackage[export]{adjustbox}
\usepackage{graphicx}
\usepackage{rotating}
\usepackage{mwe}
\graphicspath{{images/}{{\subfix{images/}}}}
\usepackage{float} 
\usepackage{multirow}
\usepackage{subfiles} 

\definecolor{asym-struc}{RGB}{0,254,0}
\definecolor{trans-struc}{RGB}{127,255,127}
\definecolor{asym-osc}{RGB}{0,149,254}
\definecolor{trans-osc}{RGB}{127,202,255}
\definecolor{asym-cyc}{RGB}{254,0,0}
\definecolor{trans-cyc}{RGB}{255,127,127}

\title{What is the relation between\\Slow Feature Analysis and the Successor Representation?}

\author[1]{Eddie Seabrook\thanks{Email: eddie.seabrook@ini.rub.de, ORCID-ID: 0000-0002-8985-1893}}
\author[1]{Laurenz Wiskott\thanks{Email: laurenz.wiskott@ini.rub.de, ORCID-ID: 0000-0001-6237-740X}}
\affil[1]{Institut für Neuroinformatik, Ruhr-Universität Bochum}
\date{\today}

\begin{document}

\maketitle

\begin{abstract}
Slow feature analysis (SFA) is an unsupervised method for extracting representations from time series data. The successor representation (SR) is a method for representing states in a Markov decision process (MDP) based on transition statistics. While SFA and SR stem from distinct areas of machine learning, they share important properties, both in terms of their mathematics and the types of information they are sensitive to. This work studies their connection along these two axes. In particular, both SFA and SR are explored analytically, and in the setting of a one-hot encoded MDP, a formal equivalence is demonstrated in terms of the grid-like representations that occur as solutions/eigenvectors. Moreover, it is shown that the columns of the matrices involved in SFA contain place-like representations, which are formally distinct from place-cell models that have already been defined using SFA.\\ 

\end{abstract}

\tableofcontents

\newpage
\section{Introduction}
\label{Intro}


One question permeating the field of machine learning research is how to learn features or representations of data that are well suited to a given model/task. Historically, this has been referred to as \textit{feature learning/selection} \citep{Guyon2003,Zhong2016}, but more recently has become known as \textit{representation learning} \citep{Bengio2013,Zhong2016}. Representation learning is one domain in which there is particularly fertile ground for the cross-pollination between machine learning and neuroscience, which can be seen from different perspectives. Firstly, representation learning methods have much to gain by taking inspiration from principles, concepts, or constraints that are known to be associated to, or at least can plausibly be associated to, brain function. Indeed, brain-inspired approaches to representation learning have historically received a lot of attention in the machine learning community \citep{Kohonen1990,Zhuang2021,Lewicki2002,Oja1982,Olshausen1996,Bell1997,Rao1999,Yamins2014,Wiskott1998b,VonDerMalsburg1973}, and neuroscience continues to be an important source of insight in the development of new methods \citep{Cox2014,Richards2019,Zaadnoordijk2022}. Secondly, something shared by all the studies just cited is that the methods used produce outputs that show some agreement with empirical observations of brain activity, which has added support to the idea that representation learning methods play a role in neural information processing. Thirdly, representation learning has become increasingly popular as a tool for analyzing, processing, and visualizing neural data (\citealp{Chapin1999,Yu2009,McKeown1998,Glaser2020}; \citealp[for reviews see][]{Cunningham2014,Humphries2021,Saxena2019}). Due to the sheer number and diversity of representation learning methods that have been introduced in the literature thus far, an important research question is how distinct methods compare in specific learning contexts, as well as how the relative performance is related to the learning principles that underlie each method. This paper tackles these questions for two representation learning methods that come from different areas of machine learning.

 
Slow feature analysis (SFA) is an unsupervised learning method for performing dimensionality reduction on time series data \citep{Wiskott1998b,Wiskott2002}. The core idea underlying this method is to find representations of a time series that have a high degree of temporal coherence. Since its invention, SFA has received significant theoretical study \citep{Blaschke2006,Creutzig2008,Sprekeler2008,Sprekeler2009,Sprekeler2011,Sprekeler2014,Wiskott2003,Richthofer2020} as well as a wide variety of technical applications \citep[for reviews, see][]{Escalante2012,Song2024}. Moreover, it has been used to model information processing in various domains of computational neuroscience \citep{Berkes2005,Franzius2007a,Franzius2008,Legenstein2010,Schoenfeld2015}.

A core concept from reinforcement learning (RL) theory is the successor representation (SR) \citep{Dayan1993}, which is a method for representing states in a Markov decision process (MDP) based on predictions about future state occupancies.  This predictive approach makes SR particularly useful for tackling sequential decision-making problems. As a result, it has achieved widespread study not only within the RL community but also in the fields of computational neuroscience and cognitive science \citep[for a review, see][]{Carvalho2024b}.

Despite coming from distinct areas of machine learning research, SFA and SR overlap in two key ways. Firstly, both are relevant to the study of spatial representations in neuroscience, such as place- and grid-cells of the hippocampal-entorhinal system \citep{Franzius2007a,Stachenfeld2014,Stachenfeld2017}. Secondly, in all of these studies, grid-like maps are generated by finding the eigenvectors of a matrix that encodes the temporal statistics of an agent moving in a spatial environment. Thus, SFA and SR agree in the type of signals they can produce as well as the mathematics with which they are formulated.

To date, some connections between SFA and SR have been pointed to, or at least implied, in the literature. For example, in \citep{Stachenfeld2017} the method used for learning eigenvectors of SR matrices is motivated conceptually by the notion of slowness. Moreover, \citet{Sprekeler2011} formulates SFA as a problem involving graph Laplacians, which have connections to SR through the framework of spectral graph theory \citep{Stachenfeld2014,Stachenfeld2017}. However, the direct relation between SFA and SR has yet to be explored formally or in significant depth. The current paper aims to fill this gap in the following way. In \cref{sec:SR}, SR is introduced and explored using various concepts from linear algebra and Markov chain theory. In \cref{sec:SFA}, the SFA algorithm is introduced, both in its classical definition as well as a number of alternative versions, each of which is formulated as a generalized or regular eigenvalue problem. In \cref{sec:Results}, the structure of each eigenvalue problem is studied in the particular case where SFA is applied to the setting of SR, i.e.\ to state trajectories generated from an MDP. The resulting family of eigenvalue problems shows various connections to quantities associated to the underlying MDP, such as the transition matrix or SR matrix generated for a given policy. Thus, by considering SFA in this specific context, a direct connection to SR is realised, which is further discussed in \cref{sec:Disc}. Moreover, by giving a broad and detailed overview of multiple versions of the SFA problem, this paper additionally acts as a general reference work on SFA for researchers that are new to the topic.

\section{Successor Representation (SR)}
\label{sec:SR}
\subsection{Reinforcement Learning (RL)}
\label{sec:RL}
Reinforcement learning (RL) is a core branch of machine learning that focuses on how agents learn behaviours by interacting with environments to maximize reward signals. The canonical paradigm of RL is a Markov decision process (MDP). This model class consists of a tuple $(\mathcal{S},\mathcal{A}, p, r)$, where  $\mathcal{S}$ and $\mathcal{A}$ are the state and action spaces that describe what situations an agent can encounter and what actions they can make, respectively. When an action $a\in\mathcal{A}$ is made in state $s\in\mathcal{S}$, the transition function $p(s'|s,a)$ determines the probability of being in state $s'\in\mathcal{S}$ at the next time point and the reward function $r(s,a)$ determines the scalar reward the agent receives. Together $p$ and $r$ define the dynamics of the MDP, and both can be deterministic or stochastic, so long as they respect the Markov property, which requires that the new state $s'$ and the reward received depend only on the current state $s$ and action $a$. Furthermore, while in the most general case $\mathcal{S}$, $\mathcal{A}$, and $t$ can be either finite or continuous, the current paper focuses on the simplest setting where all are finite.

The probability with which an agent chooses an action in a given state is given by the conditional distribution $\mu(a|s)$, which is known as the \textit{policy}, and the general goal of RL is to learn policies that maximize the reward that the agent receives over time. In order to do this, an agent first needs to be able to assess how well a given policy optimizes the reward function, which is known as the \textit{prediction problem}. Since behaviours are extended over time, it is more natural in RL to measure the cumulative reward received under a policy $\mu$ than the instantaneous reward. This cumulative notion of reward is typically referred to as the \textit{return}, and can be expressed as
\begin{equation}
\label{eq:Ret}
    g_t=\sum_{k=0}^{\infty}\gamma^k r_{t+k}
\end{equation}
where $r_t$ denotes the reward received at each time point $t$ and $\gamma\in (0,1)$ is a \textit{discount factor} that assigns larger importance to proximal over distal future rewards. Note that rewards in \cref{eq:Ret} are indexed starting at the time point $t$. This reflects the convention used in this paper whereby an action $a_t$ in state $s_t$ yields a reward $r_t$ at the same time point. An alternative convention is that the reward is received at the next time point, i.e.\ $r_{t+1}$. While both conventions commonly appear in the RL literature, the latter is somewhat more common. However, the former is used here for two reasons: 1) it is the convention used in many studies on SR \citep{Stachenfeld2014,Gershman2017,Russek2017,Stachenfeld2017,Gershman2018,Piray2021}, and 2) it simplifies various mathematical details in later sections of the paper.

In order for the return to be useful to an RL agent, two details need to be taken into account. Firstly, since MDPs often involve randomness, it is the expected value of $g_t$ that gives the most informative measure of reward for a given policy $\mu$. Secondly, the expected return naturally varies depending on which state the agent is in and what action it makes, meaning that it is natural to condition on one or more of these variables. Conditioning on states leads to the following quantity
\begin{align}
v(s_i) &=\mathbb{E}_{\mu}\Bigr[g_t |  s_t=s_i\Bigr] \\
&=\mathbb{E}_{\mu}\Biggr[\sum_{k=0}^{\infty}\gamma^k r_{t+k} \Big|  s_t=s_i\Biggr]\label{eq:VFscalar}
\end{align}
which is known as the \textit{state value function}, or simply \textit{value function}.\footnote{Sometimes the value function is denoted with a subscript or superscript $\mu$ to emphasize that it is based on an expectation w.r.t.\ the policy. However, this is dropped in the current paper since only a single policy is considered in \cref{sec:Results}.} 

A simple calculation shows that this function satisfies a self-consistency equation known as the \textit{Bellman equation} \citep{Sutton2018}. This can be expressed most compactly using a vector $\vec{v}$ containing the values of all states in $\mathcal{S}$. For this quantity, the Bellman equation is:
\begin{align}
    \vec{v} &= \vec{r}+ \gamma \mat{P}\vec{v}\label{eq:BellmanEq}
\end{align}
The matrix $\mat{P}$ is the result of combining the policy $\mu$ and the environment's transition model $p$ by marginalizing over actions, and has entries equal to:
\begin{equation}
\label{eq:TransMatPol}
P_{ij}=\sum_{a}\mu(a|s_i)p(s_j|s_i,a)
\end{equation}
Summing over the index $j$ in \cref{eq:TransMatPol} gives $\sum_jP_{ij}=\sum_{a}\mu(a|s_i)\sum_jp(s_j|s_i,a)=\sum_{a}\mu(a|s_i)=1$, meaning that the matrix $\mat{P}$ has rows that sum to $1$ and is therefore a right stochastic matrix or \textit{transition matrix} \citep{Seabrook2023}. In words, $\mat{P}$ describes the Markov chain induced across $\mathcal{S}$ when the agent behaves according to $\mu$. The vector $\vec{r}$ involves the same type of marginalization, i.e.\ $r_i=\sum_{a}\mu(a|s_i)r(s_i,a)$, and its entries describe the instantaneous reward expected in each state $s$. In this paper, it is assumed that the Markov chain described by $\mat{P}$ is ergodic, which guarantees that the Markov chain converges to a unique and strictly positive stationary distribution \citep{Seabrook2023}, henceforth denoted by $\vec{\pi}$. Ergodicity is commonly assumed in the theoretical study of RL problems, since it allows expectations to be easily evaluated and provides insight on the convergence properties of various well-known methods \citep{Sutton2018,Bojun2020,Dann2014}. Moreover, although ergodicity cannot practically be guaranteed for all kinds of RL tasks, it holds for the application domain considered in \cref{sec:Results} and is therefore a valid assumption in the current paper.

If $p$ and $r$ are fully known, \cref{eq:BellmanEq} can either be solved exactly, which is explored in the following section, or iteratively using methods from \textit{dynamic programming} \citep{Sutton2018}. However, in most practical situations, the MDP's dynamics are either (i) modelled internally by the agent - referred to as \textit{model-based learning}, or (ii) not modelled at all - referred to as \textit{model-free} learning. In either of these cases, one must resort to approximating $\vec{v}$ from sampled interactions with the environment, which come in the form of a time series, i.e.\ $(s_0,a_0, r_0)\to (s_1,a_1, r_1)\to (s_2,a_2, r_2)\to \cdots \to (s_t,a_t, r_t)$.

\subsection{Definition of SR}
\label{sec:DefSR}
\cref{eq:BellmanEq} is a self-consistency equation since $\vec{v}$ appears on both the left- and right-hand sides. Bringing both terms onto the left-hand side gives
\begin{equation}
(\mat{\mathbbm{1}}-\gamma\mat{P})^{-1}\vec{v}=\vec{r}
\end{equation}
which can then be rearranged to give
\begin{align}
   \vec{v} =&\; (\mat{\mathbbm{1}} - \gamma \mat{P})^{-1}\vec{r}\label{eq:VF-inv}\\
    :=&\;\mat{M}\vec{r}\label{eq:VF-SR}
\end{align}
The matrix-vector product in \cref{eq:VF-SR} decomposes $\vec{v}$ into two distinct quantities that are related to $\mu$: the instantaneous reward vector $\vec{r}$ from \cref{eq:BellmanEq}, and a matrix $\mat{M}$ related to the transition statistics. Markov chain theory provides a few key insights about this latter quantity \citep{Seabrook2023}, with the following being the most relevant to the current paper:
\begin{theorem}
\label{thm:SRinv+Neumann}
The matrix $\mat{M}$ is well-defined and can alternatively be expressed as
\begin{equation}
\mat{M}=\sum_{k=0}^{\infty}\gamma^k \mat{P}^k\label{eq:SRmat}
\end{equation}
which is known as a Neumann series (proof: see Appendix \ref{app:SRsum}).
\end{theorem}

\noindent In \cref{thm:SRinv+Neumann}, \cref{eq:SRmat} involves a sum over powers of the transition matrix $\mat{P}$. Each term $\mat{P}^k$ describes the \textit{$k$-step Markov chain} formed from the transitions that are realised by the policy $\mu$ at a time scale of $k$ steps \citep{Seabrook2023}, and is henceforth referred to as the \textit{$k$-step transition matrix}. The matrix $\mat{M}$ therefore extends the concept of a Markov chain such that all time scales are considered and discounted by $\gamma$.\footnote{This is similar to the concept of the fundamental matrix for absorbing Markov chains, except that in this case only transient states are included and no discounting is applied \citep{Seabrook2023}.} The entries of this matrix, $M_{ij}$, then describe the cumulative, discounted probability of occupying a state $s_j$ at some future time point given that $s_i$ is occupied initially. In analogy to \cref{eq:VFscalar}, this can be expressed as:
\begin{equation}
M_{ij}=\mathbb{E}_{\mu}\Biggr[\sum_{k=0}^{\infty}\gamma^k \mathbb{I}(s_{t+k}=s_j)\Big|  s_t=s_i\Biggr]\label{eq:SRentries}
\end{equation}
where $\mathbb{I}(s_t=s_j)=1$ if $s_t=s_j$ and $0$ otherwise. The $i$-th row of $\mat{M}$ contains the probabilities in \cref{eq:SRentries} for a given initial state $s_i$ and all possible successor states $s_j$, and therefore forms a representation of $s_i$ based on this notion of cumulative, discounted probability. For this reason, the $i$-th row of $\mat{M}$ is sometimes referred to as the \textit{successor representation} (SR) of the state $s_i$ \citep{Dayan1993}, and similarly $\mat{M}$ is referred to as the SR matrix.

In this paper, SR is considered as a ground truth quantity associated to a Markov chain for some known model $p$ and policy $\mu$. However, it is worth noting in practical settings $p$ and $\mu$ are typically not known a priori, meaning that SR needs to be approximated based on samples in a similar way to the value function $\vec{v}$, either using model-free or model-based methods \citep{Russek2017}. Unlike the value function, however, SR has no explicit reward dependence, meaning that the samples used to approximate it are simply sequences of states, i.e. $s(t_0)\to s(t_1)\to s(t_2)\to \cdots \to s(t_T)$, which are referred to henceforth as \textit{trajectories}.

\subsection{Eigenvalues and eigenvectors of SR}
\label{sec:SReig}
This section explores relationships between the eigenvalues and eigenvectors of $\mat{P}$, $\mat{P}^k$, and $\mat{M}$ that are relevant to the rest of the paper. For readers unfamiliar with eigenvalues and eigenvectors, or their respective generalized counterparts, a short introduction is provided in Appendix \ref{app:EVall}. Moreover, a number of insights from the spectral theory of Markov chains are used, for which a comprehensive overview can be found in \citep{Seabrook2023}.

\subsubsection{General case}
In the most general setting, the eigenvalues and eigenvectors of $\mat{P}$, $\mat{P}^k$, and $\mat{M}$ are related as follows:
\begin{theorem}
\label{thm:eigSR}
For a general Markov chain, if $\vec{w}$ is an eigenvector of $\mat{P}$ with eigenvalue $\lambda\in\mathbb{C}$, then it is also an eigenvector of $\mat{P}^k$ and $\mat{M}$ with eigenvalues $\lambda^k$ and $\frac{1}{1-\gamma\lambda}$, respectively, where $k>1$ (proof: see Appendix \ref{app:SREVgen}).
\end{theorem}
\noindent\cref{thm:eigSR} implies that the eigenvalues and eigenvectors of $\mat{M}$ can be obtained by first finding the same quantities for $\mat{P}$. Since the latter case is well-understood both analytically and conceptually, \cref{thm:eigSR} provides various mathematical insights for the matrix $\mat{M}$. This is particularly so for \textit{reversible Markov chains}, which are introduced in the following section.

\subsubsection{Reversible Markov chains}
\label{sec:SReigrev}
A reversible Markov chain is one for which the temporal dynamics in the stationary distribution $\vec{\pi}$ are equivalent to those of its time reversal, i.e.\ an associated Markov chain corresponding to the first one but evolving backward in time. Because of this symmetry w.r.t\ time, the transition matrices of such chains are closely related to symmetric matrices, which can be shown using the framework of spectral graph theory \citep{Seabrook2023}. The following theorem outlines four properties of such transition matrices that are particularly relevant to the current paper:
\begin{theorem}
For a reversible Markov chain with stationary distribution $\vec{\pi}$, the transition matrix $\mat{P}$ satisfies the following properties:
\begin{enumerate}
\item $\mat{\Pi}\mat{P}$ is symmetric,
\item $\mat{\Pi}^{\frac{1}{2}}\mat{P}\mat{\Pi}^{-\frac{1}{2}}$ is symmetric,
\item $\mat{P}$ is diagonalizable with real eigenvalues and eigenvectors,
\item The left and right eigenvectors of $\mat{P}$ can be chosen to be orthogonal w.r.t\ to the weighted inner products $\langle \cdot, \cdot\rangle_{\mat{\Pi}^{-1}}$ and $\langle \cdot, \cdot\rangle_{\mat{\Pi}}$, respectively,
\end{enumerate}
where $\mat{\Pi}=\textup{diag}(\vec{\pi})$ \citep[proof: see][Sections 2.6 and 4]{Seabrook2023}.
\label{thm:Prev}
\end{theorem}

\noindent The first and second points in \cref{thm:Prev} define transformed versions of $\mat{P}$ that are guaranteed to be symmetric if and only if the chain is reversible, where $\mat{\Pi}\mat{P}$ is sometimes referred to as the \textit{flow matrix} of the Markov chain \citep{Seabrook2023}. Due to these symmetries, the transition matrices of reversible Markov chains share some key properties with symmetric matrices, as outlined by the third and fourth points in \cref{thm:Prev}.

For a reversible Markov chain, \cref{thm:Prev} can be generalized to any positive power of the transition matrix, i.e.\ $\mat{P}^k$ for $k\geq 1$, as well as the SR matrix $\mat{M}$. This is described by the following theorem:
\begin{theorem}
For a reversible Markov chain with stationary distribution $\vec{\pi}$, the $k$-step transition matrix $\mat{P}^k$ and SR matrix $\mat{M}$ satisfy the following properties:
\begin{enumerate}
\item $\mat{\Pi}\mat{P}^k$ and $\mat{\Pi}\mat{M}$ are symmetric,
\item $\mat{\Pi}^{\frac{1}{2}}\mat{P}^k\mat{\Pi}^{-\frac{1}{2}}$ and $\mat{\Pi}^{\frac{1}{2}}\mat{M}\mat{\Pi}^{-\frac{1}{2}}$ are symmetric,
\item $\mat{P}^k$ and $\mat{M}$ are simultaneously diagonalizable with real eigenvalues and eigenvectors,
\item The left and right eigenvectors of $\mat{P}^k$ and $\mat{M}$ can be chosen to be orthogonal w.r.t\ to the weighted inner products $\langle \cdot, \cdot\rangle_{\mat{\Pi}^{-1}}$ and $\langle \cdot, \cdot\rangle_{\mat{\Pi}}$, respectively,
\end{enumerate}
where $\mat{\Pi}=\textup{diag}(\vec{\pi})$ and $k\geq 1$ (proof: see Appendix \ref{app:SREVrev}).
\label{thm:Pk+Mrev}
\end{theorem}

\noindent The two most important insights of \cref{thm:Pk+Mrev} to the current paper are the following: (i) since all eigenvectors are real-valued, they can easily be visualized as functions across the state space $\mathcal{S}$, and (ii) since all eigenvalues are real-valued, they can be used to order the eigenvectors \citep{Seabrook2023}. Note also that since $\mat{P}^k$ and $\mat{M}$ are simultaneously diagonalizable, it is possible to choose a common basis of eigenvectors for these matrices and only difference is the eigenvalues, which are given by \cref{thm:eigSR}. This is also insightful since it means that many concepts from the spectral theory of Markov chains carry over to SR in the time reversible setting \citep{Seabrook2023}. For example, if the eigenvectors are ordered in terms of the eigenvalues of $\mat{P}^k$ or $\mat{M}$ starting with the largest, then the first eigenvector has the smoothest profile over the state space $\mathcal{S}$ and as the eigenvalue decreases each eigenvector becomes progressively less smooth \citep{Seabrook2023}.

\cref{thm:Pk+Mrev} is relevant to all studies in the SR literature in which the underlying Markov chain is formulated as a random walk on an undirected graph, since this is equivalent to assuming time reversibility \citep{Seabrook2023}. This is a prevailing assumption in articles that feature eigenvectors of SR matrices, both in RL models \citep{Machado2018,Machado2023} and neuroscience/cognitive science applications \citep{Stachenfeld2014,Stachenfeld2017,Stoewer2022,Cothi2020}, as well as in various review papers \citep{Carvalho2024b,Momennejad2020,Gershman2018}. Note that while in some of these studies the undirected/reversible assumption is not explicitly stated, it is implicit through the visualization and analysis of eigenvectors as functions over $\mathcal{S}$, since this is only possible when they are real-valued, as stipulated by point (i) above.\footnote{To the authors' knowledge, the only study involving visualizations of SR eigenvectors without making this assumption is the model of \citet{Yu2021}, in which the eigenvectors are complex-valued and only the real parts are visualized.}\textsuperscript{,}\footnote{Note that although the theory presented in \citep{Stachenfeld2014,Stachenfeld2017} assumes undirected graphs, all eigenvectors are in practice computed using an iterative gradient descent rule in which this assumption is absent, and instead the rule is designed in such a way that only real-valued vectors can be produced. This learning rule is discussed in more depth in \cref{sec:Disc}.}

The reversible/undirected formulation of an MDP is valid in simplified settings such as a gridworld environment with a highly uniform policy (see, for example, \cref{sec:ExpSetup}). However, for a general environment and policy, Markov chains are typically non-reversible and correspond to random walks on directed graphs \citep{Seabrook2023}. In the non-reversible/directed case, $\mat{P}^k$ and $\mat{M}$ usually have complex-valued eigenvalues and eigenvectors, and the application of spectral graph theory to this setting is still an active area of research \citep{Marques2020,Sevi2023}. The simplest and most studied method for generalizing to the non-reversible/directed case are various types of symmetrization, an example of which is explored in the next section.

\subsubsection{Additive reversibilization}
A Markov chain is non-reversible if and only if it is distinct from its \textit{time reversal}, i.e.\ the Markov chain that results from the original one when the direction of time is reversed. For a Markov chain with transition matrix $\mat{P}$, the time reversal is described by the following transition matrix
\begin{equation}
\mat{P}_{\text{rev}}=\mat{\Pi}^{-1}\mat{P}^T\mat{\Pi}\label{eq:Prev}
\end{equation}
Note that for reversible chains $\mat{P}_{\text{rev}}=\mat{P}$ while for non-reversible chains $\mat{P}_{\text{rev}}\neq\mat{P}$, which is equivalent to the fact that $\mat{\Pi}\mat{P}$ is symmetric in the first case and non-symmetric in the latter case \citep{Seabrook2023}. Note that under the assumption of ergodicity, $\vec{\pi}$ is unique and strictly positive, meaning that the diagonal matrix $\mat{\Pi}^{-1}$ is well-defined and contains the reciprocals of the stationary probabilities  \citep{Seabrook2023}. This is also true in the weaker setting of a recurrent Markov chain, for which strictly positive stationary distributions exist \citep{Seabrook2023}.

While reversible Markov chains are well-characterized by a number of theoretical results, non-reversible Markov chains are generally more complex and less understood. One way to make the analysis of non-reversible Markov chains easier is to convert them to a related reversible Markov chain, which is often referred to as \textit{reversibilization} \citep{Seabrook2023}. This involves combining a non-reversible Markov chain with its time reversal in such a way that the resulting process is a reversible Markov chain. For example, given a non-reversible Markov chain, the \textit{additive reversibilization} produces a Markov chain with the following transition matrix
\begin{align}
\mat{P}_{\text{add}}:&=\frac{\mat{P}+\mat{P}_{\textup{rev}}}{2}\\
&=\frac{\mat{P}+\mat{\Pi}^{-1}\mat{P}^T\mat{\Pi}}{2}\label{eq:Padd}
\end{align}
which is an additive mixture of the transition statistics of the original Markov chain and its time reversal, hence the name \citep{Seabrook2023}.

Formally, the time reversal operation in \cref{eq:Prev} is defined only for transition matrices and not for SR matrices. However, since SR is based on transition probabilities across different time scales (see Equation \ref{eq:SRmat}), it is also possible to consider the form that $\mat{M}$ takes when time is reversed. This produces a matrix with the same form as \cref{eq:SRmat}, but involving $\mat{P}_{\textup{rev}}$ instead of $\mat{P}$, i.e.\
\begin{equation}
\mat{M}_{\text{rev}}=\sum_{k=0}^{\infty}\gamma^k (\mat{P}_{\textup{rev}})^k\label{eq:Mrev1}
\end{equation}
In the literature, this is referred to as the \textit{predecessor representation} (PR), since the entries $(\mat{M}_{\text{rev}})_{ij}$ describe the cumulative, discounted probability of occupying each state $s_j\in\mathcal{S}$ at some past time point given that $s_i$ is currently occupied \citep{Namboodiri2021,Yu2023,Jain2023}. One way to rearrange this matrix is the following
\begin{align}
\mat{M}_{\textup{rev}}\overset{(\ref{eq:Prev})}&{=}\sum_{k=0}^{\infty}\gamma^k (\mat{\Pi}^{-1}\mat{P}^T\mat{\Pi})^k\\
&=\sum_{k=0}^{\infty}\gamma^k\underbrace{(\mat{\Pi}^{-1}\mat{P}^T\mat{\Pi})(\mat{\Pi}^{-1}\mat{P}^T\mat{\Pi}) \cdots (\mat{\Pi}^{-1}\mat{P}^T\mat{\Pi})}_{k\text{ times}}\\
&=\sum_{k=0}^{\infty}\gamma^k\mat{\Pi}^{-1}\underbrace{(\mat{P}^T\mat{P}^T\cdots\mat{P}^T)}_{k\text{ times}}\mat{\Pi}\\
&=\sum_{k=0}^{\infty}\gamma^k\mat{\Pi}^{-1}(\mat{P}^T)^k\mat{\Pi}\\
&=\sum_{k=0}^{\infty}\gamma^k\mat{\Pi}^{-1}(\mat{P}^k)^T\mat{\Pi}\label{eq:Mrev2}\\
\overset{(\ref{eq:Prev})}&{=}\sum_{k=0}^{\infty}\gamma^k(\mat{P}^k)_{\textup{rev}}\label{eq:Mrev3}
\end{align}
where $(\mat{P}^k)_{\textup{rev}}$ is a transition matrix describing the time reversal of the $k$-step Markov chain. Comparing \cref{eq:Mrev1} and \cref{eq:Mrev3} offers the insight that for Markov chains the operations of reversing time and raising the transition matrix to the power of $k$ are commutative, i.e.\ $(\mat{P}^k)_{\textup{rev}}=(\mat{P}_{\textup{rev}})^k$. Another insightful way to rearrange $\mat{M}_{\textup{rev}}$ is to pull the stationary distribution matrices out of the sum in \cref{eq:Mrev2}, which gives
\begin{align}
\mat{M}_{\textup{rev}}&=\mat{\Pi}^{-1}\bigg(\sum_{k=0}^{\infty}\gamma^k(\mat{P}^k)^T\bigg)\mat{\Pi}\\
&=\mat{\Pi}^{-1}\bigg(\sum_{k=0}^{\infty}\gamma^k\mat{P}^k\bigg)^T\mat{\Pi}\\
\overset{(\ref{eq:SRmat})}&{=}\mat{\Pi}^{-1}\mat{M}^T\mat{\Pi}
\end{align}
which clearly mirrors the way $\mat{P}_{\textup{rev}}$ is defined in \cref{eq:Prev}.

It is also possible to define versions of SR in which the forward and backward temporal statistics are mixed, which in this paper are referred to generally as \textit{intercessor representations} (IR). There are various ways in which this can be defined, and one option that is particularly relevant to SFA is where SR and PR are additively mixed in equal parts, which gives
\begin{align}
\mat{M}_{\text{add}}:&=\hspace{0.03cm}\frac{\mat{M}+\mat{M}_{\text{rev}}}{2}\label{eq:Madd1}
\end{align}
in analogy to \cref{eq:Padd}. This expression can be rearranged in various ways, with the following being the most relevant to later parts of the paper
\begin{align}
\mat{M}_{\text{add}}\overset{(\ref{eq:SRmat},\ref{eq:Mrev3})}&{=}\frac{\sum_{k=0}^{\infty}\gamma^k \mat{P}^k+\sum_{k'=0}^{\infty}\gamma^{k'} (\mat{P}^{k'})_{\textup{rev}}}{2}\\
&=\sum_{k=0}^\infty\gamma^k\frac{\mat{P}^k+(\mat{P}^k)_{\textup{rev}}}{2}\\
\overset{(\ref{eq:Padd})}&{=}\sum_{k=0}^{\infty}\gamma^k  (\mat{P}^k)_{\text{add}}\label{eq:Madd2}
\end{align}
where $(\mat{P}^k)_{\text{add}}$ is a transition matrix describing the additive reversibilization of the $k$-step Markov chain. In words, \cref{eq:Madd2} says that $\mat{M}_{\text{add}}$ is the result of performing the additive reversibilization individually at each time scale $k$ and then taking the discounted sum over all time scales. Note that in the case of a reversible Markov chain the time reversal operation is redundant, which means that $\mat{M}_{\text{add}}=\mat{M}_{\text{rev}}=\mat{M}$.

It is possible to show that for ergodic Markov chains $(\mat{P}^k)_{\text{add}}$ and $\mat{M}_{\text{add}}$ satisfy all properties outlined in \cref{thm:Pk+Mrev}, except that there is no simultaneous diagonalizability in the third bullet point (see discussion below):
\begin{theorem}
\label{thm:P+Madd}
For an ergodic, not necessarily reversible Markov chain with stationary distribution $\vec{\pi}$, the matrices $(\mat{P}^k)_{\textup{add}}$ and $\mat{M}_{\textup{add}}$ satisfy the following properties:
\begin{enumerate}
\item $\mat{\Pi}(\mat{P}^k)_{\textup{add}}$ and $\mat{\Pi}\mat{M}_{\textup{add}}$ are symmetric,
\item $\mat{\Pi}^{\frac{1}{2}}(\mat{P}^k)_{\textup{add}}\mat{\Pi}^{-\frac{1}{2}}$ and $\mat{\Pi}^{\frac{1}{2}}\mat{M}_{\textup{add}}\mat{\Pi}^{-\frac{1}{2}}$ are symmetric,
\item $(\mat{P}^k)_{\textup{add}}$ and $\mat{M}_{\textup{add}}$ are diagonalizable with real eigenvalues and eigenvectors,
\item The left and right eigenvectors of $(\mat{P}^k)_{\textup{add}}$ and $\mat{M}_{\textup{add}}$ can be chosen to be orthogonal w.r.t\ to the weighted inner products $\langle \cdot, \cdot\rangle_{\mat{\Pi}^{-1}}$ and $\langle \cdot, \cdot\rangle_{\mat{\Pi}}$, respectively,
\end{enumerate}
where $\mat{\Pi}=\textup{diag}(\vec{\pi})$ and $k\geq 1$ (proof: see Appendix \ref{app:SREVadd}).
\end{theorem}
\noindent The main way in which \cref{thm:P+Madd} differs from \cref{thm:Pk+Mrev} is that the latter involves a continuity across time scales that is not present in the former. This can be seen from two perspectives. Firstly, in \cref{thm:Pk+Mrev} it is guaranteed that each power of the transition matrix $\mat{P}^k$ and the SR matrix $\mat{M}$ correspond to the same underlying Markov chain, while this is not the case in \cref{thm:P+Madd}. To see this, note that $(\mat{P}^k)_{\textup{add}}$ does not necessarily describe the $k$-step transition statistics of the chain associated to $\mat{P}_{\textup{add}}$, since additive reversibilization and raising the transition matrix to the power of $k$ are generally non-commutative, i.e.\ $(\mat{P}^k)_{\textup{add}}\neq (\mat{P}_{\textup{add}})^k$. Consequently, $\mat{M}_{\textup{add}}$ is generally not a sum over powers of $\mat{P}_{\textup{add}}$, and it is unclear whether there exists a reversible Markov chain for which $\mat{M}$, as defined by \cref{eq:SRmat}, is the same as $\mat{M}_{\textup{add}}$ of the non-reversible Markov chain. Secondly, while $\mat{M}$ and $\mat{P}^k$ are simultaneously diagonalizable in \cref{thm:Pk+Mrev}, this is generally not the case for $(\mat{P}^k)_{\textup{add}}$ and $\mat{M}_{\textup{add}}$ in \cref{thm:P+Madd}, which means that a common eigenbasis typically does not exist. One special case in which \cref{thm:P+Madd} and \cref{thm:Pk+Mrev} perfectly align is for a reversible Markov chain, since in this case the additive reversibilization is redundant and all matrices in the two theorems are the same. 

It is worth emphasizing that IR has received little attention in the literature, with the work of \citet{Keck2024} being, to the authors' knowledge, the only prior study exploring this concept. In the current section, focus is placed on $\mat{M}_{\textup{add}}$ since it is the version of IR that is most relevant to SFA. However, there are other ways to define SRs that combine the forward and backward temporal statistics of a Markov chain, some of which might be more suitable in other contexts of machine learning or computational neuroscience. Three such examples are discussed below.

Firstly, one alternative, considered by \citet{Keck2024}, is to compute the additive reversibilization only on $\mat{P}$, and then plug the resulting matrix $\mat{P}_{\text{add}}$ into the regular expression of SR. This can be denoted $\mat{M}(\mat{P}_{\text{add}})$ and is given by
\begin{align}
\mat{M}(\mat{P}_{\text{add}})\overset{(\ref{eq:SRmat})}&{=}\sum_{k=0}^\infty \gamma^k (\mat{P}_{\text{add}})^k\label{eq:M(Padd)}
\end{align}
Since this is a regular SR matrix based on the reversible chain described by $\mat{P}_{\textup{add}}$, it satisfies all properties outlined in \cref{thm:Pk+Mrev}. Thus, in contrast to $\mat{M}_{\textup{add}}$, the quantity $\mat{M}(\mat{P}_{\text{add}})$ is a sum over all $k$-step transition statistics of the unique chain described by $\mat{P}_{\text{add}}$ and a common eigenbasis is shared across all time scales. Note that one case in which $\mat{M}_{\textup{add}}$ and $\mat{M}(\mat{P}_{\text{add}})$ coincide is the case of a reversible Markov chain.

Secondly, additive mixtures need not equally balance the forward and backward statistics, and can instead assign more weight to one temporal direction. In the model of \citet{Keck2024}, this is achieved by generalizing $\mat{P}_{\textup{add}}$ to a convex combination of $\mat{P}$ and $\mat{P}_{\textup{rev}}$, i.e.
\begin{equation}
\mat{P}_{\textup{add},\alpha}=\alpha\mat{P}+(1-\alpha)\mat{P}_{\textup{rev}}\qquad\text{w/}\quad\alpha\in[0,1]\label{eq:Padd_alpha}
\end{equation}
and inserting the resulting matrix into the definition of SR, which for $\alpha=0,\frac{1}{2}$, and $1$ yields $\mat{M}_{\textup{rev}}$, $\mat{M}(\mat{P}_{\textup{add}})$, and $\mat{M}$, respectively. Note that an analogous convex combination of $\mat{M}$ and $\mat{M}_{\textup{rev}}$ is given by
\begin{equation}
\mat{M}_{\textup{add},\alpha}=\alpha\mat{M}+(1-\alpha)\mat{M}_{\textup{rev}}\qquad\text{w/}\quad\alpha\in[0,1]\label{eq:Madd_alpha}
\end{equation}
which for $\alpha=0,\frac{1}{2}$, and $1$ yields $\mat{M}_{\textup{rev}}$, $\mat{M}_{\textup{add}}$, and $\mat{M}$, respectively. One downside to using unequal additive mixtures is that the associated transition and SR matrices generally have complex eigenvalues and eigenvectors, since the non-reversible components of the probability flow are not completely removed.

Thirdly, all versions discussed so far are based on additive reversibilization. However, many reversibilization methods have been studied in the Markov chain literature \citep[for a summary, see][]{Choi2024}, each of which can potentially be used to define a version of IR. The additive reversibilization is one of the oldest and most well-known methods, appearing in the classical text by \citet{Fill1991} together with an alternative known as the \textit{multiplicative reversibilization}.

\cref{thm:P+Madd} is particularly relevant to the results of \cref{sec:Results}, which feature all matrices described in this theorem. A key insight of \cref{thm:P+Madd} for the results is that, like in the reversible case, both the eigenvalues and eigenvectors are real-valued, meaning that the latter can be easily visualized as functions across $\mathcal{S}$ and ordered in terms of the former.

\subsection{Time indexing in SR}
\label{sec:SRtime}
There are two details worth pointing out about the index $k$ in the definition of SR. Firstly, the starting value is $k=0$. In \cref{eq:SRentries}, this corresponds to the probability of being in state $s_j$ at time $t$ given that one is in state $s_i$ at time $t$, i.e.\ a probability of $1$ for $i=j$ and $0$ otherwise. Equivalently, this contributes an identity matrix $\mat{P}^0=\mat{\mathbbm{1}}$ to the sum in \cref{eq:SRmat}. This first term in SR arises due to the convention of assigning rewards at time $t$ (see \cref{sec:RL}). If the alternative convention is used whereby rewards are assigned at time $t+1$, then it is possible to show that the resulting SR is equivalent to that in \cref{eq:SRmat,eq:SRentries} except that $k$ starts at $1$ \citep[see, for example,][]{Carvalho2024b}. Note that none of the key properties established for $\mat{M}$ or $\mat{M}_{\text{add}}$ in the previous section are impacted by this detail. Secondly, in \cref{eq:SRmat,eq:SRentries} the index $k$ increases without bound, meaning that SR describes transitions between states across arbitrarily large time scales. This stems from the fact that the value function defined in \cref{eq:VFscalar} similarly involves rewards that extend arbitrarily far into the future. However, both the value function and SR can alternatively be defined with a finite temporal horizon $k_{\text{max}}$, which in some contexts is computationally a more convenient formalism to use \citep[see, for example,][]{Carvalho2024b}. Like for the previous point, none of the core results of the previous section are impacted by this choice of convention. For a finite horizon, the main difference is that in order to evaluate the eigenvalues in \cref{thm:eigSR} a finite geometric series is used. One practical difference this can lead to is that eigenvalues of $\mat{M}$ are not guaranteed to be ordered in exactly the same way as those of $\mat{P}$, however this is a finite size effect that becomes negligible for larger values of $k_{\text{max}}$. Note that the only point at which a finite horizon is used in the current paper is in the limit results of \cref{sec:LimRes}, since this is the most natural choice in the context of SFA. Moreover, in the experiments of \cref{sec:Exp}, which illustrate these results, a sufficiently large horizon is used such that the finite size effect just described is not observed.

\section{Slow Feature Analysis (SFA)}
\label{sec:SFA}
\subsection{Definition of SFA}
\label{sec:SFAdef}

Slow feature analysis (SFA) is an unsupervised learning method for performing dimensionality reduction on time series data. Given a multivariate time series $\vec{x}(t)=[x_1(t), x_2(t), ..., x_N(t)]\in\mathbb{R}^N$, it aims to find a set of scalar input-output functions such that the outputs generated vary slowly in time. Moreover, SFA stipulates that these outputs should capture meaningful features of variation without redundancy, and this is usually achieved by requiring that they have zero mean, unit variance, and are uncorrelated. The standard definition of SFA, referred to in this paper as Type 1 SFA, is described formally by the following:\\

\noindent \textbf{Type 1 SFA problem}: \textit{Given a function space $\mathcal{F}$ and an $N$-dimensional input signal $\vec{x}(t)\in\mathbb{R}^N$, find a set of $J$ real-valued input-output functions $g_j(\vec{x})$ generating the output signals $y_j(t):=g_j(\vec{x}(t))$ such that for $j=1, 2, ..., J$}

\begin{equation}
\label{eq:Slowness}
\Delta(y_j)=\langle \dot{y}_j(t)^2 \rangle_t\quad\text{is minimal}
\end{equation}

\noindent \textit{under the constraints}
\begin{align}
\langle y_j(t) \rangle_t&=0& &\text{(zero mean)}\label{eq:SFAzeromean}\\
\langle y_j(t)^2 \rangle_t&=1& &\text{(unit variance)} \label{eq:SFAunitvar}\\
\forall i<j \quad \langle y_i(t)y_j(t)\rangle_t&=0& &\text{(decorrelation and order)}\label{eq:SFAdecorr}
\end{align}
\noindent \textit{with $\langle \cdot \rangle_t$ and $\dot{y}_j(t)$ representing temporal averaging and the time derivative of $y_j(t)$, respectively.}\\

If the function space $\mathcal{F}$ is linear, i.e.\ $g_j(\vec{x})=\vec{w}_j^T\vec{x}$, then the resulting problem is typically referred to as \textit{linear SFA}. However, a key strength of SFA is that the function space $\mathcal{F}$ can be chosen flexibly, meaning that unlike other classical dimensionality reduction methods it is not constrained to linear transformations. Non-linear function spaces are typically achieved in SFA by first applying a $k$-th degree polynomial expansion to the input time series, i.e.\ $\vec{x}(t)\to\vec{h}(t)=\mathcal{P}_k(\vec{x}(t))$, and then  considering linear input-output functions on the expanded input, i.e.\ $g_j(\vec{h})=\vec{w}_j^T\vec{h}$ \citep{Wiskott2002}. Under this procedure, linear SFA corresponds either to an expansion of degree $k=1$ or to no expansion at all. Although the expansion step is an important component of SFA that is widely used in applications, it is not relevant to the results in \cref{sec:Results}. Therefore, unless otherwise stated, this paper focuses on the setting of no expansion, meaning that linear input-output functions are applied to the input time series.

The quantity in \cref{eq:Slowness} is sometimes referred to as the $\Delta$-value of the output signal $y_j(t)$, and it defines a measure of slowness of this signal. Note that due to the zero mean constraint in \cref{eq:SFAzeromean}, this measure is equivalent to the variance of the associated time derivative signal $\dot{y}_j(t)$. In practical applications, time series data is discretized over time and is of finite length, i.e.\ $t=1,2,..., T$. Therefore, in order to be able to compute the $\Delta$-value of an output signal, it is necessary to define a discrete analogue of the time derivative $\dot{y}_j(t)$. The most intuitive, and also the most commonly utilized, method for doing this is to use the differences between neighbouring values, i.e.\
\begin{equation}
\dot{y}_j(t)\approx y_j(t+1)-y_j(t)
\label{eq:TimeDeriv}
\end{equation}
which for a finite data set with $T$ time points can only be computed for $t=1,2,...,T-1$. 

\subsection{SFA based on correlation}
The objective in SFA can alternatively be expressed in terms of correlations, as explored in \citep{Blaschke2006}. This can be shown by inserting \cref{eq:TimeDeriv} into \cref{eq:Slowness}, which yields:
\begin{align}
\Delta(y_j)&=\hspace{0.25cm}\langle (y_j(t+1)-y_j(t))^2 \rangle_t\label{eq:SlownessAC1}\\
&=\langle y_j(t+1)^2\rangle_t+\langle y_j(t)^2 \rangle_t-2\underbrace{\langle y_j(t)y_j(t+1)\rangle_t}_{=\text{C}_1(y_j)}\label{eq:SlownessAC2}\\
\overset{(\ref{eq:SFAunitvar})}&{=}1+1-2\text{C}_1(y_j)\label{eq:SlownessAC3}\\
&=2\big(1-\text{C}_1(y_j))\label{eq:SlownessAC4}
\end{align}
Two details should be emphasized out about the equations above. Firstly, in \cref{eq:SlownessAC3} the first and second terms are evaluated to $1$ using the unit variance constraint of SFA. Formally, since $t$ runs from $t=0$ to $t=T-1$ for the time differences, first and second terms miss the first and last data point, respectively, meaning that they are not numerically equal to $1$. However, since SFA is typically applied to very long time series, this edge effect is in practice negligible and is therefore ignored henceforth in the current paper. Secondly, the quantity $\text{C}_1(y_j)$ is a measure of the average similarity between $y_j(t)$ and $y_j(t+1)$ over time. Provided that the zero mean and unit variance constraints have already been satisfied, it is the correlation between $y_j(t)$ and $y_j(t+1)$, and is referred to as the \textit{time-lagged correlation} of $y_j(t)$ with time-lag $1$ in this paper. \cref{eq:SlownessAC4} expresses the intuitive fact that a given signal being slow is equivalent to it having similar values over time, and therefore having a large value of $\text{C}_1(y_j)$ \citep{Blaschke2006}. Hence, minimizing $\Delta(y_j)$ is equivalent to maximizing $\text{C}_1(y_j)$.

Throughout the rest of the paper, SFA problems that use different objectives to \cref{eq:Slowness} are referred to as \textit{variants} of SFA, and the variant of SFA in which $\text{C}_1(y_j)$ is maximized is referred to as CSFA.

\subsection{SFA without the zero mean constraint}
\label{sec:T2SFA}

It is possible to define SFA both with or without the zero mean constraint, meaning that there are two \textit{types} of the SFA problem. While the previous sections focused on Type 1 SFA, in which the zero mean constraint is included, the current section explores Type 2 SFA, in which it is omitted. While Type 2 SFA has received less study in the literature \citep[but see, for example, the variational calculus formalism used in][]{Franzius2007a,Sprekeler2008,Sprekeler2009} it is introduced here because of its relevance to the results of \cref{sec:Results}. For reasons explained below, Type 2 SFA is most naturally formulated for a total of $J+1$ outputs, with the index $j$ starting at $0$. This leads to the following optimization problem for Type 2 SFA:\\

\noindent \textbf{Type 2 SFA problem}: \textit{Given a function space $\mathcal{F}$ and an $N$-dimensional input signal $\vec{x}(t)\in\mathbb{R}^N$, find a set of $J+1$ real-valued input-output functions $g_j(\vec{x})$ generating the output signals $y_j(t):=g_j(\vec{x}(t))$ such that for $j=0, 1, ..., J$}

\begin{equation}
\Delta(y_j)=\langle \dot{y}_j(t)^2 \rangle_t\quad\text{is minimal}
\end{equation}

\noindent \textit{under the constraints}
\begin{align}
\langle y_j(t)^2 \rangle_t&=1& &\text{(unit 2nd moment)} \label{eq:SFAunit2ndmoment}\\
\forall i<j \quad \langle y_i(t)y_j(t)\rangle_t&=0& &\text{(zero 2nd cross-moment and order)}\label{eq:SFAzerocrossmoment}
\end{align}
\noindent \textit{with $\langle \cdot \rangle_t$ and $\dot{y}_j(t)$ representing temporal averaging and the time derivative of $y_j(t)$, respectively.}\\

Many of the details outlined so far for Type 1 SFA apply also for Type 2 SFA. For example, non-linear function spaces are naturally achieved using polynomial expansions, and consideration in this paper is restricted to the simpler linear case. Moreover, the $\Delta$-value of Type 2 SFA can be reformulated following the same steps as in \Crefrange{eq:SlownessAC1}{eq:SlownessAC4}. This leads to a quantity $\text{C}_1(y_j)$ that is maximized, and for consistency the associated algorithm is referred to as Type 2 CSFA in this paper.

One difference between Type 1 and Type 2 SFA is that they involve different statistical quantities. This is most evident in the constraints, where in Type 2 SFA the absence of the zero mean constraint means that $\langle y(j)^2\rangle_t$ is the \textit{2nd moment} (rather than variance) of $y_j(t)$ and $\langle y_i(t)y_j(t)\rangle_t$ is the \textit{2nd cross-moment} (rather than correlation) of $y_i(t)$ and $y_j(t)$. Similarly, in the objectives of Type 2 SFA and CSFA, $\Delta (y_j)$ is the 2nd moment of $\dot{y}_j(t)$ and $\text{C}_1(y_j)$ is the \textit{time-lagged 2nd moment} of $y_j(t)$, respectively. Note that each of these statistical quantities reduces to its corresponding Type 1 counterpart if $y_j(t)$ has zero mean, and in that sense they are the most natural generalization of the Type 1 quantities. Moreover, they play similar roles to their Type 1 counterparts, with $\Delta (y_j)$ measuring the slowness of $y_j$, $\langle y(j)^2\rangle_t$ fixing the scale of variation, $\langle y_i(t)y_j(t)\rangle_t$ reducing redundancy between distinct features, and $\text{C}_1(y_j)$ measuring the continuity of each feature over time.

Another difference is that the solutions to Type 1 and Type 2 SFA are generally not the same. However, assuming that the function space $\mathcal{F}$ with input $\vec{x}(t)$ can produce a constant output, then the Type 1 and Type 2 solutions are closely related. To see this, consider the following constant signal:
\begin{equation}
\label{eq:SFAconstsignal}
y_0(t)=\pm [1,1,...,1]
\end{equation}
The first detail to note about $y_0(t)$ is that it has a mean value of $1$ and no variation, meaning that it is prohibited in Type 1 SFA, however it is allowed in Type 2 SFA since it has unit second moment. Moreover, since this signal does not change over time, it has a minimal $\Delta$-value of $0$ and a maximal correlation of $\text{C}_1=1$, and is therefore the slowest or most correlated signal possible. Therefore, under the assumption that $y_0(t)$ can be represented by $\mathcal{F}$ given $\vec{x}(t)$, it is guaranteed to be the first output for Type 2 SFA. Furthermore, following the reasoning of \citet{Sprekeler2008}, if $y_1(t)$ is the second Type 2 solution then applying the zero 2nd cross-moment constraint to $y_0(t)$ and $y_1(t)$ gives
\begin{equation}
\langle y_0(t)y_1(t)\rangle_t=\pm \langle y_1(t) \rangle_t=0
\end{equation}
meaning that this constraint effectively forces $y_1(t)$ to have zero mean, and by extension unit variance. This means that $y_1(t)$ is the slowest signal satisfying all Type 1 constraints and therefore corresponds to $j=1$ for Type 1 SFA. By a similar argument, the same correspondence holds for $j>1$. Therefore, under the assumption that $\mathcal{F}$ and $\vec{x}(t)$ can represent a constant signal, the solutions to Type 1 and Type 2 SFA are equivalent except for $y_0(t)$, meaning that there are $J$ solutions to Type 1 SFA and $J+1$ to Type 2 SFA. Moreover, while $y_0(t)$ can be useful in some specific domains, for example when approximating value functions in RL, in many common applications of SFA it is not an informative output and can therefore be discarded \citep[like in the variational calculus formalism used in][]{Franzius2007a,Sprekeler2014,Sprekeler2008,Sprekeler2009}, in which case there are $J$ solutions for both Type 1 and Type 2 SFA.

Note that there is no general guarantee that $y_0(t)$ can be represented by $\mathcal{F}$ given $\vec{x}(t)$, meaning that the relation established above between Type 1 and Type 2 SFA is not universal. However, there are two cases in which this property can be assumed. On the one hand, for some time series a constant output $y_0(t)$ is always possible, even for a linear function space $\mathcal{F}$, with one example being the time series considered in the results of this paper (see \cref{sec:OneHot}). On the other hand, for some function spaces a constant output is always possible regardless of the input $\vec{x}(t)$, for which there are the following three examples. Firstly, applications of SFA involving polynomial expansions typically include a constant feature \citep{Berkes2005}, from which a constant output can be produced simply by keeping this feature alone. Secondly, neural network implementations of SFA, such as hierarchical SFA (hSFA) \citep{Franzius2007a}, are typically expressive enough to represent a constant output for any input $\vec{x}(t)$. Thirdly, analytical studies of optimal solutions to SFA typically assume an unrestricted function space \citep{Franzius2007a, Sprekeler2008, Sprekeler2009,Sprekeler2014}, which by definition includes constant signals. Moreover, in this third case all cited papers use a variational calculus formalism in which the solutions are most naturally described in terms of a differential operator $\mathcal{D}$ that always has the constant solution as an eigenfunction.

While Type 1 and Type 2 SFA are both relevant to the results in \cref{sec:Results}, the rest of this section focuses primarily on the former, unless stated otherwise.

\subsection{Optimal free responses in discrete time}
\label{sec:SFAfree}
In most cases, it is not possible to find analytical solutions to the problems presented in the previous section. However, in \citet{Wiskott2003} it was demonstrated that analytical solutions do exist to a simplified version of Type 1 SFA in which \textit{free responses} are considered, i.e.\ the input time series $\vec{x}(t)$ and input-output function $g(\vec{x})$ are ignored and the goal is to find \textit{any} signals $y_j(t)$ that optimize the objective in \cref{eq:Slowness} subject to the constraints in \Crefrange{eq:SFAzeromean}{eq:SFAdecorr}. Formally, this problem is described as follows:\\

\noindent \textbf{Type 1 free response SFA problem}: \textit{Find a set of $J$ real-valued signals $y_j(t)$ such that for $j=1, 2, ..., J$}

\begin{equation}
\Delta(y_j)=\langle \dot{y}_j(t)^2 \rangle_t\quad\text{is minimal}
\end{equation}

\noindent \textit{under the constraints}
\begin{align}
\langle y_j(t) \rangle_t&=0& &\text{(zero mean)}\\
\langle y_j(t)^2 \rangle_t&=1& &\text{(unit variance)} \\
\forall i<j \quad \langle y_i(t)y_j(t)\rangle_t&=0& &\text{(decorrelation and order)}
\end{align}
\noindent \textit{with $\langle \cdot \rangle_t$ and $\dot{y}_j(t)$ representing temporal averaging and the time derivative of $y_j(t)$, respectively.}\\

The free response setting is a useful abstraction to SFA problems since it provides intuition regarding what the solutions typically work towards. This is insightful for two reasons. Firstly, it provides better expectations regarding what the outputs to SFA might be in a given application domain. Secondly, it provides intuition on how the $\Delta$-value relates to the form of each signal, which is useful from the perspective of parameter tuning.
 
In \citep{Wiskott2003}, methods from variational calculus were used to find optimal solutions to the problem described above in the case of continuous time. In the case of free boundary conditions, the solutions are
\begin{equation}
\label{eq:SFAFreeCont}
y_j(t)=\sqrt{2}\text{cos}\bigg(j\pi\frac{t-t_A}{t_B-t_A}\bigg)\quad\text{for}\quad t\in[t_A,t_B]\quad\text{and}\quad j=1, 2, ..., J
\end{equation}
meaning that each solution is equal to a sinusoidal wave defined over this time interval with frequency $f_j=\frac{j\pi}{2\pi(t_B-t_A)}=\frac{j}{2(t_B-t_A)}$. Due to the properties of sinusoidal functions, each solution is guaranteed to have zero mean and to be decorrelated with all others, and the scaling factor of $\sqrt{2}$ enforces unit variance. Note that in continuous time $\Delta(y_j)$ is defined through the average value of the time derivative across the signal length, which can be arbitrarily large. Moreover, \citet{Wiskott2003} showed that for the signals described by \cref{eq:SFAFreeCont} these values are a monotonic function of the frequency $f_j$, which can also be arbitrarily large since there is formally no upper bound on $J$ in the free response setting.

Like the majority of studies on SFA, this paper assumes that time is discrete. While the solutions described by \cref{eq:SFAFreeCont} do not strictly apply in this setting, they can be partially extended by letting $t_A=1,t_B=T$ and sampling at integer time points. Performing this discretization is insightful since it provides a set of approximate solutions to the discrete time version of the free response problem, however when doing this there are a number of technical details worth taking into account. Firstly, for a fixed sampling rate of $f_s$, it only possible to accurately sample continuous signals with an underlying frequency less than or equal to $f_{\text{max}}=\frac{f_s}{2}$, and for $f>f_{\text{max}}$ any sample is equivalent to another obtained for some frequency $f<f_{\text{max}}$.\footnote{In signal processing, this phenomenon is known as \textit{aliasing} and is a consequence of the \textit{Nyquist-Shannon sampling theorem} \citep{Oppenheim2016}.} Therefore, since the free response problem requires decorrelated signals, this upper bound needs to be respected when sampling the solutions described by \cref{eq:SFAFreeCont}. Discretizing at integer time points is equivalent to a sampling rate of $1$Hz, meaning that $f_{\text{max}}=0.5$Hz, and since $f_j=\frac{j}{2(T-1)}$ this frequency is reached at $j=T-1$. The discretized signal obtained at this value of $j$ oscillates between $\pm 1$, i.e.\ $y_{T-1}(t)=[1,-1,1,-1,...]$, and in order to preserve a unit variance of $1$ it is necessary to remove the scaling factor of $\sqrt{2}$ in \cref{eq:SFAFreeCont}. Secondly, although $j$ typically starts at $1$ in SFA problems, inserting $j=0$ into \cref{eq:SFAFreeCont} and removing the factor of $\sqrt{2}$ produces a constant signal with values equal to $1$. This is precisely the positive version of \cref{eq:SFAconstsignal}, and therefore including this value of $j$ offers a useful generalization to the Type 2 setting. Lastly, it is not possible to guarantee that the SFA constraints are exactly satisfied for all signals. This is because the constraints involve averages, i.e.\ $\langle \cdot \rangle_t$, which are sometimes subject to discretization effects when sampling the signals in \cref{eq:SFAFreeCont}. However, this is a finite size effect and therefore becomes less significant as $T$ is increased.

\begin{figure}[h]
\centering
\begin{subfigure}[b]{3.15cm}
\centering
\captionsetup{justification=centering} 
\includegraphics[width=\textwidth]{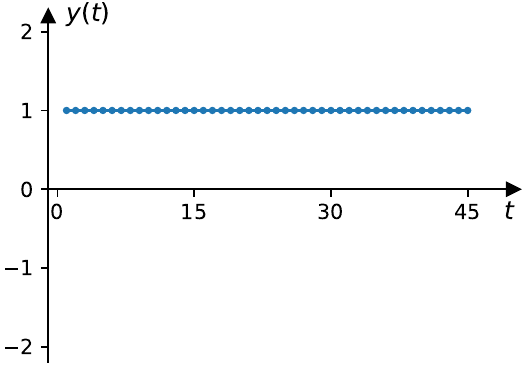}
\caption{$j=0$, $\Delta=0$, $\text{C}_1=1$}
\label{fig:ConstSig}
\end{subfigure}
\hfill
\begin{subfigure}[b]{3.15cm}
\centering
\captionsetup{justification=centering} 
\includegraphics[width=\textwidth]{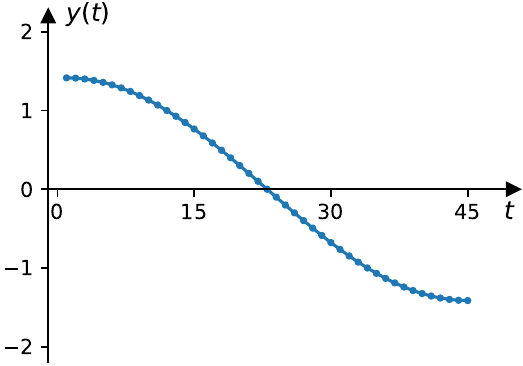}
\caption{$j=1$, $\Delta=0.005$, $\text{C}_1=0.997$}
\label{fig:SlowSig1}
\end{subfigure}
\hfill
\begin{subfigure}[b]{3.15cm}
\centering
\captionsetup{justification=centering} 
\includegraphics[width=\textwidth]{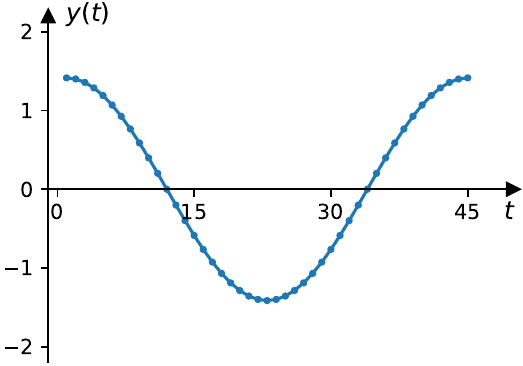}
\caption{$j=2$, $\Delta=0.02$, $\text{C}_1=0.99$}
\label{fig:SlowSig2}
\end{subfigure}
\hfill
\begin{subfigure}[b]{3.15cm}
\centering
\captionsetup{justification=centering} 
\includegraphics[width=\textwidth]{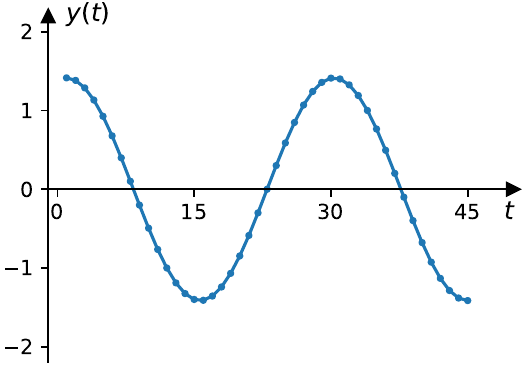}
\caption{$j=3$, $\Delta=0.046$, $\text{C}_1=0.977$}
\label{fig:SlowSig3}
\end{subfigure}
\hfill
\begin{subfigure}[b]{3.15cm}
\centering
\captionsetup{justification=centering}
\includegraphics[width=\textwidth]{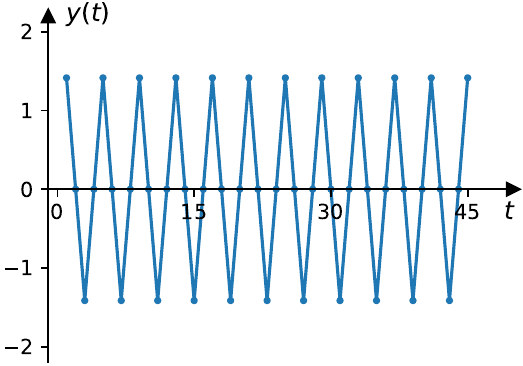}
\caption{$j=22$, $\Delta=2$, $\text{C}_1=0$}
\label{fig:MiddleSig}
\end{subfigure}
\par\bigskip
\begin{subfigure}[b]{3.15cm}
\centering
\captionsetup{justification=centering} 
\includegraphics[width=\textwidth]{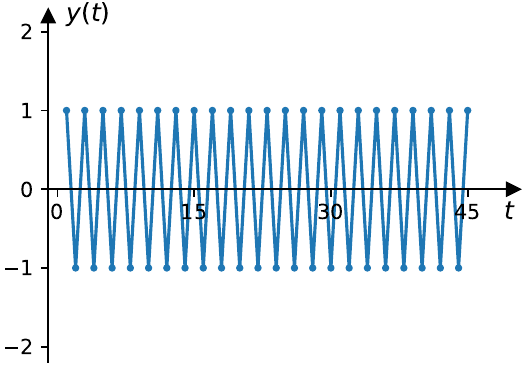}
\caption{$j=44$, $\Delta=4$, $\text{C}_1=-1$}
\label{fig:OscSig}
\end{subfigure}
\hfill
\begin{subfigure}[b]{3.15cm}
\centering
\captionsetup{justification=centering} 
\includegraphics[width=\textwidth]{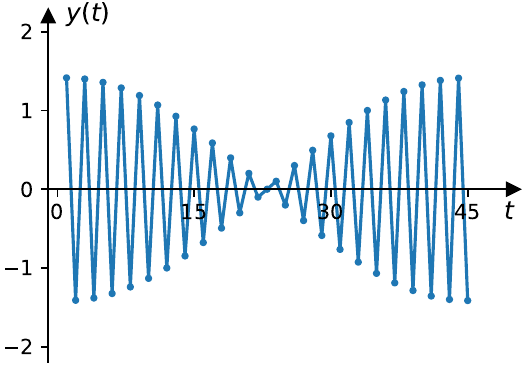}
\caption{$j=43$, $\Delta=3.995$, $\text{C}_1=-0.997$}
\label{fig:FastSig1}
\end{subfigure}
\hfill
\begin{subfigure}[b]{3.17cm}
\centering
\captionsetup{justification=centering} 
\includegraphics[width=3.15cm]{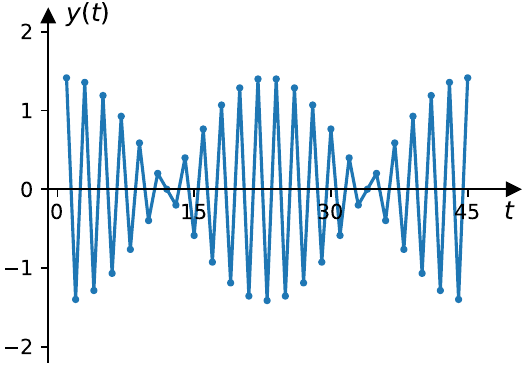}
\caption{$j=42$, $\Delta=3.98$, $\text{C}_1=-0.99$}
\label{fig:FastSig2}
\end{subfigure}
\hfill
\begin{subfigure}[b]{3.17cm}
\centering
\captionsetup{justification=centering} 
\includegraphics[width=3.15cm]{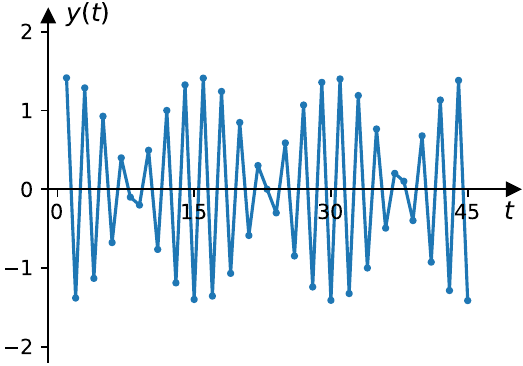}
\caption{$j=41$, $\Delta=3.954$, $\text{C}_1=-0.977$}
\label{fig:FastSig3}
\end{subfigure}
\begin{subfigure}[b]{3.17cm}
\centering
\captionsetup{justification=centering} 
\includegraphics[width=3.15cm]{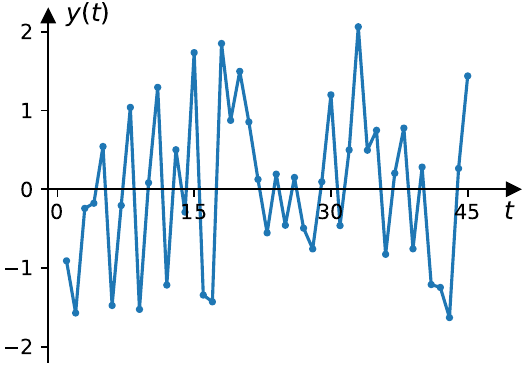}
\caption{$\Delta=1.923$, $\text{C}_1=0.038$}
\label{fig:RandSig}
\end{subfigure}
\caption{An illustration of the discrete time free responses described by \cref{eq:SFAFreeDisc,eq:SFAFreeDiscScale} for $T=45$. (a-d) and (f-i) show the signals corresponding to the four smallest and largest values of $j$, respectively, while (e) shows the signal for the middlemost value of $j$. For comparison, in (j) an output is shown resembling a realistic output of SFA with value of $\Delta$ and $\text{C}_1$ that is far away from the extreme cases considered in (a-d) and (f-i).}
\label{fig:SlowFastDelta}
\end{figure}

Provided all details outlined in the previous paragraph are taken into account, the discrete time analogues of the solutions described in \cref{eq:SFAFreeCont} are given by
\begin{equation}
\label{eq:SFAFreeDisc}
y_j(t)=c_j\text{cos}\bigg(j\pi \frac{t-1}{T-1}\bigg)\quad\text{for}\quad t=1,2,...,T\quad\text{and}\quad j=0, 1, 2, ..., T-1
\end{equation}
where
\begin{equation}
\label{eq:SFAFreeDiscScale}
c_j=\left\{
	\begin{array}{ll}
		1 & \mbox{if } j=0\mbox{ or } T-1\\
		\sqrt{2} & \mbox{otherwise }
	\end{array}
\right.
\end{equation}
A number of the signals described in \cref{eq:SFAFreeDisc,eq:SFAFreeDiscScale} are depicted in \cref{fig:SlowFastDelta} for $T=45$. In \cref{fig:ConstSig,fig:OscSig}, the aforementioned cases of $j=0$ and $j=T-1=44$ are depicted, respectively. In the former case, the signal does not change at all and therefore has $\Delta=0$, whereas in the latter case the difference between successive time points is always $2$, which means that $\Delta=4$. Moreover, since these are the slowest and fastest possible signals of length $T=45$, they provide lower and upper bounds for the $\Delta$-value, i.e.\ $\Delta\in[0,4]$. Conversely, successive values of the signal in \cref{fig:ConstSig} are perfectly correlated over time, i.e.\ $\text{C}_1=1$, whereas those of the signal in \cref{fig:OscSig} are perfectly anti-correlated over time, i.e.\ $\text{C}_1=-1$. Furthermore, the interval to which $\Delta$ belongs implies a corresponding interval of $\text{C}_1\in[-1,1]$ (see Equation \ref{eq:SlownessAC4}), which is consistent with the notion of how correlation coefficients are bounded. It should be emphasized that since $y_0(t)$ is only a possible solution if the zero mean constraint is removed, the intervals $\Delta\in[0,4]$ and $\text{C}_1\in[-1,1]$ strictly apply to Type 2 SFA, while for Type 1 SFA the values $\Delta=0$ and $\text{C}_1=1$ are not possible, in which case $\Delta\in(0,4]$ and $\text{C}_1\in[-1,1)$. Note that the restriction of $\Delta$ to these intervals is one key difference to the continuous time setting considered by \citet{Wiskott2003} in which $\Delta$ can be arbitrarily large. The reason for this difference is that in discrete time $\Delta$ is approximated with the differences between neighbouring values in $y_j(t)$ (see Equation \ref{eq:TimeDeriv}) and these differences are limited by the unit variance constraint.

In \Crefrange{fig:SlowSig1}{fig:SlowSig3}, the signals corresponding to $j=1,2,3$ are depicted, and the cosine form described in \cref{eq:SFAFreeDisc} is clearly visible. Since all of these signals do not change much between neighbouring time points, the $\Delta$-values are close to $0$ and increase with the frequency, while the values of $\text{C}_1$ are close to $1$ and decrease with the frequency. In \Crefrange{fig:FastSig1}{fig:FastSig3}, the cases corresponding to $j=T-2,T-3,T-4=43,42,41$ are depicted. Like the signal in \cref{fig:OscSig}, these signals all change a lot on average between successive time points, and therefore have $\Delta\approx 4$ and $\text{C}_1\approx -1$, again with the former (latter) increasing (decreasing) with the frequency. Visually, the signals in \Crefrange{fig:FastSig1}{fig:FastSig3} take the form of beat patterns that would emerge when performing amplitude modulation (AM) of the signal in \cref{fig:OscSig} using those in \Crefrange{fig:SlowSig1}{fig:SlowSig3}. This pairing between signals is further evident in the fact that the values of $\text{C}_1$ for the signals in \Crefrange{fig:FastSig1}{fig:FastSig3} are of equal magnitude and opposite sign to those of the signals in \Crefrange{fig:SlowSig1}{fig:SlowSig3}, or equivalently the $\Delta$-values are equally far from the lower/upper bounds of $0$ and $4$.

Equidistant from $j=0$ and $j=T-1$ is $j=\frac{T-1}{2}=22$, for which the corresponding signal is shown in \cref{fig:MiddleSig}. This signal takes the form of a fixed oscillation $y(t)=[\sqrt{2}, 0, -\sqrt{2}, 0,\sqrt{2}, \cdots]$, and is therefore analogous to the one in \cref{fig:OscSig}, but instead having a frequency half as large and the scaling factor $\sqrt{2}$ included. Moreover, this signal has $\Delta=2$, which is exactly half as large as the $\Delta$-value of the signal in \cref{fig:OscSig}.\footnote{It is worth noting that the signal for $j=\frac{T-1}{2}$ has this pattern and a $\Delta$-value of exactly $2$ only if $T$ is odd. Since this signal offers useful insight on the middle of the interval $\Delta\in[0,4]$, an odd value of $T=45$ is considered here.}

Considering the signals in \Crefrange{fig:ConstSig}{fig:FastSig3}, all display a high degree of regularity. Thus, for any of these signals, it is easy to predict future values given knowledge of a value at some time $t$, meaning that all of them, both slow and fast, are highly \textit{predictable} \citep{Creutzig2008}\footnote{Note that if features are selected based on predictability, i.e.\ the slowest and fastest signals, then the corresponding method is related to predictive coding and information bottlenecks \citep{Creutzig2008}.}. In fact, for the basis described by \cref{eq:SFAFreeDisc,eq:SFAFreeDiscScale} this property holds for all values of $j$ by virtue of the periodicity of the underlying cosine functions. However, this property is a pathological feaure of the free response problem. In real applications of SFA, it is only the signals with $\Delta\approx 0,4$ that are highly predictable, and the majority of other signals resemble noise \citep{Creutzig2008, Escalante2020}. Thus, signals with $\Delta$-values far away from $0$ and $4$ are unlikely to have a profile like the signal in \cref{fig:MiddleSig}, and are instead more likely to be highly irregular like the signal illustrated in \cref{fig:RandSig}, which is an i.i.d.\ sample of Gaussian noise with zero mean and unit variance.

\subsection{Generalization to time-lags larger than 1}
\label{sec:SFAgentaugeq1}
The quantities $\Delta(y_j)$ and $\text{C}_1(y_j)$ introduced in \cref{sec:SFAdef} are both computed using successive data points in the time series $\vec{x}(t)$. For the $\Delta$-value, this is necessary in order to maintain an analogy to the continuous-time derivative. By contrast, temporal correlations can be computed between data points that are separated by time lags larger than $1$. In particular, provided that $y_j(t)$ satisfies the zero mean and unit variance constraints of Type 1 SFA, then the time-lagged correlation of $y_j(t)$ for a general time-lag $\tau\geq 1$ is given by
\begin{equation}
\text{C}_\tau(y_j)=\langle y_j(t)y_j(t+\tau)\rangle_t\label{eq:TimeLagACtau}
\end{equation}
which reduces to $\text{C}_1(y_j)$ for $\tau=1$, and therefore generalizes this quantity. 

Replacing the CSFA objective with the maximization of $\text{C}_\tau(y_j)$ is one way to generalize this problem to larger time scales. In the literature, this variant of SFA has initially been explored as a method for performing \textit{independent component analysis} (ICA), in particular when statistical independence is defined in terms of temporal decorrelation \citep{Belouchrani1997,Molgedey1994,Ziehe1998,Blaschke2006}. More recently, this method has become a popular tool for the construction of so-called \textit{Markov state models} in the field of molecular dynamics simulation, where it is referred to as \textit{time-lagged independent component analysis} (tICA) \citep{Naritomi2011,Perez-Hernandez2013,Moffett2017,Klus2018,Schultze2021,Husic2018}. However, in order to emphasize that this is a generalization of CSFA to larger time-lags \citep{Blaschke2006}, and to make the terminology of the current paper more consistent, this variant of SFA is henceforth referred to as $\tau$SFA.

Like in the case of SFA and CSFA, $\tau$SFA can be defined either with or without the zero mean constraint, which for consistency is referred to as Type 1 and Type 2 $\tau$SFA, respectively. Moreover, all details outlined in \cref{sec:T2SFA} regarding the relationship between Type 1 and Type 2 SFA/CSFA apply equally in the case of $\tau$SFA.

Some intuition regarding the relationship of $\text{C}_\tau(y_j)$ and $\text{C}_1(y_j)$ can be gained by applying the former to the signals in \cref{fig:SlowFastDelta} and comparing to the insights found in \cref{sec:SFAfree}. The most important factor that determines whether $\text{C}_\tau(y_j)$ orders these signals in the same way as the $\tau=1$ case is the parity of $\tau$. For even $\tau$, slow signals are ordered in the same way, but fast signals are assigned large positive correlation, while for odd $\tau$ the ordering is in full agreement with the $\tau=1$ case. For example, for a time-lag of $\tau=2$, the values of the constant signal in \cref{fig:ConstSig} are perfectly correlated over time, meaning that $\text{C}_2=1$, and the other slow signals in \Crefrange{fig:SlowSig1}{fig:SlowSig3} have highly correlated values for this time lag, meaning that $\text{C}_2\approx 1$. Therefore, for $\tau=2$ all slow signals have large correlations either equal or close to $1$, in agreement with the $\tau=1$ case. By contrast, the signal in \cref{fig:OscSig} oscillates between $\pm 1$, meaning that the values $2$ steps apart are equal, and therefore $\text{C}_2=1$. Likewise, for the other fast signals in \Crefrange{fig:FastSig1}{fig:FastSig3}, the values $2$ steps apart are similar over time, and therefore $\text{C}_2\approx 1$. Note in particular that due to the structure of the cosine waves, for $\tau=2$ the value of $\text{C}_2$ for each signal in \Crefrange{fig:ConstSig}{fig:SlowSig3} is exactly same as the value of its fast counterpart in \Crefrange{fig:OscSig}{fig:FastSig3}, which contrasts with the $\tau=1$ case. Now compare this to a time-lag of $\tau=3$. In this case, the slow signals in \Crefrange{fig:ConstSig}{fig:SlowSig3} have high values of $\text{C}_3$ either equal or close to $1$, in agreement with both the $\tau=1,2$ cases, which can be demonstrated using the same argument as above. However, for the fast signals the values that are spaced by $3$ steps are very different, meaning that they have low values of $\text{C}_3$ either equal or close to $-1$, which agrees with the $\tau=1$ case but disagrees with the $\tau=2$ case. To summarize, many of the intuitive properties regarding how $\text{C}_1(y_j)$ measures the variation of a signal over time carry over to $\text{C}_\tau(y_j)$, however for even values of $\tau$ it is important to be careful when dealing with quickly oscillating signals.

\subsection{Generalization to multiple time-lags}
\label{sec:SFAgentaumulti}
A number of studies have extended $\tau$SFA by considering correlations across multiple time-lags simultaneously \citep{Ziehe1998,Blaschke2006, Wang2020}. These methods all aim to maximize a quantity that sums over correlations at different time-lags, i.e.\
\begin{equation}
\text{F}_\gamma(y_j)=\sum_{\tau=0}^{\tau_{\text{max}}}\kappa_\tau\text{C}_\tau(y_j)\label{eq:LFquantity1}
\end{equation}
where $\kappa_\tau$ are a set of weights specifying the contribution at each time-lag $\tau$. While \citet{Ziehe1998} and \citet{Wang2020} consider all time-lags equally, i.e.\ $\kappa_\tau=1\;\forall \tau$, \citet{Blaschke2006} use a set of exponentially decaying weights, i.e.\ $\kappa_\tau:=\text{exp}(-\beta\tau)$ with $\beta$ controlling the speed of decay. The latter formulation is focused on in this paper as it is more relevant to the results in \cref{sec:Results}. Assuming a parameter $\gamma\in(0,1)$, analogous to the one in RL, exponentially decaying weights can be written as $\kappa_\tau=\gamma^\tau$, meaning that
\begin{equation}
\text{F}_\gamma(y_j)=\sum_{\tau=0}^{\tau_{\text{max}}}\gamma^\tau\text{C}_\tau(y_j)\label{eq:LFquantity2}
\end{equation}
is the exact quantity to be maximized. Since the $\tau=1$ term in \cref{eq:LFquantity2} corresponds to CSFA, this is perhaps a natural choice for the starting value of $\tau$. However, like in \citet{Ziehe1998}, this paper uses a starting value of $\tau=0$, and there are two reasons for this choice. Firstly, starting at $\tau=0$ leads to more insightful results in \cref{sec:Results}. Secondly, assuming the unit variance constraint has been satisfied, then including an additional $\tau=0$ term in practice only adds $\gamma^0\langle y_j(t)y_j(t)\rangle_t=\langle y_j(t)^2\rangle_t=1$ to the sum in \cref{eq:LFquantity2}, meaning that the ordering of signals is invariant to this detail. The maximum value $\tau_{\text{max}}$ is left as a free parameter, however it is important to note that it is in practice bound from above by the length of the input time series.

The variant of SFA involving the maximization of $\text{F}_\gamma(y_j)$ has, like $\tau$SFA, mostly been studied as a technique for performing ICA based on temporal decorrelation \citep{Ziehe1998,Blaschke2006, Wang2020}. In \citet{Blaschke2006}, it was referred to as \textit{linear filtering}, and this terminology can be understood through the following rearrangement of \cref{eq:LFquantity2}:
\begin{align}
\text{F}_\gamma(y_j)&=\sum_{\tau=0}^{\tau_{\text{max}}}\gamma^\tau\text{C}_\tau(y_j)\\
&=\sum_{\tau=0}^{\tau_{\text{max}}}\gamma^\tau \langle y_j(t)y_j(t+\tau)\rangle_t\\
&=\bigg\langle y_j(t)\bigg(\sum_{\tau=0}^{\tau_{\text{max}}} \gamma^\tau y_j(t+\tau)\bigg)\bigg\rangle_t\label{eq:LinearFiltercorr}
\end{align}
\cref{eq:LinearFiltercorr} says that computing $\text{F}_\gamma(y_j)$ is equivalent to first applying an exponentially decaying filter to $y_j(t)$ and then computing the correlation of $y_j(t)$ with the resulting signal.\footnote{Note that in \citep{Foldiak1991} an analogous quantity is used to model memory traces as part of a learning rule for extracting slowly varying features.} For this reason, $\text{F}_\gamma(y_j)$ is referred to henceforth as the \textit{linearly filtered correlation} of $y_j(t)$, and the variant of SFA that seeks to maximize this quantity is referred to in this paper as \textit{linear filtering slow feature analysis} (LFSFA). Note that due to the use of exponentially decaying weights, maximizing \cref{eq:LinearFiltercorr} strongly favours highly correlated/slow varying signals, which in some sense is related to low-pass filtering. As explained in \citep{Boehmer2015}, in some learning contexts this can lead to pathological outputs, and in such cases it might be more appropriate to experiment with choices of weights $\kappa_\tau$ that instead correspond to band-pass filtering.

Like all problems introduced so far, LFSFA can be defined either with or without the zero mean constraint, which is referred to as Type 1 and Type 2 LFSFA, respectively. Moreover, the two types are related in an analogous way to the case of SFA and CSFA, as outlined in \cref{sec:T2SFA}.

Note that $\text{F}_\gamma(y_j)$ is not applied to the free response signals in \cref{fig:SlowFastDelta} since this provides less insight in comparison to $\Delta(y_j)$ and $\text{C}_\tau(y_j)$.

\subsection{Solving Type 1 SFA problems}
\label{sec:Type1sol}

This section describes how each variant of Type 1 SFA presented in the previous section can be solved in the form of a generalized or regular eigenvalue problem. It should be emphasized that while the linear setting of each of these problems is considered in this paper, the extension to non-linear transformations can be dealt with in an equivalent manner but with the input $\vec{x}(t)$ exchanged for a polynomially expanded signal $\vec{h}(t)$ \citep{Sprekeler2014}. An introduction to both regular and generalized eigenvalue problems can be found in Appendix \ref{app:EVall}. Moreover, note that while the vast majority of approaches to SFA are based, at least in part, on the framework of eigenvalue problems, one alternative is the gradient-based approach developed by \citet{Schueler2019}. 

\subsubsection{SFA}
\label{sec:Type1solSFA}
If a centering process is applied to the raw signal $\vec{x}(t)$, i.e.\ $\vec{c}(t)=\vec{x}(t)-\langle \vec{x}(t)\rangle_t$, then the linear Type 1 SFA problem reduces to the following generalized eigenvalue problem \citep{Berkes2005,Creutzig2008}
\begin{equation}
\label{eq:linSFAgenEVmat1}
\dot{\mat{\Sigma}}\mat{W}=\mat{\Sigma}\mat{W\Lambda}'
\end{equation}
where $\mat{\Sigma}=\langle \vec{c}(t)\vec{c}(t)^T\rangle_t$ and $\dot{\mat{\Sigma}}=\langle \dot{\vec{c}}(t)\dot{\vec{c}}(t)^T\rangle_t$ are the \textit{covariance matrices} of the input signal $\vec{x}(t)$ and its time derivative $\dot{\vec{x}}(t)$, respectively, with the former arising from the constraints in \cref{eq:SFAunitvar,eq:SFAdecorr} and the latter from the definition of slowness in \cref{eq:Slowness}. The matrix $\mat{W}\in\mathbb{R}^{N\times J}$ contains $J$ generalized eigenvectors of $(\dot{\mat{\Sigma}},\mat{\Sigma})$, which are precisely the optimal weight vectors $\vec{w}_j$ and which satisfy the following relation:
\begin{equation}
\label{eq:linSFAgenEV_orthog}
\mat{W}^T\mat{\Sigma}\mat{W}=\mat{\mathbbm{1}}
\end{equation}
Since $\mat{\Sigma}$ is a covariance matrix, it is guaranteed to be positive semi-definite \citep{Potters2020}, and in the case where it is positive definite \cref{eq:linSFAgenEV_orthog} says that the weight vectors are orthonormal w.r.t\ the associated inner product $(\cdot, \cdot)_{\mat{\Sigma}}$ (for a definition of positive definite and positive demi-definite matrices, see Appendix \ref{app:EVprob}). The matrix $\mat{\Lambda}'\in\mathbb{R}^{J\times J}$ is a diagonal matrix containing the corresponding generalized eigenvalues on the diagonals, which are equal to the $\Delta$-values and are therefore bound by $\Lambda'_{jj}\in (0,4]$. Therefore, SFA consists in finding the generalized eigenvectors of $(\dot{\mat{\Sigma}},\mat{\Sigma})$ that have the smallest generalized eigenvalues.

It should be noted that due to the centering step, each output $y_j(t)$ is given by
\begin{align}
y_j(t)&=\vec{w}_j^T\vec{c}(t)\label{eq:SFAaffine1}\\
&=\vec{w}_j^T(\vec{x}(t)-\langle \vec{x}(t)\rangle_t)\\
&=\vec{w}_j^T\vec{x}(t)-\vec{w}_j^T\langle \vec{x}(t)\rangle_t\\
&=\vec{w}_j^T\vec{x}(t)-a_j\label{eq:SFAaffine4}
\end{align}
for some scalar constant $a_j$. Therefore, contrary to the terminology, linear Type 1 SFA formally involves affine input-output functions.

\subsubsection{CSFA}
\label{sec:Type1solCSFA}
In \cref{sec:SFAdef}, the objectives of SFA and CSFA are related by \cref{eq:SlownessAC4}. By applying a similar set of steps as in \Crefrange{eq:SlownessAC1}{eq:SlownessAC4}, it is possible to deduce an analogous relationship between $\dot{\mat{\Sigma}}$ and a matrix based on time-lagged statistics of the input \citep{Blaschke2006,Creutzig2008}. In particular:
\begin{align}
\dot{\mat{\Sigma}}&=\langle\dot{\vec{c}}(t)\dot{\vec{c}}(t))^T\rangle_t\label{eq:ACmatfirst}\\
&=\langle(\vec{c}(t+1)-\vec{c}(t))(\vec{c}(t+1)-\vec{c}(t)^T)\rangle_t\label{eq:ACmatsecond}\\
&=\langle\vec{c}(t+1)\vec{c}(t+1)^T\rangle_t+\langle\vec{c}(t)\vec{c}(t)^T\rangle_t-\langle\vec{c}(t)\vec{c}(t+1)^T\rangle_t-\langle\vec{c}(t+1)\vec{c}(t)^T\rangle_t\\
&=\langle\vec{c}(t+1)\vec{c}(t+1)^T\rangle_t+\langle\vec{c}(t)\vec{c}(t)^T\rangle_t-2\bigg(\underbrace{\frac{\langle\vec{c}(t)\vec{c}(t+1)^T\rangle_t+\langle\vec{c}(t+1)\vec{c}(t)^T\rangle_t}{2}}_{=\mat{\Omega}_1}\bigg)\\
&=2(\mat{\Sigma}-\mat{\Omega}_1)\label{eq:ACmatlast}
\end{align}
The matrix $\mat{\Omega}_1$ can be thought of as a matrix analogue of the quantity $\text{C}_1(y_j)$ from the objective of CSFA, and before moving on it is worth pointing out a number of details about this quantity. Firstly, since $\langle\vec{c}(t+1)\vec{c}(t)^T\rangle_t=(\langle\vec{c}(t)\vec{c}(t+1)^T\rangle_t)^T$, the matrix $\mat{\Omega}_1$ has the form $\frac{1}{2}(\mat{A}+\mat{A}^T)$ for $\mat{A}=\langle\vec{c}(t)\vec{c}(t+1)^T\rangle_t$ and is therefore symmetric. Secondly, $\mat{A}$ contains second moments between the entries of $\vec{c}(t)$ and $\vec{c}(t+1)$, or equivalently covariances between the entries of $\vec{x}(t)$ and $\vec{x}(t+1)$, and can therefore be interpreted as a \textit{time-lagged covariance matrix}. By extension, $\mat{\Omega}_1$ can be interpreted as a \textit{symmetrized time-lagged covariance matrix}, or \textit{STC matrix} for short. Thirdly, while $(\mat{A})_{ij}=\langle c_i(t)c_j(t+1)^T\rangle_t$ describes covariances between $x_i$ and $x_j$ in the forward direction, i.e.\ $t\to t+1$, the corresponding entry of $\mat{A}^T$ can be expressed as $(\mat{A}^T)_{ij}=\langle c_i(t+1)c_j(t)^T\rangle_t=\langle c_i(t)c_j(t-1)^T\rangle_t$ and therefore describes covariances between $x_i$ and $x_j$ in the backward direction, i.e.\ $t\to t-1$. Thus, the symmetrization in $\mat{\Omega}_1$ effectively averages the forward and backward covariances and is therefore analogous to the additive reversibilization described in \cref{sec:SReig}. This insight is relevant to the results in \cref{sec:Results}.

The generalized eigenvalue problem described in \cref{eq:linSFAgenEVmat1} can be transformed into one based on $\mat{\Omega}_1$, which describes the solutions to the Type 1 CSFA problem. In particular, inserting \cref{eq:ACmatlast} into \cref{eq:linSFAgenEVmat1} gives
\begin{equation}
2(\mat{\Sigma}-\mat{\Omega}_1)\mat{W}=\mat{\Sigma}\mat{W\Lambda}'\label{eq:linSFAgenEV_swapAC}
\end{equation}
which can be easily rearranged to
\begin{align}
\mat{\Omega}_1\mat{W}&=\mat{\Sigma}\mat{W}(\mat{\mathbbm{1}}-\frac{1}{2}\mat{\Lambda}')\\
&=\mat{\Sigma}\mat{W}\mat{\Lambda}\label{eq:linSFAgenEVmatAC}
\end{align}
Note that \cref{eq:linSFAgenEVmat1} and \cref{eq:linSFAgenEVmatAC} have the same generalized eigenvectors, which is in agreement with the insight from \cref{sec:SFAdef,sec:SFAfree} that the solutions of SFA and CSFA are the same. The main difference between the equations is the eigenvalue arrays, which are related by $\mat{\Lambda}=(\mat{\mathbbm{1}}-\frac{1}{2}\mat{\Lambda}')$, meaning that for CSFA the generalized eigenvectors with largest generalized eigenvalues are optimal. Moreover, each eigenvalue associated to the CSFA problem can be expressed as
\begin{align}
\Lambda_{jj}&=1-\frac{1}{2}\Lambda'_{jj}\\
&=1-\frac{1}{2}\Delta(y_j)\\
\overset{(\ref{eq:SlownessAC4})}&{=}
1-\frac{2(1-\text{C}_1(y_j))}{2}\label{eq:EigenvalCorr}\\
&=1-1+\text{C}_1(y_j)\\
&=\text{C}_1(y_j)
\end{align}
meaning that $\Lambda_{jj}\in [-1,1)$. To summarize, the generalized eigenvalue problems described by \cref{eq:linSFAgenEVmatAC,eq:linSFAgenEVmat1} capture the notion that SFA and CSFA differ only in terms of whether slowness is measured with $\Delta(y_j)$ or $\text{C}_1(y_j)$, respectively.

\subsubsection{$\tau$SFA}
\label{sec:Type1soltauSFA}
Since $\tau$SFA is a generalization of CSFA to larger time-lags, its corresponding generalized eigenvalue problem shares much in common with the one presented in the previous section. In particular, the solutions to the $\tau$SFA problem can be formulated as \citep{Blaschke2006}
\begin{equation}
\mat{\Omega}_\tau\mat{W}=\mat{\Sigma}\mat{W}\mat{\Lambda}
\label{eq:ACgenEVtau}
\end{equation}
where
\begin{align}
\mat{\Omega}_\tau&=\frac{1}{2}\big(\langle\vec{c}(t)\vec{c}(t+\tau)^T\rangle_t+\langle\vec{c}(t+\tau)\vec{c}(t)^T\rangle_t\big)\label{eq:ACMattau}
\end{align}
is the STC matrix with time-lag $\tau$, which corresponds to the objective of $\tau$SFA and generalizes $\mat{\Omega}_1$ to $\tau\geq 1$. Therefore, \cref{eq:ACgenEVtau} generalizes \cref{eq:linSFAgenEVmatAC} to $\tau\geq 1$, in agreement with the relation of $\tau$SFA to CSFA. Moreover, for this problem it is possible to show that $\Lambda_{jj}=\text{C}_\tau(y_j)$,\footnote{In particular, this requires the method of Lagrange multipliers as in \citet{Creutzig2008} but with $\dot{\mat{\Sigma}}$ exchanged for $\mat{\Omega}_\tau$.} meaning that $\Lambda_{jj}\in[-1,1)$ and the generalized eigenvectors with largest generalized eigenvalues are optimal, like in CSFA. It should be emphasized that the symbols $\mat{\Lambda}$ and $\mat{W}$ in \cref{eq:ACgenEVtau} are used simply as place holders for eigenvalues and eigenvectors, are generally not the same as the quantities labelled in this way in \cref{eq:linSFAgenEVmatAC}. This notation is maintained henceforth throughout the paper.

In order to make the following sections of the paper more concise, focus is placed on the more general method of $\tau$SFA, with CSFA being discussed as a special case whenever necessary.

\subsubsection{LFSFA}
\label{sec:Type1solLFSFA}
Since the LFSFA objective involves correlations at multiple time-lags, with all other aspects of the problem equivalent to the others presented so far, the solutions to this algorithm resemble those of $\tau$SFA provided that multiple time scales are included with temporal discounting. In particular, the generalized eigenvalue problem corresponding to LFSFA is \citep{Blaschke2006}
\begin{equation}
\mat{\Psi}_\gamma\mat{W}=\mat{\Sigma}\mat{W}\mat{\Lambda}
\label{eq:LFgenEV}
\end{equation}
where
\begin{align}
\mat{\Psi}_\gamma&=\sum_{\tau=0}^{\tau_{\text{max}}}\gamma^\tau\mat{\Omega}_\tau\\
&=\sum_{\tau=0}^{\tau_{\text{max}}}\frac{\gamma^\tau}{2}\big(\langle\vec{c}(t)\vec{c}(t+\tau)^T\rangle_t+\langle\vec{c}(t+\tau)\vec{c}(t)^T\rangle_t\big)\label{eq:LFmat}
\end{align}
is referred to as the \textit{LF matrix}. In analogy to the previous sections, it can be shown that $\Lambda_{jj}=\text{F}_\gamma(y_j)$, and although this does not imply a specific interval for the eigenvalues, it does offer the insight that the generalized eigenvectors with largest generalized eigenvalues are optimal, like for $\tau$SFA. Moreover, the $\tau=0$ term contributes the following term to the LF matrx:
\begin{align}
\frac{\gamma^0}{2}\big(\langle\vec{c}(t)\vec{c}(t)^T\rangle_t+\langle\vec{c}(t)\vec{c}(t)^T\rangle_t\big)&=\frac{1}{2}\big(2\mat{\Sigma}\big)\\
&=\mat{\Sigma}
\end{align}
Thus, in comparison to using a starting value of $\tau=1$, all generalized eigenvalues $\Lambda_{jj}$, or equivalently all linearly filtered correlations $\text{F}_\gamma(y_j)$, are increased by a value of $1$, which is consistent with the analysis of $\text{F}_\gamma(y_j)$ given in \cref{sec:SFAgentaumulti}.

\subsubsection{Corresponding regular eigenvalue problems}
\label{sec:Type1Norm}

The preceding sections show that the solutions to each Type 1 variant of SFA considered in this paper are given by 
\begin{equation}
\mat{AW}=\mat{\Sigma W\Lambda}\label{eq:T1GenEV}
\end{equation}
where $\mat{A}$ is symmetric and $\mat{\Sigma}$ is the data covariance matrix, which is guaranteed to be positive semi-definite \citep{Potters2020}. These problems can either be solved directly using various numerical techniques, or they can first be converted to regular eigenvalue problems. The latter option is relevant both analytically, since regular eigenvalue problems are typically easier to solve, but also practically, since most approaches to SFA involve such a conversion either implicitly or explicitly. Moreover, since the solutions to SFA problems are essentially independent of whether a generalized or regular eigenvalue problem is used, the specific choice is referred to as a \textit{formulation} of SFA in this paper. In particular, generalized eigenvalue problems are referred to as \textit{unnormalized} formulations while regular eigenvalue problems are referred to as \textit{normalized} formulations.

If $\mat{\Sigma}$ is positive definite, i.e.\ if it has no eigenvalues equal to zero, then \cref{eq:T1GenEV} can be converted to a regular eigenvalue problem using either symmetric or left normalization, which are defined in Appendix \ref{app:GenEVprob}. This gives
\begin{equation}
\mat{\Sigma}^{-\frac{1}{2}}\mat{A}(\mat{\Sigma}^{-\frac{1}{2}})^T\widetilde{\mat{W}}=\widetilde{\mat{W}}\mat{\Lambda}\label{eq:T1EVsym}
\end{equation}
and
\begin{equation}
\mat{\Sigma}^{-1}\mat{A}\mat{W}=\mat{W\Lambda}\label{eq:T1EVleft}
\end{equation}
respectively, where $\widetilde{\mat{W}}=(\mat{\Sigma}^{\frac{1}{2}})^T\mat{W}$ and where $\mat{\Sigma}^{\frac{1}{2}}$ can be any positive root of $\mat{\Sigma}$ (see Appendix \ref{app:SqrtPSD}). In words, \cref{eq:T1EVsym,eq:T1EVleft} describe the symmetric and left normalized formulations of Type 1 SFA problems.

In the more general case where $\mat{\Sigma}$ has at least one eigenvalue equal to zero, symmetric and left normalization are not well-defined because they both require matrix inversion, i.e.\ $\mat{\Sigma}^{-\frac{1}{2}}$ and $\mat{\Sigma}^{-1}$, respectively. A covariance matrix has eigenvalues equal to zero whenever one or more variables in a data set can be expressed as a linear combination of the others, or equivalently when the data points live in a lower-dimensional subspace of $\mathbb{R}^N$ with the remaining dimensions having zero variation. If $\mat{\Sigma}$ has $M$ eigenvalues equal to zero, then this subspace has dimension $N-M$, which is also the total number of SFA solutions that are possible in this case. In applications of SFA where $\mat{\Sigma}$ is non-invertible, the normalization methods can be generalized in the following two ways.

Firstly, a preprocessing phase can be included in which the input $\vec{x}(t)$ is perturbed by noise. This introduces variations into the $M$ dimensions described above and therefore removes any corresponding eigenvalues of zero. Thus, after adding noise both $\mat{\Sigma}^{-\frac{1}{2}}$ and $\mat{\Sigma}^{-1}$ can be computed, meaning that the normalization methods can be applied. One disadvantage of this method is that $\vec{x}(t)$ is perturbed in all directions, including the $N-M$ dimensions that already contain variation. Therefore, this method is not specifically aimed at eliminating linear dependencies since it also impacts any genuine slow features in $\vec{x}(t)$. Moreover, in cases where constant signals can be represented, the noise added along the $N-M$ dimensions that contain variation disrupts the equivalence between Type 1 and Type 2 SFA solutions (see \cref{sec:T2SFA}). However, one advantage of this method is that for Gaussian noise with variance $\sigma^2$, which is a typical choice for SFA \citep{Konen2009,Franzius2007a}, the structure of $\mat{\Sigma}$ remains the same in the large data limit except for an additional diagonal term $\sigma^2\mat{\mathbbm{1}}$ (see Appendix \ref{app:LinSFAnoise}). Thus, for a suitable level of noise and a long enough time series this preserves many aspects of the generalized eigenvalue problem while allowing it to be normalized. Because of this, noise is used to deal with non-invertible covariance matrices in \cref{sec:Results}, where the focus is on large data limits of SFA problems. Note that noise is added only when necessary, in which case the covariance matrix is written as $\mat{\Sigma}+\sigma^2\mat{\mathbbm{1}}$ in order to explicitly acknowledge the impact this has.

Secondly, the matrix inverses $\mat{\Sigma}^{-\frac{1}{2}}$ and $\mat{\Sigma}^{-1}$ can be swapped for the respective \textit{pseudoinverses}, which can be computed using the eigendecomposition of $\mat{\Sigma}$, as in the original SFA publication \citep{Wiskott2002}.\footnote{Note that \citet{Wiskott2002} refer to this as the singular value decomposition (SVD) of $\mat{\Sigma}$, but since this matrix is positive semi-definite the two are equivalent.} This effectively constrains the process of matrix inversion to the subspace spanned by the eigenvectors of $\mat{\Sigma}$ with $\lambda>0$, or equivalently to the $N-M$ subspace in which $\vec{x}(t)$ has variation. This method has various advantages over adding noise. For example, by leaving the input data $\vec{x}(t)$ intact, the slow features in $\vec{x}(t)$ are undisturbed. Moreover, in cases where constant signals can be represented, performing SFA with the pseudoinverse preserves exactly the correspondence between Type 1 and Type 2 SFA solutions. In particular, $N-M$ slow features are obtained as well as an additional constant feature in the case of Type 2 SFA (see \cref{sec:T2SFA}).

For each Type 1 SFA problem explored in the previous sections, all formulations are shown in the upper half of column A in \cref{tab:AllEigenvalueProblems}. Since a general input $\vec{x}(t)$ without zero eigenvalues in $\mat{\Sigma}$ is assumed, no diagonal terms associated to noise are specified in the normalized problems in column A. Moreover, the prime symbol used to denote generalized eigenvalues for SFA in \cref{sec:Type1solSFA} is dropped in order to simplify notation.

\begin{table}
\renewcommand{\arraystretch}{2}
\centering
\small
\begin{tabular}{ |c|c|c|c|c|} 
\hline
\rotatebox[origin=c]{90}{Types} & \rotatebox[origin=c]{90}{\hspace{0.2cm}Formulations\hspace{0.2cm}} & \rotatebox[origin=c]{90}{Variants} & A. General input & B. Markovian one-hot trajectory, \(T\to\infty\) \\[1ex]
\hline
\multirow{9}{*}{T1} 
  & \multirow{3}{*}{\makecell{No\\norm.}} 
    & SFA    & \(\textcolor{blue}{\dot{\mat{\Sigma}}}\mat{W}=\mat{\Sigma}\mat{W}\mat{\Lambda}\) & \((2\textcolor{blue}{\mat{L}_{\textup{dir}}}+\textcolor{red}{\vec{\pi \pi}^T})\mat{W}=(\mat{\Pi}-\textcolor{red}{\vec{\pi \pi}^T})\mat{W}\mat{\Lambda}\) \\
\cline{3-5}
  &   & \(\tau\)SFA & \(\textcolor{blue}{\mat{\Omega}_\tau}\mat{W}=\mat{\Sigma}\mat{W}\mat{\Lambda}\) & \((\mat{\Pi}\textcolor{blue}{(\mat{P}^\tau)_{\textup{add}}}-\textcolor{red}{\vec{\pi \pi}^T})\mat{W}=(\mat{\Pi}-\textcolor{red}{\vec{\pi \pi}^T})\mat{W}\mat{\Lambda}\) \\
\cline{3-5}
  &   & LFSFA  & \(\textcolor{blue}{\mat{\Psi}_\gamma}\mat{W}=\mat{\Sigma}\mat{W}\mat{\Lambda}\) & \((\mat{\Pi}\textcolor{blue}{\mat{M}_{\textup{add}}}-\textcolor{red}{\alpha\vec{\pi \pi}^T})\mat{W}=(\mat{\Pi}-\textcolor{red}{\vec{\pi \pi}^T})\mat{W}\mat{\Lambda}\) \\
\cline{2-5}
  & \multirow{3}{*}{\makecell{Sym.\\norm.}} 
    & SFA    & \(\mat{\Sigma}^{-\frac{1}{2}}\textcolor{blue}{\dot{\mat{\Sigma}}}(\mat{\Sigma}^{-\frac{1}{2}})^T\widetilde{\mat{W}}=\widetilde{\mat{W}}\mat{\Lambda}\) & \((\mat{\Pi}-\textcolor{red}{\vec{\pi \pi}^T}+\textcolor{Green}{\sigma^2\mat{\mathbbm{1}}})^{-\frac{1}{2}}(2\textcolor{blue}{\mat{L}_{\textup{dir}}}+\textcolor{red}{\vec{\pi \pi}^T})(\mat{\Pi}-\textcolor{red}{\vec{\pi \pi}^T}+\textcolor{Green}{\sigma^2\mat{\mathbbm{1}}})^{-\frac{1}{2}}\widetilde{\mat{W}}=\widetilde{\mat{W}}\mat{\Lambda}\) \\
\cline{3-5}
  &   & \(\tau\)SFA & \(\mat{\Sigma}^{-\frac{1}{2}}\textcolor{blue}{\mat{\Omega}_\tau}(\mat{\Sigma}^{-\frac{1}{2}})^T\widetilde{\mat{W}}=\widetilde{\mat{W}}\mat{\Lambda}\) & \((\mat{\Pi}-\textcolor{red}{\vec{\pi \pi}^T}+\textcolor{Green}{\sigma^2\mat{\mathbbm{1}}})^{-\frac{1}{2}}(\mat{\Pi}\textcolor{blue}{(\mat{P}^\tau)_{\textup{add}}}-\textcolor{red}{\vec{\pi \pi}^T})(\mat{\Pi}-\textcolor{red}{\vec{\pi \pi}^T}+\textcolor{Green}{\sigma^2\mat{\mathbbm{1}}})^{-\frac{1}{2}}\widetilde{\mat{W}}=\widetilde{\mat{W}}\mat{\Lambda}\) \\
\cline{3-5}
  &   & LFSFA  & \(\mat{\Sigma}^{-\frac{1}{2}}\textcolor{blue}{\mat{\Psi}_\gamma}(\mat{\Sigma}^{-\frac{1}{2}})^T\widetilde{\mat{W}}=\widetilde{\mat{W}}\mat{\Lambda}\) & \((\mat{\Pi}-\textcolor{red}{\vec{\pi \pi}^T}+\textcolor{Green}{\sigma^2\mat{\mathbbm{1}}})^{-\frac{1}{2}}(\mat{\Pi}\textcolor{blue}{\mat{M}_{\textup{add}}}-\textcolor{red}{\alpha\vec{\pi \pi}^T})(\mat{\Pi}-\textcolor{red}{\vec{\pi \pi}^T}+\textcolor{Green}{\sigma^2\mat{\mathbbm{1}}})^{-\frac{1}{2}}\widetilde{\mat{W}}=\widetilde{\mat{W}}\mat{\Lambda}\) \\
\cline{2-5}
  & \multirow{3}{*}{\makecell{Left\\norm.}} 
    & SFA    & \(\mat{\Sigma}^{-1}\textcolor{blue}{\dot{\mat{\Sigma}}}\mat{W}=\mat{W\Lambda}\) & \((\mat{\Pi}-\textcolor{red}{\vec{\pi \pi}^T}+\textcolor{Green}{\sigma^2\mat{\mathbbm{1}}})^{-1}(2\textcolor{blue}{\mat{L}_{\textup{dir}}}+\textcolor{red}{\vec{\pi \pi}^T})\mat{W}=\mat{W}\mat{\Lambda}\) \\
\cline{3-5}
  &   & \(\tau\)SFA & \(\mat{\Sigma}^{-1}\textcolor{blue}{\mat{\Omega}_\tau}\mat{W}=\mat{W\Lambda}\) & \((\mat{\Pi}-\textcolor{red}{\vec{\pi \pi}^T}+\textcolor{Green}{\sigma^2\mat{\mathbbm{1}}})^{-1}(\mat{\Pi}\textcolor{blue}{(\mat{P}^\tau)_{\textup{add}}}-\textcolor{red}{\vec{\pi \pi}^T})\mat{W}=\mat{W}\mat{\Lambda}\) \\
\cline{3-5}
  &   & LFSFA  & \(\mat{\Sigma}^{-1}\textcolor{blue}{\mat{\Psi}_\gamma}\mat{W}=\mat{W\Lambda}\) & \((\mat{\Pi}-\textcolor{red}{\vec{\pi \pi}^T}+\textcolor{Green}{\sigma^2\mat{\mathbbm{1}}})^{-1}(\mat{\Pi}\textcolor{blue}{\mat{M}_{\textup{add}}}-\textcolor{red}{\alpha\vec{\pi \pi}^T})\mat{W}=\mat{W}\mat{\Lambda}\) \\
\hline
\multirow{9}{*}{T2} 
  & \multirow{3}{*}{\makecell{No\\norm.}} 
    & SFA    & \(\textcolor{red}{\widehat{\textcolor{blue}{\dot{\mat{\Sigma}}}}}\mat{W}=\textcolor{red}{\widehat{\textcolor{black}{\mat{\Sigma}}}}\mat{W}\mat{\Lambda}\) & \(2\textcolor{blue}{\mat{L}_{\textup{dir}}}\mat{W}=\mat{\Pi}\mat{W}\mat{\Lambda}\) \\
\cline{3-5}
  &   & \(\tau\)SFA & \(\textcolor{red}{\widehat{\textcolor{blue}{\mat{\Omega}}}_{\textcolor{blue}{\tau}}}\mat{W}=\textcolor{red}{\widehat{\textcolor{black}{\mat{\Sigma}}}}\mat{W}\mat{\Lambda}\) & \(\mat{\Pi}\textcolor{blue}{(\mat{P}^\tau)_{\textup{add}}}\mat{W}=\mat{\Pi}\mat{W}\mat{\Lambda}\) \\
\cline{3-5}
  &   & LFSFA  & \(\textcolor{red}{\widehat{\textcolor{blue}{\mat{\Psi}}}}_{\textcolor{blue}{\gamma}}\mat{W}=\textcolor{red}{\widehat{\textcolor{black}{\mat{\Sigma}}}}\mat{W}\mat{\Lambda}\) & \(\mat{\Pi}\textcolor{blue}{\mat{M}_{\textup{add}}}\mat{W}=\mat{\Pi}\mat{W}\mat{\Lambda}\) \\
\cline{2-5}
  & \multirow{3}{*}{\makecell{Sym.\\norm.}} 
    & SFA    & \(\textcolor{red}{\widehat{\textcolor{black}{\mat{\Sigma}}}}^{-\frac{1}{2}}\textcolor{red}{\widehat{\textcolor{blue}{\dot{\mat{\Sigma}}}}}(\textcolor{red}{\widehat{\textcolor{black}{\mat{\Sigma}}}}^{-\frac{1}{2}})^T\widetilde{\mat{W}}=\widetilde{\mat{W}}\mat{\Lambda}\) & \(2\mat{\Pi}^{-\frac{1}{2}}\textcolor{blue}{\mat{L}_{\textup{dir}}}\mat{\Pi}^{-\frac{1}{2}}\widetilde{\mat{W}}=\widetilde{\mat{W}}\mat{\Lambda}\) \\
\cline{3-5}
  &   & \(\tau\)SFA & \(\textcolor{red}{\widehat{\textcolor{black}{\mat{\Sigma}}}}^{-\frac{1}{2}}\textcolor{red}{\widehat{\textcolor{blue}{\mat{\Omega}}}_{\textcolor{black}{\tau}}}(\textcolor{red}{\widehat{\textcolor{black}{\mat{\Sigma}}}}^{-\frac{1}{2}})^T\widetilde{\mat{W}}=\widetilde{\mat{W}}\mat{\Lambda}\) & \(\mat{\Pi}^{\frac{1}{2}}\textcolor{blue}{(\mat{P}^\tau)_{\textup{add}}}\mat{\Pi}^{-\frac{1}{2}}\widetilde{\mat{W}}=\widetilde{\mat{W}}\mat{\Lambda}\) \\
\cline{3-5}
  &   & LFSFA  & \(\textcolor{red}{\widehat{\textcolor{black}{\mat{\Sigma}}}}^{-\frac{1}{2}}\textcolor{red}{\widehat{\textcolor{blue}{\mat{\Psi}}}}_{\textcolor{blue}{\gamma}}(\textcolor{red}{\widehat{\textcolor{black}{\mat{\Sigma}}}}^{-\frac{1}{2}})^T\widetilde{\mat{W}}=\widetilde{\mat{W}}\mat{\Lambda}\) & \(\mat{\Pi}^{\frac{1}{2}}\textcolor{blue}{\mat{M}_{\textup{add}}}\mat{\Pi}^{-\frac{1}{2}}\widetilde{\mat{W}}=\widetilde{\mat{W}}\mat{\Lambda}\) \\
\cline{2-5}
  & \multirow{3}{*}{\makecell{Left\\norm.}} 
    & SFA    & \(\textcolor{red}{\widehat{\textcolor{black}{\mat{\Sigma}}}}^{-1}\textcolor{red}{\widehat{\textcolor{blue}{\dot{\mat{\Sigma}}}}}\mat{W}=\mat{W\Lambda}\) & \(2\mat{\Pi}^{-1}\textcolor{blue}{\mat{L}_{\textup{dir}}}\mat{W}=\mat{W}\mat{\Lambda}\) \\
\cline{3-5}
  &   & \(\tau\)SFA & \(\textcolor{red}{\widehat{\textcolor{black}{\mat{\Sigma}}}}^{-1}\textcolor{red}{\widehat{\textcolor{blue}{\mat{\Omega}}}_{\textcolor{blue}{\tau}}}\mat{W}=\mat{W\Lambda}\) & \(\textcolor{blue}{(\mat{P}^\tau)_{\textup{add}}}\mat{W}=\mat{W}\mat{\Lambda}\) \\
\cline{3-5}
  &   & LFSFA  & \(\textcolor{red}{\widehat{\textcolor{black}{\mat{\Sigma}}}}^{-1}\textcolor{red}{\widehat{\textcolor{blue}{\mat{\Psi}}}}_{\textcolor{blue}{\gamma}}\mat{W}=\mat{W\Lambda}\) & \(\textcolor{blue}{\mat{M}_{\textup{add}}}\mat{W}=\mat{W}\mat{\Lambda}\) \\
\hline
\end{tabular}
\caption{The generalized and regular eigenvalue problems corresponding to all types, formulations, and variants of SFA considered in this paper. In column A, the problems are shown for a general input $\vec{x}(t)$, as introduced in \cref{sec:Type1sol}, with the matrices that correspond to the objectives (i.e.\ $\dot{\mat{\Sigma}}$, $\widehat{\dot{\mat{\Sigma}}}$, $\mat{\Omega}_\tau$, $\widehat{\mat{\Omega}}_\tau$, $\mat{\Psi}_\gamma$, and $\widehat{\mat{\Psi}}_\gamma$) indicated in blue and with all hats indicated in red in order to emphasize the distinction between Type 1 and Type 2 versions of each SFA problem. In column B, each problem is shown for the special case where $\vec{x}(t)$ is a Markovian one-hot trajectory, assuming the limit $T\to\infty$, as introduced in \cref{sec:Results}. Moreover, a similar color scheme is used as in column A. For example, the basic graph theoretic quantities associated to the objectives in this limit (i.e.\ $\mat{L}_{\textup{dir}}$, $(\mat{P}^\tau)_{\textup{add}}$, and $\mat{M}_{\textup{add}}$) are highlighted in blue, and the quantities highlighted in red are the ones which differ between Type 1 and Type 2 versions of each SFA problem. Moreover, in column B, the additional terms $\sigma^2\mat{\mathbbm{1}}$ that occur whenever it is necessary to add noise, i.e.\ in the Type 1 normalized formulations, are shown in green.}
\label{tab:AllEigenvalueProblems}
\end{table}

Two details are worth emphasizing about the material of this section. Firstly, while a left normalized formulation of SFA has been used in at least one study \citep{Creutzig2008},\footnote{Note that \citet{Creutzig2008} describe SFA as a generalized eigenvalue problem involving left eigenvectors, which is normalized by multiplying with $\mat{\Sigma}^{-1}$ from the right rather than from the left. However, provided that the problem is instead formulated using right eigenvectors, like in the current paper, then this multiplication happens from the left, like in \cref{eq:T1EVleft}.} symmetric normalized formulations are more common since for Type 1 SFA problems they correspond to the use of whitening in the data preprocessing phase (see Appendix \ref{app:SymNormWhite}), which is a common method for enforcing the constraints in all Type 1 SFA problems considered so far \citep{Ziehe1998,Wiskott2002,Blaschke2006,Wang2020}. Secondly, while the eigenvectors in $\widetilde{\mat{W}}$ are orthogonal w.r.t.\ to the Euclidean inner product, those in $\mat{W}$ are orthogonal w.r.t.\ to the weighted inner product $(\cdot,\cdot)_{\mat{\Sigma}}$ (see \cref{sec:Type1solSFA} and Appendix \ref{app:GenEVprob}). In practice, most eigenvalue solvers choose vectors with Euclidean norm $1$ by default, in which case the eigenvectors in $\mat{W}$ must be rescaled by $\vec{w}_i\to \frac{\vec{w}_i}{|\vec{w}_i|_{\mat{\Sigma}}}$, so that $|\vec{w}_i|_{\mat{\Sigma}}=1$, which ensures that the unit variance constraint is met.

\subsection{Solving Type 2 SFA Problems}
\label{sec:Type2sol}
For Type 2 SFA problems, the solutions can similarly be described in terms of generalized eigenvalue problems, with the only difference being that the zero mean constraint is no longer applied. For each of the Type~1 SFA problems considered in the previous section, this constraint is enforced through the centering transformation $\vec{x}(t)\to\vec{c}(t)$ \citep{Berkes2005,Creutzig2008}. Therefore, the solutions to the corresponding Type 2 SFA problems can be computed by omitting this transformation and leaving all other steps the same. The two main differences that result from leaving out the centering step are the following. Firstly, the input-output functions are linear, in comparison to the Type 1 case where they are affine (see Equations \ref{eq:SFAaffine1}-\ref{eq:SFAaffine4}). Secondly, replacing the vector $\vec{c}(t)$ with $\vec{x}(t)$ in all problems considered so far leads to a set of generalized eigenvalue problems involving new matrices. In order to emphasize the correspondence of these new matrices with those of the last section, they are denoted using the same symbol but with an additional hat. For example, replacing $\vec{c}(t)$ for $\vec{x}(t)$ in $\mat{\Sigma}$ and $\dot{\mat{\Sigma}}$ gives
\begin{align}
\widehat{\mat{\Sigma}}&=\langle \vec{x}(t)\vec{x}(t)^T\rangle_t\\
\widehat{\dot{\mat{\Sigma}}}&=\langle \dot{\vec{x}}(t)\dot{\vec{x}}(t)^T\rangle_t
\end{align}
which are the \textit{second moment matrices} of $\vec{x}(t)$ and $\dot{\vec{x}}(t)$, respectively. Doing the same for $\mat{\Omega}_\tau$ and $\mat{\Psi}_\gamma$ gives
\begin{align}
\widehat{\mat{\Omega}}_\tau&=\frac{1}{2}\big(\langle\vec{x}(t)\vec{x}(t+\tau)^T\rangle_t+\langle\vec{x}(t+\tau)\vec{x}(t)^T\rangle_t\big)\\
\widehat{\mat{\Psi}}_\gamma&=\sum_{\tau=0}^{\tau_{\text{max}}}\frac{\gamma^\tau}{2}\big(\langle\vec{x}(t)\vec{x}(t+\tau)^T\rangle_t+\langle\vec{x}(t+\tau)\vec{x}(t)^T\rangle_t\big)
\end{align}
which for simplicity are not given new names. Like in the previous section, $\widehat{\dot{\mat{\Sigma}}}$, $\widehat{\mat{\Omega}}_\tau$, and $\widehat{\mat{\Psi}}_\gamma$ arise from the objectives of Type 2 SFA, $\tau$SFA, and LFSFA, respectively, while $\widehat{\mat{\Sigma}}$ arises from the constraints of these problems. Moreover, in analogy to the Type 1 case the unnormalized formulation of these problems are given by
\begin{equation}
\mat{AW}=\widehat{\mat{\Sigma}}\mat{W\Lambda}\label{eq:T2GenEV}
\end{equation}
where for $\mat{A}=\widehat{\dot{\mat{\Sigma}}}$ it is the generalized eigenvectors with smallest generalized eigenvalues that are optimal, while for $\mat{A}=\widehat{\mat{\Omega}}_\tau$ or $\widehat{\mat{\Psi}}_\gamma$ it is those with largest generalized eigenvalues.

It should be emphasized that the analytical comparisons drawn between Type 1 and Type 2 SFA problems in \cref{sec:T2SFA} apply equally to their associated generalized eigenvalue problems. For example, Type 2 SFA problems come with the additional possibility of a constant output, which corresponds to a generalized eigenvector of $\widehat{\dot{\mat{\Sigma}}}$ and $\widehat{\mat{\Omega}}_\tau$ with generalized eigenvalue $0$ and $1$, respectively. Therefore, in contrast to the Type 1 case, the eigenvalue intervals of these matrices are $[0,4]$ and $[-1,1]$, respectively. Moreover, when a constant output does occur, all other outputs correspond to solutions from the Type 1 setting.

Furthermore, many of the details regarding normalization of Type 1 problems carry over to the Type 2 case. For example, if $\widehat{\mat{\Sigma}}$ is positive definite then the symmetric and left normalized formulations are given by
\begin{equation}
\widehat{\mat{\Sigma}}^{-\frac{1}{2}}\mat{A}(\widehat{\mat{\Sigma}}^{-\frac{1}{2}})^T\widetilde{\mat{W}}=\widetilde{\mat{W}}\mat{\Lambda}
\end{equation}
and
\begin{equation}
\widehat{\mat{\Sigma}}^{-1}\mat{A}\mat{W}=\mat{W\Lambda}
\end{equation}
respectively, and if instead $\widehat{\mat{\Sigma}}$ has one or more eigenvalues equal to zero then it is necessary to either add noise or compute the pseudoinverse (see \cref{sec:Type1Norm}). Moreover, the addition of Gaussian noise affects $\widehat{\mat{\Sigma}}$ in the same way as $\mat{\Sigma}$ in the large data limit, i.e.\ $\widehat{\mat{\Sigma}}\to\widehat{\mat{\Sigma}}+\sigma^2\mat{\mathbbm{1}}$ (see Appendix \ref{app:LinSFAnoise}), and this is how zero eigenvalues are dealt with in \cref{sec:Results} when necessary. The only difference in the case of Type~2 problems is that the symmetric normalized formulations are not related to whitening, since whitening is formally defined only for data that has zero mean (see Appendix \ref{app:SymNormWhite}). Instead, applying symmetric normalization to Type~2 SFA problems corresponds to a process whereby the second moments of the input data are normalized to unity, and it should be noted that the definition of this normalization process has the same degree of freedom as explored for whitening in Appendix \ref{app:SymNormWhite}.

All formulations of each Type 2 SFA problem are shown in the lower half of column A in \cref{tab:AllEigenvalueProblems}, and in analogy to the Type 1 Problems shown in this column a general input $\vec{x}(t)$ without zero eigenvalues in $\widehat{\mat{\Sigma}}$ is assumed.

\subsection{Representation of the input signal}
\label{sec:RepInput}
One context in which it is insightful to study SFA is where the input signal $\vec{x}(t)\in\mathbb{R}^N$ is generated by a lower dimensional set of source signals or latent variables, i.e.\ $\vec{u}(t)=[u_1(t), u_2(t), ..., u_n(t)]\in\mathbb{R}^n$ with $n\ll N$, through some non-linear function, i.e.\ $f:\mathbb{R}^n\mapsto\mathbb{R}^N$, $f(\vec{u}(t))=\vec{x}(t)$. In this setting, which is conceptually related to the problem of non-linear blind source separation \citep{Sprekeler2014}, it is possible to show that the optimal SFA solutions are independent of the function $f$ and can be formulated purely in terms of the source signal $\vec{u}(t)$ \citep{Franzius2007a, Sprekeler2008, Sprekeler2009,Sprekeler2014}. In order for this argument to hold, two assumptions are necessary. Firstly, the function $f$ that maps $\vec{u}(t)$ to $\vec{x}(t)$ must be injective, or in other words one-to-one. Note that when $f$, $\vec{u}(t)$, and $\vec{x}(t)$ are all continuous, this implies that the input data is embedded on an $n$-dimensional manifold in $\mathbb{R}^N$ that is parametrized by $\vec{u}$. Secondly, the SFA function space $\mathcal{F}$ must be unrestricted. Together, these two assumptions imply that any function of the input signal $\vec{x}(t)$ can equivalently be described as a function of the corresponding source signal $\vec{u}(t)$, i.e.\
\begin{equation}
\label{eq:SFAoutputsourceinput}
\tilde{g}_j(\vec{x}(t))=\tilde{g}_j(f(\vec{u}(t)))=g_j(\vec{u}(t))\quad\text{where}\quad\tilde{g}_j,g_j\in\mathcal{F}
\end{equation}
where $g_j=\tilde{g}_j\circ f$. In words, \cref{eq:SFAoutputsourceinput} says that while the optimal functions depend on the choice of coordinate system, the optimal outputs do not. Thus, under both assumptions, the SFA problem can equivalently be formulated in terms of the sources $\vec{u}(t)$ \citep[see][Optimization Problem 2]{Franzius2007a}. The key utility of this result is that the application of variational calculus to the SFA problem is typically more tractable when formulated in terms of $\vec{u}(t)$, since (i) by assumption, $\vec{u}(t)$ has much lower dimensionality than $\vec{x}(t)$, and (ii) it is often more mathematically convenient to work in the space of sources signals $\vec{u}(t)$. By applying variational calculus in this way, one arrives at a partial differential equation (PDE) in terms of $\vec{u}(t)$ and $\dot{\vec{u}}(t)$ that describe the optimal SFA solutions \citep{Franzius2007a, Sprekeler2008, Sprekeler2009,Sprekeler2014}. It is worth noting that since it is more natural to omit the zero mean constraint in the variational calculus formalism, the analytical solutions studied in these papers formally correspond to Type 2 SFA. However, since in each study the constant signal is ignored and an unrestricted function space is assumed, the solutions are equivalent to those of Type 1 SFA (see \cref{sec:SFAdef}).

Clearly, an unrestricted function space is in practice not realizable, which means that the argument given above formally applies to an abstracted version of the SFA problem. However, it is possible to achieve something close to an unrestricted function space by performing SFA with a highly expressive function space, such as that realised by an hSFA network, in which case the PDE described above applies to a good degree of approximation \citep{Franzius2007a}.

A further simplification studied in \citep{Franzius2007a, Sprekeler2008, Sprekeler2009,Sprekeler2014} is the case where each of the source signals $u_1(t)$, $u_2(t)$, ..., $u_n(t)$ are statistically independent. Under this assumption, the optimal SFA functions factorize into products of functions defined over each individual source $u_\alpha(t)$, i.e.
\begin{equation}
g_j(\vec{u}(t))=\prod_{\alpha}g_{j\alpha}(u_\alpha(t))\label{eq:StatIndepSource}
\end{equation}
where $g_{j\alpha}$ can be thought of as a harmonic function of the source $u_\alpha(t)$ with a frequency that increases with the index $j$ \citep{Sprekeler2008,Sprekeler2014}. Moreover, if Type 2 SFA is considered, then the $j=0$ harmonic of each source is guaranteed to be a constant. Therefore, setting $j=0$ for all but one source $u_\alpha(t)$ in \cref{eq:StatIndepSource} shows that the corresponding single source harmonics, i.e.\  $g_{j\alpha}$ $\forall j$, are also solutions to the full problem \citep{Sprekeler2008,Sprekeler2014}.

\subsection{SFA in spatial environments}
\label{sec:SFAspatial}
A particularly relevant application of the theory from the previous section is the case where $\vec{x}(t)$ is generated from sensors monitoring an agent moving around in a spatial environment. In this setting, a plausible set of source signals $\vec{u}(t)$ are the variables that spatially describe the agent's state in the environment. For example, in the model of \citet{Franzius2007a}, an agent explores a virtual open-field environment with dimensions $L_x\times L_y$ and with a visual data stream as input that is determined by the agent's position and head direction, i.e.\ $\vec{u}=[x,y,\varphi]$. For this model, the mapping between $\vec{x}(t)$ and $\vec{u}(t)$ is injective\footnote{In particular, the environment studied is sufficiently rich such that each input image $\vec{x}$ corresponds to a unique position and head direction, and the environment does not change over time such that each position and head direction corresponds to a unique image $\vec{x}$.} and an hSFA network is used, which offers a degree of flexibility that approaches the abstract setting of an unrestricted function space $\mathcal{F}$. Thus, to a good degree of approximation the optimal SFA solutions in this context are analytically described by a PDE involving the sources $\vec{u}(t)$ and the corresponding time derivatives or velocities $\dot{\vec{u}}=[v_x, v_y, \omega]$. In \citep{Franzius2007a}, this PDE is further simplified by assuming that the positions and orientations are homogeneously distributed over time and that the velocities are decorrelated, which the authors referred to as the \textit{simple movement paradigm}. Since this movement paradigm satisfies the condition of statistical independence discussed in the last section, the optimal SFA solutions factorize into products of functions over $x$, $y$, and $\varphi$, and are given by
\begin{equation}
g_{jln}(x,y,\varphi)=\left\{
	\begin{array}{ll}
		\sqrt{8}\;\text{cos}\bigg(j\pi\frac{x}{L_x}\bigg)\text{cos}\bigg(l\pi\frac{y}{L_y}\bigg)\text{sin}\bigg(\frac{n+1}{2}\varphi\bigg) & \mbox{for } n \mbox{ odd}\\[1em]
		\sqrt{8}\;\text{cos}\bigg(j\pi\frac{x}{L_x}\bigg)\text{cos}\bigg(l\pi\frac{y}{L_y}\bigg)\text{sin}\bigg(\frac{n}{2}\varphi\bigg) & \mbox{for } n \mbox{ even}
	\end{array}
\right.\label{eq:SimpleMovementxytheta}
\end{equation}
and the associated $\Delta$-values by
\begin{equation}
\Delta_{jln}=\left\{
	\begin{array}{ll}
		\pi^2\langle v^2\rangle\bigg(\frac{j^2}{L_x^2}+\frac{l^2}{L_y^2}\bigg)+\langle w^2\rangle\frac{(n+1)^2}{4} & \mbox{for } n \mbox{ odd}\\[1em]
		\pi^2\langle v^2\rangle\bigg(\frac{j^2}{L_x^2}+\frac{l^2}{L_y^2}\bigg)+\langle w^2\rangle\frac{n^2}{4} & \mbox{for } n \mbox{ even}
	\end{array}
\right.\label{eq:SimpleMovementxythetaDelta}
\end{equation}
where $j,l,n=0,1,2,...$ and the solution corresponding to $j=l=n=0$ is discarded. Note that since these solutions are derived in terms of the sources rather than the inputs, the sampling rate of the input data is not taken into account and so the $\Delta$-values are not restricted to the domain $[0,4]$ like in the discrete time setting. Instead, they increase without bound like in the case of continuous time free responses \citep[see \cref{sec:SFAfree} and][]{Wiskott2003}.

\begin{figure}[h]
\centering
\includegraphics[width=0.8\textwidth]{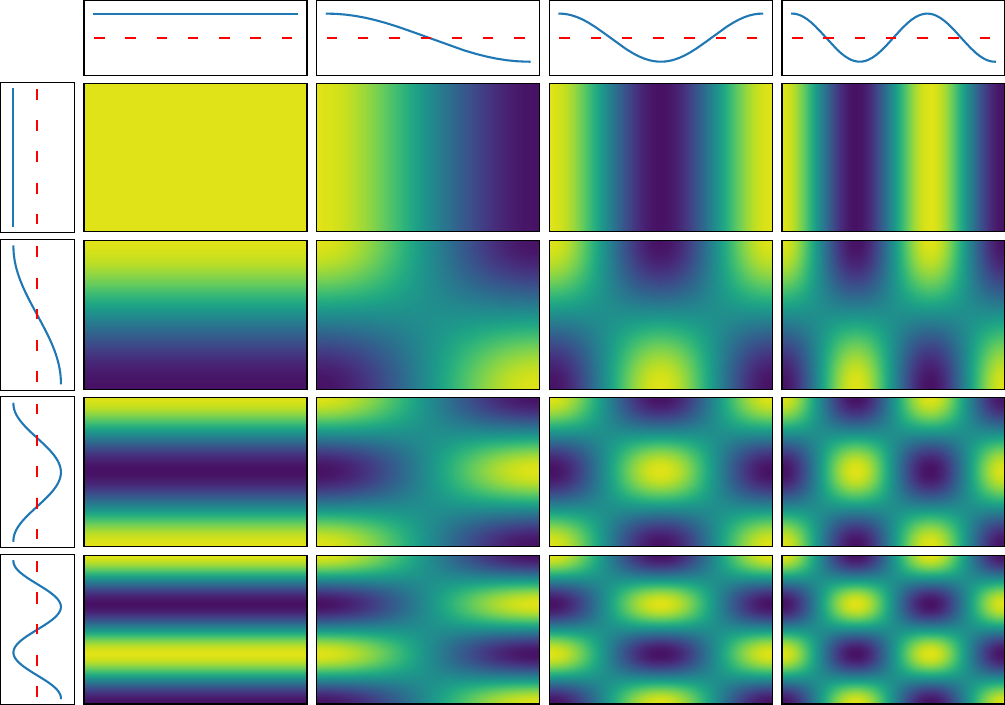}
 \caption{An illustration of the solutions described by \Crefrange{eq:SimpleMovementxy1}{eq:SimpleMovementxy3} for a rectangular open field with $\frac{L_x}{L_y}=1.5$ and for $j,l=0,1,2,3$ ordered from left to right and top to bottom, respectively.}
\label{fig:Spatial2Dcont}
\end{figure}

Using an analogous derivation, it is possible to study the optimal solutions for this movement paradigm but where the source signals only involve location, i.e.\ $\vec{u}=[x,y]$, which is more relevant to the experiments performed in \cref{sec:Results}. In this case, the optimal SFA solutions and associated $\Delta$-values are given by
\begin{align}
g_{jl}(x,y)&=\sqrt{4}\;\text{cos}\bigg(j\pi\frac{x}{L_x}\bigg)\text{cos}\bigg(l\pi\frac{y}{L_y}\bigg)\label{eq:SimpleMovementxy1}\\
&=\underbrace{\sqrt{2}\;\text{cos}\bigg(j\pi\frac{x}{L_x}\bigg)}_{=g_j(x)}\underbrace{\sqrt{2}\;\text{cos}\bigg(l\pi\frac{y}{L_y}\bigg)}_{=g_l(y)}\label{eq:SimpleMovementxy2}\\
&=g_j(x)g_l(y)\label{eq:SimpleMovementxy3}
\end{align}
and
\begin{equation}
\Delta_{jl}=\pi^2\langle v^2\rangle\bigg(\frac{j^2}{L_x^2}+\frac{l^2}{L_y^2}\bigg)\label{eq:SimpleMovementxyDelta}
\end{equation}
respectively. \Crefrange{eq:SimpleMovementxy1}{eq:SimpleMovementxy3} describe a set of 2D standing waves defined over a rectangular open field environment that decompose into a product of two 1D standing waves $g_j(x)$ and $g_l(y)$ defined over the horizontal and vertical axes of the environment, respectively. Note that Type 2 SFA was considered in \citep{Franzius2007a}, which means that $g_j(x)$ and $g_l(y)$ are also solutions to the full problem, as explained in the previous section. Moreover, it should be acknowledged that $g_j(x)$ and $g_l(y)$ are equivalent to the continuous time free responses in \cref{eq:SFAFreeCont}, provided that the relative positions $\frac{x}{L_x}$ or $\frac{y}{L_y}$ are replaced by a relative time $\frac{t-t_A}{t_B-t_A}$, respectively. \cref{eq:SimpleMovementxyDelta} says that the $\Delta$-value of each output $g_{jl}(x,y)$ is a sum of two terms, each associated with either $g_j(x)$ or $g_l(y)$. Furthermore, each of these terms is equivalent to the $\Delta$-values found by \citet{Wiskott2003} for the continuous free responses provided that relative position is swapped for relative time and $\langle v^2\rangle$ is set to $1$. The grid of plots in \cref{fig:Spatial2Dcont} illustrates the solutions described by \Crefrange{eq:SimpleMovementxy1}{eq:SimpleMovementxy3} for $j,l=0,1,2,3$ and $\frac{L_x}{L_y}=1.5$. The first row and column in this grid depict $g_j(x)$ and $g_l(y)$ as line plots and the remaining plots depict the corresponding products $g_{j,l}(x,y)$ as heat maps over the open field.

Plotting SFA outputs in terms of underlying source variables, i.e.\ $y_j(\vec{u})$, can be particularly insightful when source variables are spatial. This is true not only for the optimal outputs predicted by theory, such as in \cref{fig:Spatial2Dcont}, but also for real outputs of an SFA computation. In the latter case, once a set of SFA weight vectors have been generated, it is necessary to perform a post-processing phase in which the space of source signals is explored and the corresponding outputs $y_j(\vec{u})=g_j(\vec{u})$ are computed \citep[as in][]{Franzius2007a}. Moreover, while the theory of the current and previous sections applies only to SFA, it is possible to visualize the outputs of CSFA, $\tau$SFA and LFSFA in an equivalent manner when the underlying environment is spatial, which is done in \cref{sec:Results}.

\section{Results}
\label{sec:Results}

This section presents the main results of this paper, which identify analytical and conceptual relationships between SR and all variants, formulations, and types of the SFA problem introduced in \cref{sec:SFA}. \cref{sec:OneHot} introduces the time series $\vec{x}(t)$ for which these relationships are best explored and  \cref{sec:LimRes} presents four theorems that describe the particular form that the SFA problems take in the limit $T\to\infty$ for this type of time series. Then, \cref{sec:Exp} explores these theorems empirically by considering a particular type of environment and policy.

\subsection{Markovian one-hot trajectories}
\label{sec:OneHot}
Since SFA and SR are defined in different contexts of machine learning, making a formal comparison between the two requires adapting one to the setting of the other. One fact that makes such an adaptation straightforward is that both are based on the notion of a time series. In the case of SFA, this is trivially true since it is a time series method, while for SR it stems from the fact that RL is the study of temporal decision making processes. One key difference between the two is that in SFA the input can essentially be any time series, while SR has two crucial restrictions. Firstly, SR is defined in terms of state trajectories that occur in a finite MDP, for which the temporal statistics are guaranteed to be Markovian. Secondly, in finite MDPs it is typical to represent state occupation using a one-hot encoding,\footnote{Some extensions of SR beyond this encoding and to continuous state spaces have been formulated using tools from deep learning \citep[for a review, see][]{Carvalho2024b}} i.e.\
\begin{equation}
x_i(t)=\left\{
	\begin{array}{ll}
		1 & \mbox{if } s_i \mbox{ is occupied}\\
		0 & \mbox{otherwise }
	\end{array}
\right.
\end{equation}
Therefore, the time series underlying SR, henceforth referred to as \textit{Markovian one-hot trajectories}, are considerably less general than those considered in SFA, in which neither Markovian statistics nor one-hot representations are necessary.

\begin{figure}[h]
\centering
\includegraphics[width=0.95\textwidth]{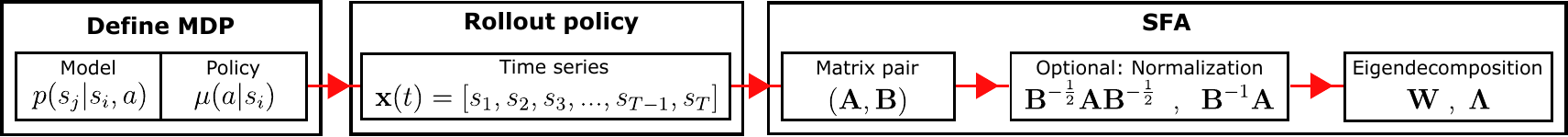}
 \caption{A data pipeline for applying SFA, $\tau$SFA, or LFSFA to Markovian one-hot trajectories sampled from an MDP.}
\label{fig:MDP-SFA}
\end{figure}

The above insights suggest that one possible way to adapt SFA to the setting of SR is to constrain the input of SFA by assuming that it is a Markovian one-hot trajectory resulting from an RL agent exploring its environment using some policy $\mu$. The general pipeline corresponding to this is illustrated in \cref{fig:MDP-SFA}, where the model and policy are initially defined, before rolling out the policy to obtain the input time series $\vec{x}(t)$, and then solving one of the SFA problems presented in \cref{sec:SFA} for this input. In the last step of \cref{fig:MDP-SFA}, each unnormalized formulation is denoted by an ordered pair of matrices $(\mat{A}, \mat{B})$,\footnote{In the literature, this is sometimes referred to as the \textit{matrix pencil} underlying a generalized eigenvalue problem \citep{Parlett1991,Ikramov1993}.} where the Type 1 versions involve $\mat{A}\in\{\dot{\mat{\Sigma}}, \mat{\Omega}_\tau,\mat{\Psi}_\gamma\}$ and $\mat{B}=\mat{\Sigma}$, while the Type 2 versions involve $\mat{A}\in\{\widehat{\dot{\mat{\Sigma}}}, \widehat{\mat{\Omega}}_\tau,\widehat{\mat{\Psi}}_\gamma\}$ and $\mat{B}=\widehat{\mat{\Sigma}}$. Moreover, as described in \cref{sec:Type1Norm,sec:Type2sol}, each of these problems can optionally be converted to the corresponding symmetric or left normalized formulations, i.e.\ the regular eigenvalue problem associated to $\mat{B}^{-\frac{1}{2}}\mat{A}\mat{B}^{-\frac{1}{2}}$ or $\mat{B}^{-1}\mat{A}$, respectively. In order for the normalization methods to be well-defined, Gaussian noise with a suitable variance is added to $\vec{x}(t)$ whenever $\mat{\Sigma}$ or $\widehat{\mat{\Sigma}}$ have one or more eigenvalues equal to zero (see \cref{sec:Type1Norm} and \cref{sec:Type2sol}). One detail worth emphasizing about this application of SFA is that for a one-hot encoding it is always possible to produce a constant output. Namely, if $\vec{w}=\vec{1}$ is a constant vector of ones, then the output at each time point is always $1$, i.e. $y(t)=\vec{1}^T\vec{x}(t)=1$. Therefore, for this time series the Type 1 and Type 2 solutions are the same except for the additional constant output in the latter case (see \cref{sec:SFAdef}).

\subsection{Limit theorems}
\label{sec:LimRes}

This section explores the specific form that the SFA problems of \cref{sec:SFA} take in the case of a Markovian one-hot trajectory by studying the matrices involved in these problems in the limit $T\to\infty$. \cref{sec:LimResType1,sec:LimResType2} focus on Type 1 and Type 2 SFA problems, respectively. Note that like in \cref{sec:SR}, the Markov chain underlying the input data is assumed to be ergodic, which plays an important role in the proofs of the results that are presented. Note also that although the results of this section are motivated by connections between SFA and SR, they in principle apply to any use of SFA as a state representation method for finite MDPs, provided that $\vec{x}(t)$ satisfies the conditions outlined in \cref{sec:OneHot} and that ergodicity holds.

\subsubsection{Type 1 SFA Problems}
\label{sec:LimResType1}

Type 1 SFA problems are equivalent to eigenvalue problems involving the matrices $\mat{\Sigma}$, $\dot{\mat{\Sigma}}$, $\mat{\Omega}_\tau$, and $\mat{\Psi}_\gamma$ (see \cref{tab:AllEigenvalueProblems}). While for a general time series $\vec{x}(t)$ there is no rule about what the exact entries of these matrices are, for a Markovian one-hot trajectory they are equal to values associated to the underlying Markov chain in the limit $T\to\infty$, as outlined by the following two results:

\begin{theorem}
\label{thm:Type1LimConstraint}
For an ergodic Markovian one-hot trajectory $\vec{x}(t)$, the matrix $\mat{\Sigma}$ associated to the Type 1 constraints obeys the following limit:
\begin{align}
\lim_{T\to\infty}\mat{\Sigma}&=\mat{\Pi}-\vec{\pi \pi}^T
\end{align}
where $\vec{\pi}$ is the unique stationary distribution of the underlying Markov chain and $\mat{\Pi}=\textup{diag}(\vec{\pi})$. Moreover, the limiting matrix $\mat{\Pi}-\vec{\pi \pi}^T$ is guaranteed to have a single eigenvalue of $0$ (proof: see Appendix \ref{app:LimType1}).
\end{theorem}

\begin{theorem}
\label{thm:Type1LimObjective}
For an ergodic Markovian one-hot trajectory $\vec{x}(t)$, the matrices $\dot{\mat{\Sigma}}$, $\mat{\Omega}_\tau$, and $\mat{\Psi}_\gamma$ associated to the Type 1 objectives obey the following limits:
\begin{align}
\lim_{T\to\infty}\dot{\mat{\Sigma}}&=2\mat{L}_{\textup{dir}}+\vec{\pi \pi}^T\\
\lim_{T\to\infty}\mat{\Omega}_\tau&=\mat{\Pi}(\mat{P}^\tau)_{\textup{add}}-\vec{\pi \pi}^T\\
\lim_{T\to\infty}\mat{\Psi}_\gamma&=\mat{\Pi}\mat{M}_{\textup{add}}-\alpha\vec{\pi}\vec{\pi}^T
\end{align}
where $\mat{L}_{\textup{dir}}$ is the combinatorial directed Laplacian matrix associated to the underlying Markov chain, $\mat{P}$ is the transition matrix, $\mat{M}$ is the SR matrix with discount factor $\gamma$, and $\alpha=\frac{1-\gamma^{\tau_{\textup{max}}+1}}{1-\gamma}\in[1,\frac{1}{1-\gamma})$ is a scalar constant (proof: see Appendix \ref{app:LimType1}).
\end{theorem}
\noindent The main insight from \cref{thm:Type1LimConstraint,thm:Type1LimObjective} is that for this type of input in the limit $T\to\infty$ the Type~1 generalized eigenvalue problems are given by:
\begin{align}
(2\mat{L}_{\textup{dir}}+\vec{\pi \pi}^T)\mat{W}&=(\mat{\Pi}-\vec{\pi \pi}^T)\mat{W}\mat{\Lambda}&\text{(SFA)}&\label{eq:GenEVType1SFA}\\
(\mat{\Pi}(\mat{P}^\tau)_{\textup{add}}-\vec{\pi \pi}^T)\mat{W}&=(\mat{\Pi}-\vec{\pi \pi}^T)\mat{W}\mat{\Lambda}& \text{($\tau$SFA)}&\label{eq:GenEVType1tauSFA}\\
(\mat{\Pi}\mat{M}_{\textup{add}}-\alpha\vec{\pi}\vec{\pi}^T)\mat{W}&=(\mat{\Pi}-\vec{\pi \pi}^T)\mat{W}\mat{\Lambda}& \text{(LFSFA)}&\label{eq:GenEVType1LFSFA}
\end{align}
\cref{eq:GenEVType1SFA} relates the SFA problem to the combinatorial directed graph Laplacian matrix $\mat{L}_{\textup{dir}}$ associated to the underlying Markov chain \citep[originally defined in][]{Chung2005}. \cref{eq:GenEVType1tauSFA} relates the $\tau$SFA problem to $\mat{\Pi}(\mat{P}^\tau)_{\textup{add}}$, which is the flow matrix associated to the additive reversibilization of the underlying $\tau$-step Markov chain (see \cref{sec:SReig}). Thus, the time scale used in $\tau$SFA extracts information relating to the $\tau$-step transition statistics of the underlying process. For $\tau=1$, \cref{eq:GenEVType1tauSFA} relates the CSFA problem to $\mat{\Pi}\mat{P}_{\textup{add}}$, and comparing this to \cref{eq:GenEVType1SFA} indicates that swapping time derivatives for time-lagged correlations corresponds to swapping $\mat{L}_{\textup{dir}}$ for $\mat{\Pi}\mat{P}_{\textup{add}}$, which is intuitive since in both settings the generalized eigenvectors remain the same and only the generalized eigenvalues change \citep[see \cref{sec:Type1solCSFA} and][]{Seabrook2023}. \cref{eq:GenEVType1LFSFA} relates the LFSFA problem to $\mat{\Pi}\mat{M}_{\textup{add}}$, which generalizes \cref{eq:GenEVType1tauSFA} across different time scales and applies temporal discounting. Note that the relation of SFA to graph Laplacians is consistent with both the study of \citet{Sprekeler2011} as well as unpublished work by Merlin Schüler (personal communication, 2021), but that the relations of $\tau$SFA and LFSFA to transition matrices and SR, respectively, are novel insights.

An important general feature of \Crefrange{eq:GenEVType1SFA}{eq:GenEVType1LFSFA} is that all matrices on the left-hand side involve symmetrization, which arises from the definitions of $\dot{\mat{\Sigma}}$, $\mat{\Omega}_\tau$, and $\mat{\Psi}_\gamma$. In \cref{eq:GenEVType1SFA}, $\mat{L}_{\textup{dir}}$ is formulated in terms of a directed graph, which is converted to an undirected graph through this symmetrization (see Appendix \ref{app:LimType1}). In \cref{eq:GenEVType1tauSFA,eq:GenEVType1LFSFA}, the symmetrization is related to the notion of additive reversibilization, which is consistent with the observation that $\mat{\Omega}_1$ involves an additive mixture of the forward and backward correlations (see the discussion in \cref{sec:Type1solCSFA}), as do $\mat{\Omega}_\tau$ and $\mat{\Psi}_\gamma$. Practically, the role of symmetrization in \Crefrange{eq:GenEVType1SFA}{eq:GenEVType1LFSFA} is crucial since it guarantees that all matrices on the left-hand side can be orthogonally diagonalized with real eigenvalues and eigenvectors (in the case of $\mat{\Pi}(\mat{P}^\tau)_{\textup{add}}$ and $\mat{\Pi}\mat{M}_{\textup{add}}$ these properties are proven in \cref{thm:P+Madd}). It should be emphasized that some studies on SFA introduce additional assumptions that circumvent the need for symmetrization, since this makes various analytical details more straightforward. For example, \citet{Sprekeler2011} considers only Laplacians defined on undirected graphs and \citet{Blaschke2006} and \citet{Creutzig2008} both assume that the temporal statistics of the input data are time reversible to begin with. Conceptually, this is equivalent to the assumption made in some SR studies that the underlying Markov chain is formulated as a random walk on an undirected graph \citep{Stachenfeld2014,Stachenfeld2017}. The absence of such assumptions in the limit results of this paper reflects the fact that SFA generally receives an input $\vec{x}(t)$ generated from a non-reversible process and produces features that are related to a symmetrized/reversibilized version of this process. To the authors' knowledge, the only other study in which this has been explored is unpublished work of Merlin Schüler (personal communication, 2021), which also relates SFA to $\mat{L}_{\textup{dir}}$.

As explained in \cref{sec:Type1Norm}, solving SFA problems typically involves preprocessing steps that transform the underlying generalized eigenvalue problem into a regular one. In particular, while each Type 1 generalized eigenvalue problem involves the ordered pair of matrices $(\mat{A},\mat{\Sigma})$ where $\mat{A}\in\{\dot{\mat{\Sigma}}, \mat{\Omega}_\tau,\mat{\Psi}_\gamma\}$, the corresponding symmetric and left normalized formulations are regular eigenvalue problems involving the single matrices $\mat{\Sigma}^{-\frac{1}{2}}\mat{A}(\mat{\Sigma}^{-\frac{1}{2}})^T$ and $\mat{\Sigma}^{-1}\mat{A}$, respectively. Therefore, in order to determine the large data limits of the normalized formulations it is sufficient to evaluate these two matrices in the limit $T\to\infty$, which can be done in two steps.

Firstly, it is known from calculus that if $\{\mat{Y}_t\}$ and $\{\mat{Z}_t\}$ are two matrix sequences that converge to $\mat{Y}$ and $\mat{Z}$, respectively, then the product sequence $\{\mat{Y}_t\mat{Z}_t\}$ converges to $\mat{YZ}$ \citep[see][, chapter 19]{Seber2008}. Therefore, the large data limits of $\mat{\Sigma}^{-\frac{1}{2}}\mat{A}(\mat{\Sigma}^{-\frac{1}{2}})^T$ and $\mat{\Sigma}^{-1}\mat{A}$ can be written as
\begin{align}
\lim_{T\to\infty}\mat{\Sigma}^{-\frac{1}{2}}\mat{A}(\mat{\Sigma}^{-\frac{1}{2}})^T&=\bigg(\lim_{T\to\infty}\mat{\Sigma}^{-\frac{1}{2}}\bigg)\bigg(\lim_{T\to\infty}\mat{A}\bigg)\bigg(\lim_{T\to\infty}\mat{\Sigma}^{-\frac{1}{2}}\bigg)^T\\
&=\bigg(\lim_{T\to\infty}\mat{\Sigma}^{-\frac{1}{2}}\bigg)\bigg(\lim_{T\to\infty}\mat{A}\bigg)\bigg(\lim_{T\to\infty}\mat{\Sigma}^{-\frac{1}{2}}\bigg)\label{eq:GenEVType1sym}
\end{align}
and
\begin{equation}
\lim_{T\to\infty}\mat{\Sigma}^{-1}\mat{A}=\bigg(\lim_{T\to\infty} \mat{\Sigma}^{-1}\bigg)\bigg(\lim_{T\to\infty} \mat{A} \bigg)\label{eq:GenEVType1left}
\end{equation}
respectively. As explained in Appendix \ref{app:SymNormWhite}, there are different ways to define $\mat{\Sigma}^{-\frac{1}{2}}$, which correspond to different versions of whitening, and the definition corresponding to ZCA whitening is assumed here since it preserves the correspondence between $x_i$ and $s_i$ that exists for Markovian one-hot trajectories. This choice allows the transpose to be dropped in \cref{eq:GenEVType1sym} since $\mat{\Sigma}^{-\frac{1}{2}}$ is symmetric for ZCA whitening.


Secondly, since the limit of $\mat{A}$ is already described by \cref{thm:Type1LimObjective}, the only parts of \cref{eq:GenEVType1sym,eq:GenEVType1left} that need to be considered are the limits of $\mat{\Sigma}^{-\frac{1}{2}}$ and $\mat{\Sigma}^{-1}$, respectively. Note that these limits exist only if the limit of $\mat{\Sigma}$ is positive definite. In this case, matrix inversion commutes with the limit operation since the former is a continuous function around any positive definite matrix. Thus, the remaining terms in \cref{eq:GenEVType1sym,eq:GenEVType1left} are given by
\begin{equation}
\lim_{T\to\infty}\mat{\Sigma}^{-\frac{1}{2}}=\bigg(\lim_{T\to\infty}\mat{\Sigma}\bigg)^{-\frac{1}{2}}
\end{equation}
and
\begin{equation}
\lim_{T\to\infty}\mat{\Sigma}^{-1}=\bigg(\lim_{T\to\infty} \mat{\Sigma}\bigg)^{-1}
\end{equation}
respectively. One issue with the second step in this argument is that for Type 1 SFA problems the limiting covariance matrix is $\mat{\Pi}-\vec{\pi\pi}^T$, which has an eigenvalue of $0$ and is therefore not positive definite (see \cref{thm:Type1LimConstraint}).  Therefore, in order for the normalized formulations of Type 1 SFA problems to have well-defined large data limits for a Markovian one-hot trajectory, it is necessary that noise is added to $\vec{x}(t)$, so that the large data limit of the covariance matrix becomes
\begin{equation}
\lim_{T\to\infty}\mat{\Sigma}=\mat{\Pi}-\vec{\pi\pi}^T+\sigma^2\mat{\mathbbm{1}}
\end{equation}
which is positive definite (see Appendix \ref{app:LinSFAnoise}). By the argument just given, this means that the limits of $\mat{\Sigma}^{-\frac{1}{2}}$ and $\mat{\Sigma}^{-1}$ are
\begin{equation}
\lim_{T\to\infty}\mat{\Sigma}^{-\frac{1}{2}}=\bigg(\mat{\Pi}-\vec{\pi\pi}^T+\sigma^2\mat{\mathbbm{1}}\bigg)^{-\frac{1}{2}}\label{eq:lim_invroot_cov}
\end{equation}
and
\begin{equation}
\lim_{T\to\infty}\mat{\Sigma}^{-1}=\bigg(\mat{\Pi}-\vec{\pi\pi}^T+\sigma^2\mat{\mathbbm{1}}\bigg)^{-1}\label{eq:lim_inv_cov}
\end{equation}
respectively. Inserting \cref{eq:lim_invroot_cov,eq:lim_inv_cov} into \cref{eq:GenEVType1sym,eq:GenEVType1left}, respectively, yields 
\begin{align}
\lim_{T\to\infty}\mat{\Sigma}^{-\frac{1}{2}}\mat{A}(\mat{\Sigma}^{-\frac{1}{2}})^T&=\bigg(\mat{\Pi}-\vec{\pi \pi}^T+\sigma^2\mat{\mathbbm{1}}\bigg)^{-\frac{1}{2}}\bigg(\lim_{T\to\infty}\mat{A}\bigg)\bigg(\mat{\Pi}-\vec{\pi \pi}^T+\sigma^2\mat{\mathbbm{1}}\bigg)^{-\frac{1}{2}}\label{eq:GenEVType1sym2}
\end{align}
and
\begin{equation}
\lim_{T\to\infty}\mat{\Sigma}^{-1}\mat{A}=\bigg(\mat{\Pi}-\vec{\pi\pi}^T+\sigma^2\mat{\mathbbm{1}}\bigg)^{-1}\bigg(\lim_{T\to\infty} \mat{A} \bigg)\label{eq:GenEVType1left2}
\end{equation}
which describe the large data limits of the Type 1 symmetric and left normalized SFA problems, respectively, where the terms involving $\mat{A}$ are described by \cref{thm:Type1LimObjective}.

\subsubsection{Type 2 SFA Problems}
\label{sec:LimResType2}
For a Markovian one-hot trajectory, the matrices $\widehat{\mat{\Sigma}}$, $\widehat{\dot{\mat{\Sigma}}}$, $\widehat{\mat{\Omega}}_\tau$ and $\widehat{\mat{\Psi}}_\gamma$ associated to the Type 2 SFA problems are described in the limit $T\to\infty$ by the following pair of results that are analogous to those of the previous section:

\begin{theorem}
\label{thm:Type2LimConstraint}
For an ergodic Markovian one-hot trajectory $\vec{x}(t)$, the matrix $\widehat{\mat{\Sigma}}$ associated to the Type 2 constraints obeys the following limit:
\begin{align}
\lim_{T\to\infty}\widehat{\mat{\Sigma}}&=\mat{\Pi}
\end{align}
where $\vec{\pi}$ is the unique stationary distribution of the underlying Markov chain and $\mat{\Pi}=\textup{diag}(\vec{\pi})$. Moreover, the limiting matrix $\mat{\Pi}$ is guaranteed to have no eigenvalues of $0$ (proof: see Appendix \ref{app:LimType2}).
\end{theorem}

\begin{theorem}
\label{thm:Type2LimObjective}
For an ergodic Markovian one-hot trajectory $\vec{x}(t)$, the matrices $\widehat{\dot{\mat{\Sigma}}}$, $\widehat{\mat{\Omega}}_\tau$, and $\widehat{\mat{\Psi}}_\gamma$ associated to the Type 2 objectives obey the following limits:
\begin{align}
\lim_{T\to\infty}\widehat{\dot{\mat{\Sigma}}}&=2\mat{L}_{\textup{dir}}\\
\lim_{T\to\infty}\widehat{\mat{\Omega}}_\tau&=\mat{\Pi}(\mat{P}^\tau)_{\textup{add}}\\
\lim_{T\to\infty}\widehat{\mat{\Psi}}_\gamma&=\mat{\Pi}\mat{M}_{\textup{add}}
\end{align}
where $\mat{L}_{\textup{dir}}$ is the combinatorial directed Laplacian matrix associated to the underlying Markov chain, $\mat{P}$ is the transition matrix, and $\mat{M}$ is the SR matrix with discount factor $\gamma$ (proof: see Appendix \ref{app:LimType2}).
\end{theorem}

\noindent The main insight from \cref{thm:Type2LimConstraint} and \cref{thm:Type2LimObjective} is that for this type of input in the limit $T\to\infty$ the Type 2 generalized eigenvalue problems are given by
\begin{align}
2\mat{L}_{\textup{dir}}\mat{W}&=\mat{\Pi}\mat{W}\mat{\Lambda}& \text{(SFA)}&\label{eq:GenEVType2SFA}\\
\mat{\Pi}(\mat{P}^\tau)_{\textup{add}}\mat{W}&=\mat{\Pi}\mat{W}\mat{\Lambda}&\text{($\tau$SFA)}&\label{eq:GenEVType2tauSFA}\\
\mat{\Pi}\mat{M}_{\textup{add}}\mat{W}&=\mat{\Pi}\mat{W}\mat{\Lambda} &\text{(LFSFA)}&\label{eq:GenEVType2LFSFA}
\end{align}
Thus, like in the previous section, SFA is related to $\mat{L}_{\textup{dir}}$, $\tau$SFA to $\mat{\Pi}(\mat{P}^\tau)_{\textup{add}}$, and LFSFA to $\mat{\Pi}\mat{M}_{\textup{add}}$. The only difference in comparison to Type 1 SFA problems is that the terms involving the outer product $\vec{\pi}\vec{\pi}^T$ are no longer present on either side of the equations, which is due to the fact that the centering step is omitted (see Appendix \ref{app:LimType2}).

One important consequence of \cref{thm:Type2LimConstraint} is that for a Markovian one-hot trajectory the limit of the second moment matrix $\widehat{\mat{\Sigma}}$ is $\mat{\Pi}$, which has no eigenvalues of zero and is therefore positive definite. This makes analyzing the large data limits of the normalized formulations of Type 2 SFA problems more straightforward than in the Type 1 setting. In particular, the Type 2 symmetric and left normalized SFA problems are described by $\widehat{\mat{\Sigma}}^{-\frac{1}{2}}\mat{A}(\widehat{\mat{\Sigma}}^{-\frac{1}{2}})^T$ and $\widehat{\mat{\Sigma}}^{-1}\mat{A}$ where $\mat{A}\in\{\widehat{\dot{\mat{\Sigma}}}, \widehat{\mat{\Omega}}_\tau,\widehat{\mat{\Psi}}_\gamma\}$, and for the reasons given in the previous section the limits of these matrices can be written as
\begin{align}
\lim_{T\to\infty}\widehat{\mat{\Sigma}}^{-\frac{1}{2}}\mat{A}(\widehat{\mat{\Sigma}}^{-\frac{1}{2}})^T&=\bigg(\lim_{T\to\infty}\widehat{\mat{\Sigma}}^{-\frac{1}{2}}\bigg)\bigg(\lim_{T\to\infty}\mat{A}\bigg)\bigg(\lim_{T\to\infty}\widehat{\mat{\Sigma}}^{-\frac{1}{2}}\bigg)^T\\
&=\bigg(\lim_{T\to\infty}\widehat{\mat{\Sigma}}^{-\frac{1}{2}}\bigg)\bigg(\lim_{T\to\infty}\mat{A}\bigg)\bigg(\lim_{T\to\infty}\widehat{\mat{\Sigma}}^{-\frac{1}{2}}\bigg)\label{eq:GenEVType2sym}
\end{align}
and
\begin{equation}
\lim_{T\to\infty}\widehat{\mat{\Sigma}}^{-1}\mat{A}=\bigg(\lim_{T\to\infty} \widehat{\mat{\Sigma}}^{-1}\bigg)\bigg(\lim_{T\to\infty} \mat{A} \bigg)\label{eq:GenEVType2left}
\end{equation}
respectively. Note that although $\widehat{\mat{\Sigma}}^{-\frac{1}{2}}$ does not formally correspond to whitening, its definition has the same degree of freedom that is explored for whitening in Appendix \ref{app:SymNormWhite}, and like in the previous section $\widehat{\mat{\Sigma}}^{-\frac{1}{2}}$ is chosen such that it both preserves the correspondence between $x_i$ and $s_i$ and is symmetric, which is why the transpose is dropped in \cref{eq:GenEVType2sym}. Moreover, since the large data limit of $\widehat{\mat{\Sigma}}$ is positive definite, the limits of $\widehat{\mat{\Sigma}}^{-\frac{1}{2}}$ and $\widehat{\mat{\Sigma}}^{-1}$ are well-defined without the need to add noise to $\vec{x}(t)$, and are given by
\begin{equation}
\lim_{T\to\infty}\widehat{\mat{\Sigma}}^{-\frac{1}{2}}=\mat{\Pi}^{-\frac{1}{2}}\label{eq:lim_invroot_2ndmom}
\end{equation}
and
\begin{equation}
\lim_{T\to\infty}\widehat{\mat{\Sigma}}^{-1}=\mat{\Pi}^{-1}\label{eq:lim_inv_2ndmom}
\end{equation}
respectively. These are diagonal matrices containing the reciprocal square roots and reciprocals of the stationary probabilities, i.e.\ $\frac{1}{\sqrt{\pi_i}}$ and $\frac{1}{\pi_i}$, respectively, where in the former case this property depends on the choice of $\widehat{\mat{\Sigma}}^{-\frac{1}{2}}$ used. Inserting \cref{eq:lim_invroot_2ndmom,eq:lim_inv_2ndmom} into \cref{eq:GenEVType2sym,eq:GenEVType2left}, respectively, yields
\begin{align}
\lim_{T\to\infty}\widehat{\mat{\Sigma}}^{-\frac{1}{2}}\mat{A}(\widehat{\mat{\Sigma}}^{-\frac{1}{2}})^T&=\mat{\Pi}^{-\frac{1}{2}}\bigg(\lim_{T\to\infty}\mat{A}\bigg)\mat{\Pi}^{-\frac{1}{2}}\label{eq:GenEVType2sym2}
\end{align}
and
\begin{equation}
\lim_{T\to\infty}\widehat{\mat{\Sigma}}^{-1}\mat{A}=\mat{\Pi}^{-1}\bigg(\lim_{T\to\infty} \mat{A} \bigg)\label{eq:GenEVType2left2}
\end{equation}
which describe the large data limits of the Type 2 symmetric and left normalized SFA problems, respectively, where the terms involving $\mat{A}$ are described by \cref{thm:Type2LimObjective}. These limits have a somewhat simpler form than in the Type 1 setting, which becomes particularly evident by considering how they evaluate for the different choices of $\mat{A}$. For example, if $\mat{A}=\widehat{\dot{\mat{\Sigma}}}$ then \cref{eq:GenEVType2sym2,eq:GenEVType2left2} become
\begin{align}
\lim_{T\to\infty}\widehat{\mat{\Sigma}}^{-\frac{1}{2}}\widehat{\dot{\mat{\Sigma}}}(\widehat{\mat{\Sigma}}^{-\frac{1}{2}})^T&=2\mat{\Pi}^{-\frac{1}{2}}\mat{L}_{\textup{dir}}\mat{\Pi}^{-\frac{1}{2}}
\end{align}
and
\begin{equation}
\lim_{T\to\infty}\widehat{\mat{\Sigma}}^{-1}\widehat{\dot{\mat{\Sigma}}}=2\mat{\Pi}^{-1}\mat{L}_{\textup{dir}}
\end{equation}
respectively. Note that in the literature on Laplacian matrices, $\mat{\Pi}^{-\frac{1}{2}}\mat{L}_{\textup{dir}}\mat{\Pi}^{-\frac{1}{2}}$ and $\mat{\Pi}^{-1}\mat{L}_{\textup{dir}}$ are referred to as the \textit{symmetric normalized} and \textit{random walk normalized} versions of $\mat{L}_{\textup{dir}}$, respectively \citep{Wiskott2019}. Moreover, if $\mat{A}=\widehat{\mat{\Omega}}_\tau$ then \cref{eq:GenEVType2sym2,eq:GenEVType2left2} become
\begin{align}
\lim_{T\to\infty}\widehat{\mat{\Sigma}}^{-\frac{1}{2}}\widehat{\mat{\Omega}}_\tau(\widehat{\mat{\Sigma}}^{-\frac{1}{2}})^T&=\mat{\Pi}^{\frac{1}{2}}(\mat{P}^\tau)_{\textup{add}}\mat{\Pi}^{-\frac{1}{2}}
\end{align}
and
\begin{equation}
\lim_{T\to\infty}\widehat{\mat{\Sigma}}^{-1}\widehat{\mat{\Omega}}_\tau=(\mat{P}^\tau)_{\textup{add}}
\end{equation}
repectively, while if $\mat{A}=\widehat{\mat{\Psi}}_\gamma$ then they become
\begin{align}
\lim_{T\to\infty}\widehat{\mat{\Sigma}}^{-\frac{1}{2}}\widehat{\mat{\Psi}}_\gamma(\widehat{\mat{\Sigma}}^{-\frac{1}{2}})^T&=\mat{\Pi}^{\frac{1}{2}}\mat{M}_{\textup{add}}\mat{\Pi}^{-\frac{1}{2}}
\end{align}
and
\begin{equation}
\lim_{T\to\infty}\widehat{\mat{\Sigma}}^{-1}\widehat{\mat{\Psi}}_\gamma=\mat{M}_{\textup{add}}
\end{equation}
respectively. Note that the eigenvalues and eigenvectors of $\mat{\Pi}^{\frac{1}{2}}(\mat{P}^\tau)_{\textup{add}}\mat{\Pi}^{-\frac{1}{2}}$, $(\mat{P}^\tau)_{\textup{add}}$, $\mat{\Pi}^{\frac{1}{2}}\mat{M}_{\textup{add}}\mat{\Pi}^{-\frac{1}{2}}$, and $\mat{M}_{\textup{add}}$ are all explored in \cref{thm:P+Madd}.\\

In \cref{tab:AllEigenvalueProblems}, column B shows the limits of all generalized and regular eigenvalue problems corresponding to the Type 1 and Type 2 SFA problems considered in this paper, where the general structure of the table remains the same as for column A, with the only differences being that: (i) $\vec{x}(t)$ is assumed to be a Markovian one-hot trajectory, (ii) the limit $T\to\infty$ is taken, and (iii) the transposes are dropped in the rows corresponding to symmetric normalization.

\subsection{Experimentation}
\label{sec:Exp}
This section illustrates the results of the previous section by following the pipeline outlined in \cref{fig:MDP-SFA} for a specific environment and policy. Since the main novel insight of these results concerns the relationship of $\tau$SFA and LFSFA to transition matrices and SRs, respectively, focus is placed on these two variants of SFA. Moreover, the Type 2 left normalized formulations of these problems are considered since the relationship to transition matrices and SRs is clearest for these cases (see \cref{tab:AllEigenvalueProblems}). In \cref{sec:ExpSetup} the choice of environment and policy is discussed, and in \cref{sec:ExpSim} the matrices involved in these problems as well as the corresponding outputs are computed for a time series $\vec{x}(t)$ resulting from this policy and environment.

\subsubsection{Setup}
\label{sec:ExpSetup}

The environment considered in this paper is a \textit{gridworld}, which is a canonical choice of toy model in RL research. In a gridworld, the state space $\mathcal{S}$ is a finite 2D lattice of states, often having some degree of spatial regularity, and from each state the agent can make actions that typically induce deterministic transitions to neighbouring states. This paper considers the most simple example of a gridworld, namely a rectangular open field with no barriers or obstacles, and with fixed boundaries (i.e.\ no cyclic topology). States are arranged in a square geometry, with the action set fixed across all states and involving the four cardinal directions and the possibility of remaining in the same state, i.e.\ $\mathcal{A}=(\leftarrow,\rightarrow,\uparrow, \downarrow, \bullet)$. When a forbidden action is made, such as moving right at a state on the rightmost edge, this results in a transition in the opposite direction, i.e.\ left. This rule is referred to as a \textit{reflective boundary condition} in this paper. 

The policy considered in this paper is one in which actions are selected with equal probability in each state, which is referred to henceforth as a \textit{uniform policy} and which corresponds to  $p(\leftarrow,\rightarrow,\uparrow, \downarrow, \bullet)=\frac{1}{5}=0.2$, for the environment just described. The combination of this policy together with the environment produces a Markov chain with transition statistics that are highly uniform, depending only on whether the starting state is located in the bulk, on an edge, or in a corner. This is referred to as a \textit{uniform random walk} in the current paper.

Despite being highly simplified, the choice of environment and policy is fitting to this paper for three reasons. Firstly, the results of the previous section only require that the Markov chain is ergodic, or equivalently that it is irreducible and aperiodic \citep{Seabrook2023}. For a uniform random walk, the former property holds because the underlying gridworld is connected and there is an equal probability of moving in all directions, and the latter property holds because self-transitions are included \citep{Seabrook2023}. Secondly, many SR models consider uniform random walks as a baseline case of interest \citep{Stachenfeld2014,Stachenfeld2017}. Thirdly, a rectangular open field together with a uniform policy is the closest discrete analogue of the visual environment and simple movement paradigm considered by \citet{Franzius2007a}, meaning that useful comparisons can be made to the theoretical and experimental results of this study.

\subsubsection{Simulation}
\label{sec:ExpSim}

The simulations of this section involve a rectangular open field gridworld with dimensions $L_x=15$ and $L_y=10$, which an agent explores using a uniform policy. This policy is rolled out for $T=10^6$ steps, producing a Markovian one-hot trajectory $\vec{x}(t)$ where each vector has $150$ entries corresponding to each possible state that can be occupied. The Type 2 left normalized formulations of $\tau$SFA and LFSFA are then applied to this time series, where the matrices involved have shape $150\times 150$.

It is worth noting that the uniform random walk is guaranteed to be reversible since there are no cycles in the state space that involve a net flow of probability \citep{Seabrook2023}. Thus, for the resulting time series $\vec{x}(t)$ the additive reversibilization can w.l.o.g.\ be dropped in all limits listed  in \cref{tab:AllEigenvalueProblems}. For the Type~2 left normalized formulations of $\tau$SFA and LFSFA, which generally involve the matrices $\widehat{\mat{\Sigma}}^{-1}\widehat{\mat{\Omega}}_\tau$ and $\widehat{\mat{\Sigma}}^{-1}\widehat{\mat{\Psi}}_\gamma$, respectively, this means that the limits become
\begin{align}
\mat{P}^\tau\mat{W}=\mat{W}\mat{\Lambda}\label{eq:tauSFAleft}
\end{align}
and
\begin{align}
\mat{M}\mat{W}=\mat{W}\mat{\Lambda}\label{eq:LFSFAleft}
\end{align}
Note that by \cref{thm:eigSR}, it is possible to choose a single eigenbasis that satisfies \cref{eq:tauSFAleft} for all $\tau$ as well as \cref{eq:LFSFAleft}.

\begin{figure}[h!]
\captionsetup[subfigure]{justification=centering, labelformat=empty, font=normal}
\centering
\begin{subfigure}[b]{0.16\textwidth}
         \centering
         \includegraphics[width=\textwidth]{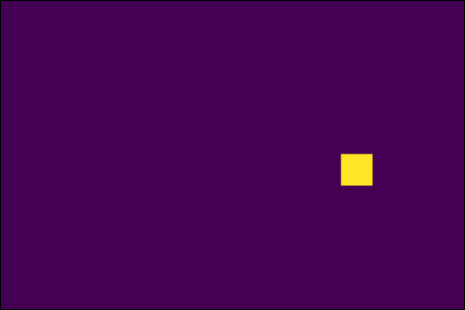}
        \caption{$\tau=0$}
        \label{fig:SimPlotColtau0}
\end{subfigure}
\hfill
\begin{subfigure}[b]{0.16\textwidth}
         \centering
         \includegraphics[width=\textwidth]{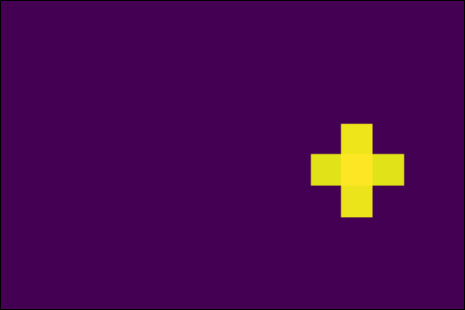}
        \caption{$\tau=1$}
        \label{fig:SimPlotColtau1}
\end{subfigure}
\hfill
\begin{subfigure}[b]{0.16\textwidth}
         \centering
         \includegraphics[width=\textwidth]{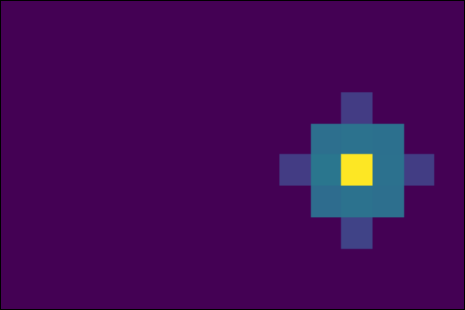}
        \caption{$\tau=2$}
        \label{fig:SimPlotColtau2}
\end{subfigure}
\hfill
\begin{subfigure}[b]{0.16\textwidth}
         \centering
         \includegraphics[width=\textwidth]{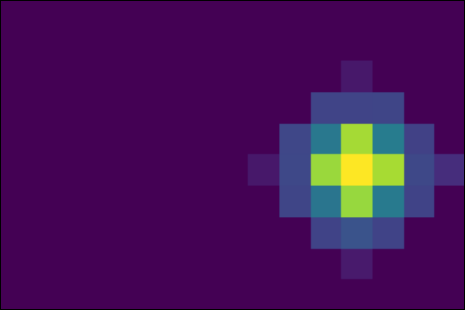}
        \caption{$\tau=3$}
        \label{fig:SimPlotColtau3}
\end{subfigure}
\hfill
\begin{subfigure}[b]{0.16\textwidth}
         \centering
         \includegraphics[width=\textwidth]{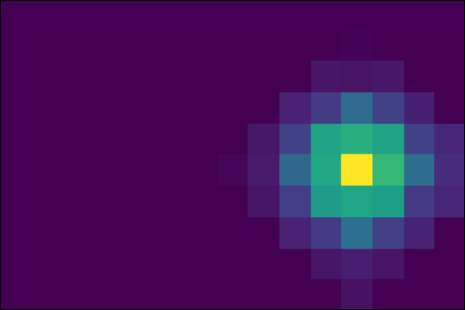}
        \caption{$\tau=4$}
        \label{fig:SimPlotColtau4}
\end{subfigure}
\hfill
\begin{subfigure}[b]{0.16\textwidth}
         \centering
         \includegraphics[width=\textwidth]{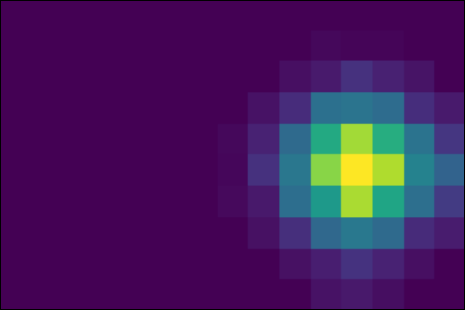}
        \caption{$\tau=5$}
        \label{fig:SimPlotColtau5}
\end{subfigure}

\vspace{-0.8em} 
\rule{\textwidth}{0.4pt} 
\vspace{-0.8em} 

\begin{subfigure}[b]{0.16\textwidth}
         \centering
         \includegraphics[width=\textwidth]{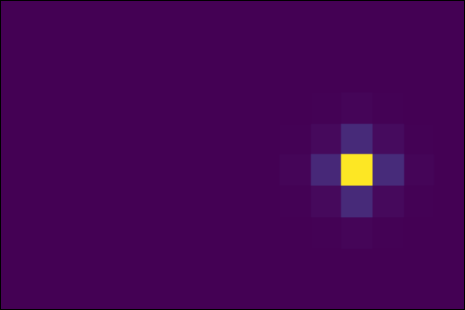}
        \caption{$\gamma=0.5$}
        \label{fig:SimPlotColgamma0.5}
\end{subfigure}
\hfill
\begin{subfigure}[b]{0.16\textwidth}
         \centering
         \includegraphics[width=\textwidth]{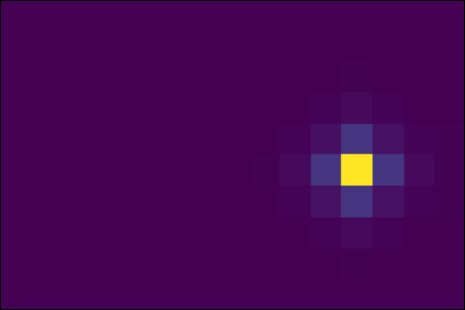}
        \caption{$\gamma=0.6$}
        \label{fig:SimPlotColgamma0.6}
\end{subfigure}
\hfill
\begin{subfigure}[b]{0.16\textwidth}
         \centering
         \includegraphics[width=\textwidth]{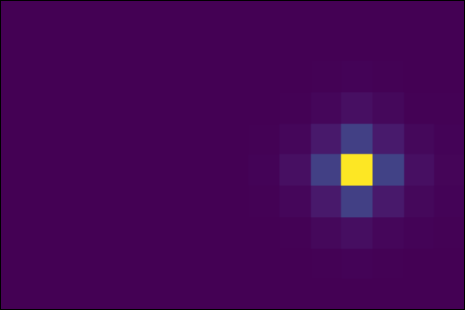}
        \caption{$\gamma=0.7$}
        \label{fig:SimPlotColgamma0.7}
\end{subfigure}
\hfill
\begin{subfigure}[b]{0.16\textwidth}
         \centering
         \includegraphics[width=\textwidth]{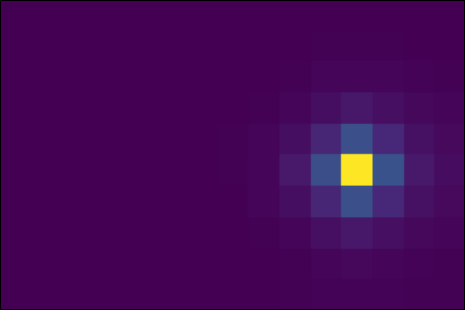}
        \caption{$\gamma=0.8$}
        \label{fig:SimPlotColgamma0.8}
\end{subfigure}
\hfill
\begin{subfigure}[b]{0.16\textwidth}
         \centering
         \includegraphics[width=\textwidth]{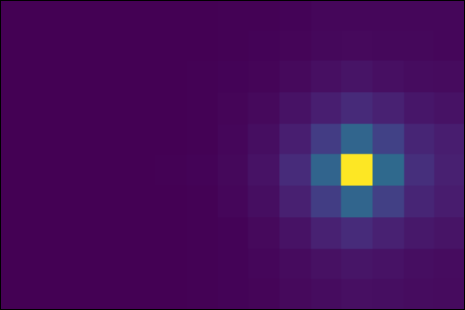}
        \caption{$\gamma=0.9$}
        \label{fig:SimPlotColgamma0.9}
\end{subfigure}
\hfill
\begin{subfigure}[b]{0.16\textwidth}
         \centering
         \includegraphics[width=\textwidth]{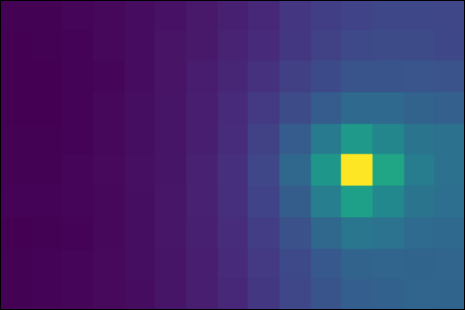}
        \caption{$\gamma=0.99$}
        \label{fig:SimPlotColgamma0.99}
\end{subfigure}
 \caption{The column of $\widehat{\mat{\Sigma}}^{-1}\widehat{\mat{\Omega}}_\tau$ (top row) and $\widehat{\mat{\Sigma}}^{-1}\widehat{\mat{\Psi}}_\gamma$ (bottom row) corresponding to a single state $s$ in the right half of the environment, for $\tau=0,1,2,3,4,5$ and $\gamma=0.5,0.6,0.7,0.8,0.9,0.99$, respectively, which equally well serve as visualizations of $\mat{P}^\tau$ and $\mat{M}$, respectively. Note that the color scales are normalized for each subplot individually.}
 \label{fig:SimPlotCol}
\end{figure}

The performance of $\tau$SFA and LFSFA in this setting can first be analyzed at the level of the matrices $\widehat{\mat{\Sigma}}^{-1}\widehat{\mat{\Omega}}_\tau$ and $\widehat{\mat{\Sigma}}^{-1}\widehat{\mat{\Psi}}_\gamma$, respectively. One way to visualize the matrices is to reshape the rows or columns so that they match the shape of the environment, i.e.\ $\mathbb{R}^{150}\to\mathbb{R}^{10\times 15}$, and then plot the resulting arrays as heat maps over the state space $\mathcal{S}$. In this paper, columns rather than rows are plotted since (i) it is more consistent with the convention of the current paper in which both input data points and eigenvectors are column vectors, and (ii) it offers a closer comparison with other studies on SR, which typically focus on the column structure \citep{Stachenfeld2014,Stachenfeld2017,Gershman2018}. The $\tau$SFA problem is illustrated in the top row of \cref{fig:SimPlotCol}, where a single column of $\widehat{\mat{\Sigma}}^{-1}\widehat{\mat{\Omega}}_\tau$ is shown for $\tau=0,1,2,3,4,5$, corresponding to a state $s$ in the right-hand side of the open field. For $\tau=1$, the shape of the field shows equal values for the state and its neighbours, and zero for all other states. This can be understood from the fact that $\widehat{\mat{\Sigma}}^{-1}\widehat{\mat{\Omega}}_1$ is close to $\mat{P}$ in the large data limit (see Equation \ref{eq:tauSFAleft}), and so the corresponding column of this matrix represents the set of transition probabilities that lead into state $s$, which are equal to $0.2$ for the state itself and the nearest neighbours but $0$ for all others. For $\tau>1$, $\widehat{\mat{\Sigma}}^{-1}\widehat{\mat{\Omega}}_\tau$ is close to $\mat{P}^\tau$, which contains the $\tau$-step transition statistics. As $\tau$ increases, the columns of $\mat{P}^\tau$ become increasingly diffuse since the number of states from which it is possible to access a given state $s$ increases. This is demonstrated in \cref{fig:SimPlotCol}, where the columns of $\widehat{\mat{\Sigma}}^{-1}\widehat{\mat{\Omega}}_\tau$ corresponding to the reference state become more and more smeared out as $\tau$ increases. For $\tau=0$, the field is non-zero only for the state itself, which agrees with the fact that $\mat{P}^0=\mat{\mathbbm{1}}$. The LFSFA problem is illustrated in the bottom row of \cref{fig:SimPlotCol}, where a single column of $\widehat{\mat{\Sigma}}^{-1}\widehat{\mat{\Psi}}_\gamma$ is shown for multiple values of $\gamma$, corresponding to the same reference state as in the top row of \cref{fig:SimPlotCol}. Note that in the simulations considered in this section, $\widehat{\mat{\Sigma}}^{-1}\widehat{\mat{\Psi}}_\gamma$ is computed with a time lag ranging from $0$ up to $\tau_{\text{max}}=100$. Note also that in the large data limit this matrix is close to the SR matrix $\mat{M}$ computed with discount factor $\gamma$, and that the shape of the fields in the bottom row of \cref{fig:SimPlotCol} is consistent with prior studies of SR for a uniform policy and open field environment \citep{Stachenfeld2014,Stachenfeld2017,Stoewer2022}. As $\gamma$ increases, the fields become less localized, which reflects how smaller (larger) values of $\gamma$ assign less (more) weight to the larger values of $\tau$. In particular, for $\gamma=0.5,0.6$ the fields are very localized (bottom row), with the $\tau=0,1$ correlations being the only significant contributions (top row). By contrast, for $\gamma=0.9,0.99$ there are significant contributions from correlations at large time scales, such as those illustrated in \cref{fig:SimPlotCol} for $\tau\geq 2$, and the resulting columns of $\widehat{\mat{\Sigma}}^{-1}\widehat{\mat{\Psi}}_\gamma$ are very diffuse. One important feature of all plots in \cref{fig:SimPlotCol} is that the activity is maximum at the reference state and decreases monotonically as a function of the distance from this state, which is henceforth referred to as being \textit{place-like}. In particular, the columns of $\widehat{\mat{\Sigma}}^{-1}\widehat{\mat{\Omega}}_\tau$ (top row) resemble discretized 2D Gaussian distributions, which results from the fact that uniform random walks diffuse symmetrically along the horizontal and vertical axes of the gridworld. Moreover, the columns of $\widehat{\mat{\Sigma}}^{-1}\widehat{\mat{\Psi}}_\gamma$ (bottom row) have a closer resemblance to discretized Laplace distributions. This can be explained by the fact that LFSFA involves temporal discounting, which is equivalent to applying an exponentially decaying filter over time.

\begin{figure}[h]
\captionsetup[subfigure]{justification=centering, labelformat=empty, font=normal}
\centering
\begin{subfigure}[b]{0.16\textwidth}
         \centering
         \includegraphics[width=\textwidth]{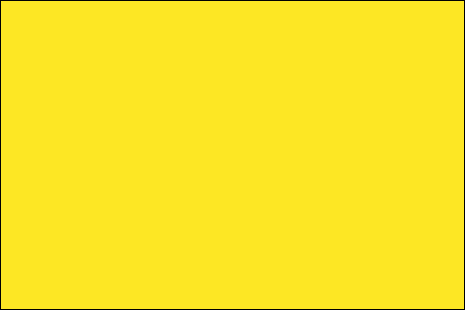}
        \caption{$j=0$}
        \label{fig:SimPlotEig0}
\end{subfigure}
\hfill
\begin{subfigure}[b]{0.16\textwidth}
         \centering
         \includegraphics[width=\textwidth]{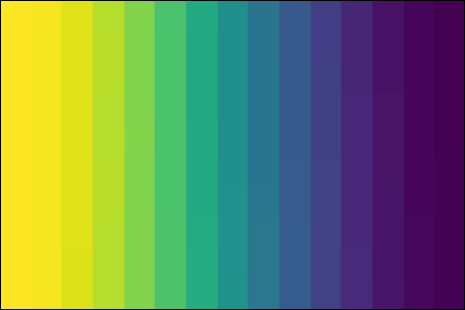}
        \caption{$j=1$}
        \label{fig:SimPlotEig1}
\end{subfigure}
\hfill
\begin{subfigure}[b]{0.16\textwidth}
         \centering
         \includegraphics[width=\textwidth]{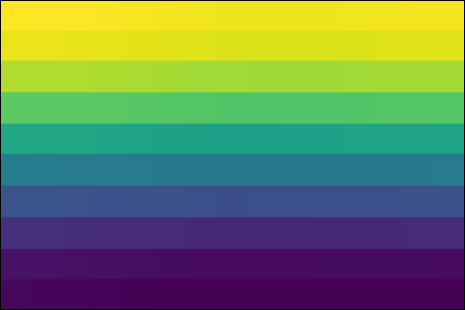}
        \caption{$j=2$}
        \label{fig:SimPlotEig2}
\end{subfigure}
\hfill
\begin{subfigure}[b]{0.16\textwidth}
         \centering
         \includegraphics[width=\textwidth]{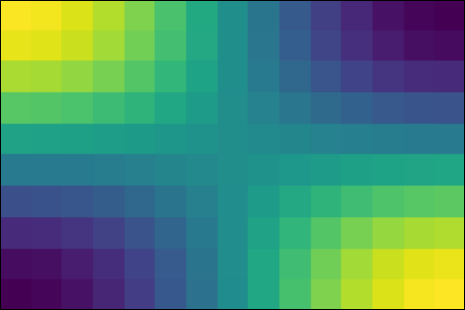}
        \caption{$j=3$}
        \label{fig:SimPlotEig3}
\end{subfigure}
\hfill
\begin{subfigure}[b]{0.16\textwidth}
         \centering
         \includegraphics[width=\textwidth]{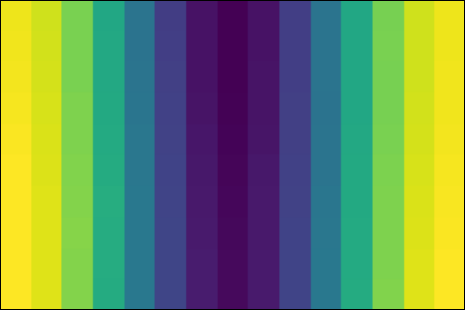}
        \caption{$j=4$}
        \label{fig:SimPlotEig4}
\end{subfigure}
\hfill
\begin{subfigure}[b]{0.16\textwidth}
         \centering
         \includegraphics[width=\textwidth]{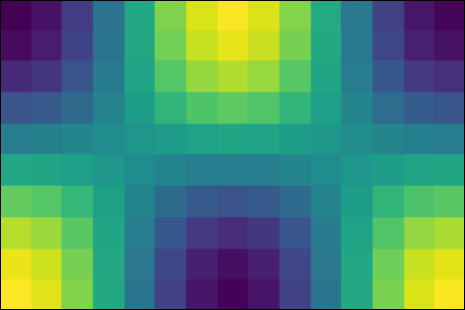}
        \caption{$j=5$}
        \label{fig:SimPlotEig5}
\end{subfigure}
\par\smallskip
\begin{subfigure}[b]{0.16\textwidth}
         \centering
         \includegraphics[width=\textwidth]{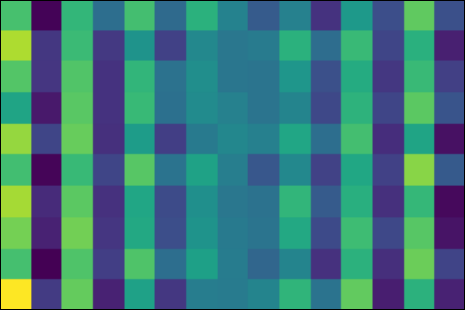}
        \caption{$j=72$}
        \label{fig:SimPlotEig25}
\end{subfigure}
\hfill
\begin{subfigure}[b]{0.16\textwidth}
         \centering
         \includegraphics[width=\textwidth]{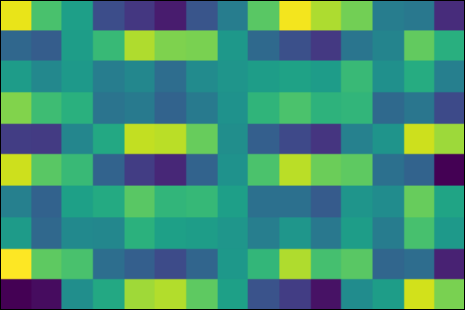}
        \caption{$j=73$}
        \label{fig:SimPlotEig45}
\end{subfigure}
\hfill
\begin{subfigure}[b]{0.16\textwidth}
         \centering
         \includegraphics[width=\textwidth]{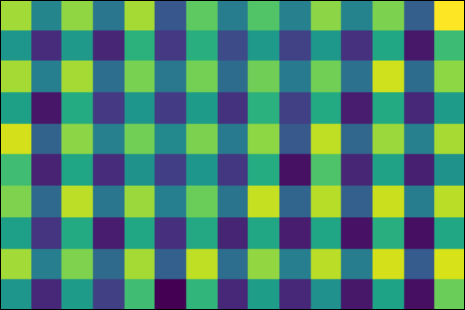}
        \caption{$j=74$}
        \label{fig:SimPlotEig65}
\end{subfigure}
\hfill
\begin{subfigure}[b]{0.16\textwidth}
         \centering
         \includegraphics[width=\textwidth]{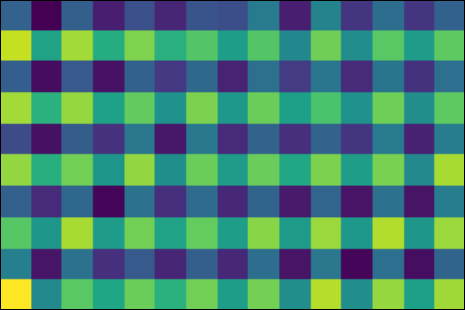}
        \caption{$j=75$}
        \label{fig:SimPlotEig85}
\end{subfigure}
\hfill
\begin{subfigure}[b]{0.16\textwidth}
         \centering
         \includegraphics[width=\textwidth]{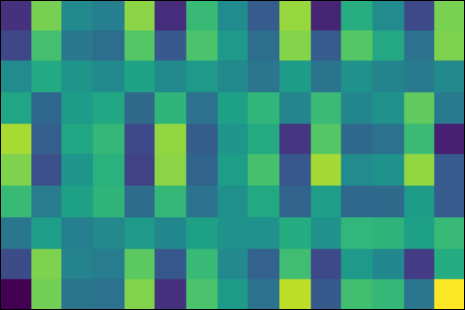}
        \caption{$j=76$}
        \label{fig:SimPlotEig105}
\end{subfigure}
\hfill
\begin{subfigure}[b]{0.16\textwidth}
         \centering
         \includegraphics[width=\textwidth]{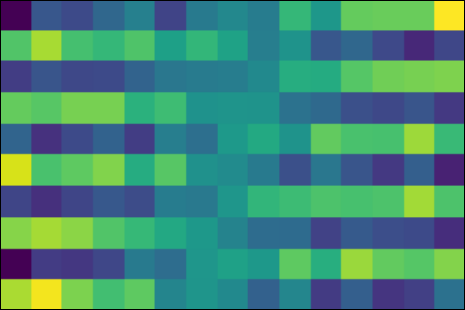}
        \caption{$j=77$}
        \label{fig:SimPlotEig125}
\end{subfigure}
\par\smallskip
\begin{subfigure}[b]{0.16\textwidth}
         \centering
         \includegraphics[width=\textwidth]{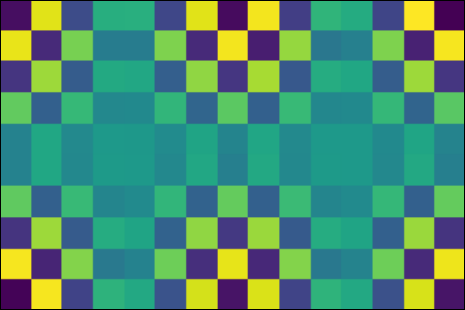}
        \caption{$j=144$}
        \label{fig:SimPlotEig144}
\end{subfigure}
\hfill
\begin{subfigure}[b]{0.16\textwidth}
         \centering
         \includegraphics[width=\textwidth]{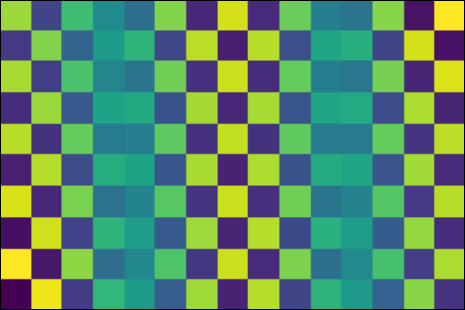}
        \caption{$j=145$}
        \label{fig:SimPlotEig145}
\end{subfigure}
\hfill
\begin{subfigure}[b]{0.16\textwidth}
         \centering
         \includegraphics[width=\textwidth]{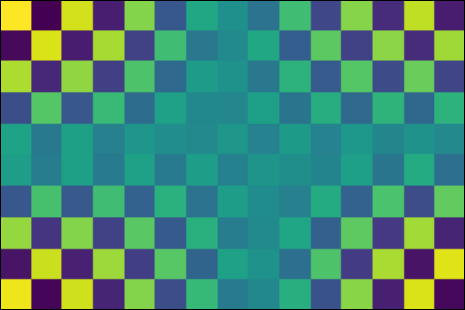}
        \caption{$j=146$}
        \label{fig:SimPlotEig146}
\end{subfigure}
\hfill
\begin{subfigure}[b]{0.16\textwidth}
         \centering
         \includegraphics[width=\textwidth]{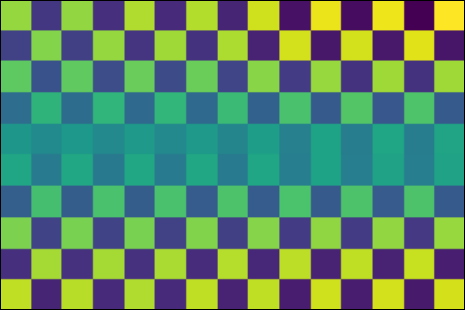}
        \caption{$j=147$}
        \label{fig:SimPlotEig147}
\end{subfigure}
\hfill
\begin{subfigure}[b]{0.16\textwidth}
         \centering
         \includegraphics[width=\textwidth]{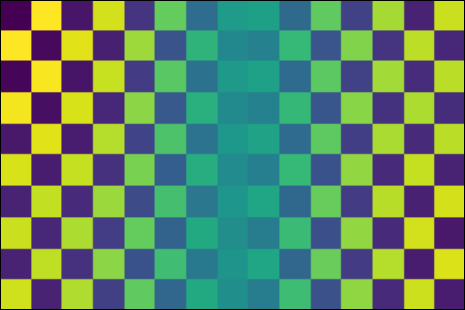}
        \caption{$j=148$}
        \label{fig:SimPlotEig148}
\end{subfigure}
\hfill
\begin{subfigure}[b]{0.16\textwidth}
         \centering
         \includegraphics[width=\textwidth]{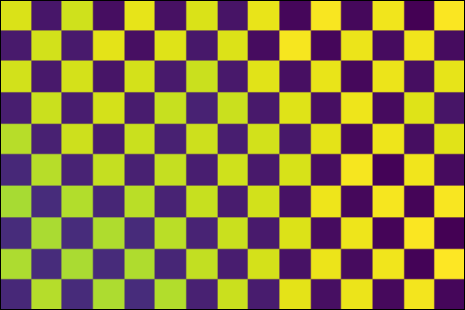}
        \caption{$j=149$}
\label{fig:SimPlotEig149}
\end{subfigure}
 \caption{A selection of $18$ outputs of the Type 2 left normalized CSFA problem, computed with the eigenvectors of $\widehat{\mat{\Sigma}}^{-1}\widehat{\mat{\Omega}}_1$, with the index of each output indicated below each subplot. Note that the color scales are normalized for each subplot individually.}
 \label{fig:SimPlotEig}
\end{figure}

Since \cref{eq:tauSFAleft,eq:LFSFAleft} can be solved with a common set of eigenvectors, for a long enough time series the solutions to $\tau$SFA are also solutions to LFSFA. This section first considers $\tau$SFA for $\tau=1$, i.e.\ CSFA, and studies how the correlation values, or equivalently the eigenvalues, of these solutions vary between the problems. The CSFA solutions are computed by performing an eigendecomposition of $\widehat{\mat{\Sigma}}^{-1}\widehat{\mat{\Omega}}_1\in\mathbb{R}^{150\times 150}$. This produces $150$ solutions in total, which are ordered by an index $j$ starting at $0$, since this convention is more natural in the Type 2 setting (see \cref{sec:T2SFA}). Moreover, for a given eigenvector, the output can be computed for every state in $\mathcal{S}$, producing an output array that can be visualized as a heat map across $\mathcal{S}$. In \cref{fig:SimPlotEig}, a selection of the CSFA outputs are visualized in this way, with the index $j$ indicated beneath each plot. The first and last row contain the $6$ most highly correlated and anti-correlated outputs, respectively, while the middle row contains the $6$ most intermediate outputs. The correlation/eigenvalue $\text{C}_1$ corresponding to each output is shown in the second row of \cref{tab:CorrValstauSFA}.

\begin{table}[h]
\centering
\renewcommand{\arraystretch}{2}
\resizebox{\textwidth}{!}{
\begin{tabular}{|c|c c c c c c c c c c c c c c c c c c |} 
\hline
$j$ & 0 & 1 & 2 & 3 & 4 & 5 & 72 & 73 & 74 & 75 & 76 & 77 & 144 & 145 & 146 & 147 & 148 & 149 \\ [1ex]
 \hline
$\text{C}_1$ & 1 & 0.99 & 0.98 & 0.97 & 0.96 & 0.94 & 0.21 & 0.21 & 0.2 & 0.2 & 0.19 & 0.19 & -0.54 & -0.56 & -0.57 & -0.58 & -0.59 & -0.6 \\[1ex] 
 \hline
$\text{C}_2$ & 1 & 0.98 & 0.95 & 0.93 & 0.92 & 0.88 & 0.05 & 0.04 & 0.04 & 0.04 & 0.04 & 0.04 & 0.29 & 0.31 & 0.32 & 0.33 & 0.35 & 0.36 \\[1ex] 
\hline 
$\text{C}_3$ & 1 & 0.97 & 0.93 & 0.9 & 0.89 & 0.82 & 0.01 & 0.01 & 0.01 & 0.01 & 0.01 & 0.01 & -0.15 & -0.18 & -0.18 & -0.19 & -0.21 & -0.22 \\[1ex] 
 \hline
  $\text{F}_{0.8}$ & 5 & 4.81 & 4.56 & 4.4 & 4.32 & 3.98 & 1.2 & 1.2 & 1.19 & 1.19 & 1.18 & 1.18 & 0.7 & 0.69 & 0.69 & 0.68 & 0.68 & 0.68 \\[1ex] 
 \hline
 $\text{F}_{0.9}$ & 10 & 9.17 & 8.21 & 7.64 & 7.38 & 6.35 & 1.23 & 1.23 & 1.22 & 1.22 & 1.21 & 1.21 & 0.67 & 0.67 & 0.66 & 0.66 & 0.65 & 0.65 \\[1ex] 
 \hline
\end{tabular}}
\caption{The values of the time-lagged correlations $\text{C}_1$, $\text{C}_2$, and $\text{C}_3$, as well as the linearly filtered correlations $\text{F}_{0.8}$ and $\text{F}_{0.9}$, for all outputs visualized in \cref{fig:SimPlotEig}.}
\label{tab:CorrValstauSFA}
\end{table}

The CSFA outputs in \cref{fig:SimPlotEig} can be interpreted using \cref{eq:tauSFAleft}, which describes the form of this problem in the limit $T\to\infty$. In particular, in the large data limit the matrix $\widehat{\mat{\Sigma}}^{-1}\widehat{\mat{\Omega}}_1$ is equal to the underlying transition matrix $\mat{P}$. Such matrices are guaranteed to have a constant right eigenvector with eigenvalue $\lambda=1$ \citep{Seabrook2023}, which corresponds to the constant output that is guaranteed whenever the zero mean constraint is omitted from SFA. This output is the first one illustrated in \cref{fig:SimPlotEig}. Moreover, since the underlying Markov chain is reversible, the other eigenvectors of $\mat{P}$ can be understood using spectral graph theory, which states that the corresponding eigenvalues describe the smoothness of each eigenvector across the state space $\mathcal{S}$ \citep{Seabrook2023}. In particular, eigenvectors with large positive eigenvalues have similar values for nearby states, while those with large negative eigenvalues alternate a lot between nearby states, which can be clearly seen from the outputs in the first and last row of \cref{fig:SimPlotEig}, respectively, as well as the corresponding correlations/eigenvalues in \cref{tab:CorrValstauSFA}.

Alternatively, the plots of \cref{fig:SimPlotEig} can be interpreted using various theoretical insights from \cref{sec:SFA}. In particular, each output has a \textit{grid-like} structure that resembles a discretized 2D cosine function, or in other words a discrete analogue of the analytical solutions studied in \citep{Franzius2007a} and explored in \cref{sec:SFAspatial}. Moreover, in analogy to those latter solutions, each output illustrated in \cref{fig:SimPlotEig} factorizes into a product of 1D cosine functions defined over the $x$ and $y$ axes of the gridworld, with each 1D function having a correspondence to the discrete free responses explored in \cref{sec:SFAfree}. One difference between the two cases is that while the analytical solutions studied in \citep{Franzius2007a} all have positive correlations, the outputs in the second row of \cref{fig:SimPlotEig}, as well as some in the middle row, have a negative correlation (see \cref{tab:CorrValstauSFA}). These negatively correlated solutions occur as a result of the environment being discretized, in analogy to the discretized free responses explored in \cref{sec:SFAfree}.

It should be emphasized that the exact form of the CSFA outputs, as well as the associated correlation values, depends in part on various implementation details of the underlying model. For example, while the simulations of this paper use a reflective boundary condition, other choices of boundary condition produce similar outputs to those in \cref{fig:SimPlotEig} but with additional edge effects. A reflective boundary condition is used in this paper since it offers the closest correspondence to the solutions studied in \cref{sec:SFAfree} and \cref{sec:SFAspatial}. Moreover, self-transitions are allowed in the random walk since this is one way to ensure ergodicity, which is required for the results of \cref{sec:LimRes} to hold. One side effect this has is that the correlation values of all outputs except $j=1$ are slightly larger than their counterparts in the free response setting. For example, the output for $j=150$ resembles a product of signals that correspond to the strictly alternating free response in \cref{fig:OscSig}. While the latter is perfectly anti-correlated, i.e.\ $\text{C}_1=-1$, the former is only moderately anti-correlated, i.e.\ $\text{C}_1=-0.6$. This difference arises because, in the current model, the random walker remains in the same state with probability $0.2$ at each time point. If this happens, the random walker observes the same output values from one time step to the next, preventing perfect anti-correlation. One way to reduce this disagreement while maintaining ergodicity is to decrease the probability of self-transitions to some value $\epsilon\ll 1$.

For each output $y_j(t)=\vec{w}_j^T\vec{x}(t)$ of the CSFA problem, the quantities $\text{C}_\tau$ and $\text{F}_\gamma$ defined in \cref{sec:SFAgentaumulti,sec:SFAgentaugeq1}, respectively, can be computed using the following \textit{generalized Rayleigh quotients} \citep{Ghojogh2023}:
\begin{align}
\text{C}_\tau&=\frac{\vec{w}_j^T\widehat{\mat{\Omega}}_\tau\vec{w}_j}{\vec{w}_j^T\widehat{\mat{\Sigma}}\vec{w}_j}\\
\text{F}_\gamma&=\frac{\vec{w}_j^T\widehat{\mat{\Psi}}_\gamma\vec{w}_j}{\vec{w}_j^T\widehat{\mat{\Sigma}}\vec{w}_j}
\end{align}
The third and fourth rows of \cref{tab:CorrValstauSFA} show the values of $\text{C}_2$ and $\text{C}_3$ for each of the outputs in \cref{fig:SimPlotEig}, respectively. Note that outputs with positive $\text{C}_1$ have positive $\text{C}_2$ and $\text{C}_3$, while outputs with negative $\text{C}_1$ have positive $\text{C}_2$ and negative $\text{C}_3$. In agreement with the free responses presented in \cref{sec:SFAfree}, this reflects the fact that even values of $\tau$ tend to order outputs in a different way to $\tau=1$ while odd values tend to maintain the same ordering. Moreover, \cref{eq:tauSFAleft} implies that $\text{C}_2$ and $\text{C}_3$ are approximately the eigenvalues of $\mat{P}^2$ and $\mat{P}^3$, respectively, which are known to be $\lambda^2$ and $\lambda^3$ for each eigenvalue of $\mat{P}$ \citep{Seabrook2023}. Analogously, in \cref{tab:CorrValstauSFA} the values of $\text{C}_2$ and $\text{C}_3$ are approximately equal to $\text{C}_1^2$ and $\text{C}_1^3$ for each output.\footnote{Note that in order for the correlations to be related in this way, it is necessary that the underlying chain is reversible, so that all additive reversibilizations can be dropped, which was assumed in \cref{eq:tauSFAleft}.} Note that as $\tau$ increases the associated eigenvalues $\lambda^\tau$ become increasingly clustered around $0$, which can be seen clearly by comparing the $\tau=1,2,3$ correlation values for the intermediate outputs in \cref{tab:CorrValstauSFA}. By contrast, the correlation values for $\tau=1$  are less clustered, which in practice means that the associated outputs are more numerically stable than for larger $\tau$. The fifth and sixth rows of \cref{tab:CorrValstauSFA} show the values of $\text{F}_{0.8}$ and $\text{F}_{0.9}$ for each of the signals in \cref{fig:SimPlotEig}, respectively. The values of $\text{F}_{\gamma}$ have a magnitude that is inversely related to $\gamma$, in analogy to how the eigenvalues of $\mat{M}$ relate to $\gamma$, and they are ordered in the same way as $\text{C}_1$, in analogy to how the eigenvalues of $\mat{M}$ are ordered in the same way as those of $\mat{P}$ for a sufficiently large horizon (see \cref{thm:eigSR,sec:SRtime}).

It is also important to acknowledge that all outputs visualized in \cref{fig:SimPlotEig} have a high degree of spatial regularity. Part of this regularity stems from the type of input $\vec{x}(t)$ used. In particular, $T$ is large in comparison to $|\mathcal{S}|$, so that the limits described by \cref{eq:tauSFAleft} can be assumed to a good degree of approximation, and since the underlying environment and policy are highly uniform, so too are the approximate temporal statistics. An additional contributing factor is that the SFA transformation used is also in some sense regular, i.e.\ the addition of noise is not necessary since $\widehat{\mat{\Sigma}}\approx\mat{\Pi}$ has no eigenvalues of $0$ and a linear transformation is performed. By contrast, general applications of SFA do not produce such regular outputs due to the impact of various types of noise (e.g. finite sample errors, irregular input statistics, addition of noise due to eigenvalues of $0$, as well as noise that arises from using finite function spaces in high-dimensional settings). For the simulations considered in this section, there are two simple manipulations that can increase the level of noise in $\vec{x}(t)$. Firstly, one can increase the value of $\tau$ used in $\tau$SFA, or equivalently the value of $\gamma$ in LFSFA, which increases the sample error since transition statistics become more noisy as $\tau$ increases. Secondly, one can decrease the relative size of $T$ in comparison to $|\mathcal{S}|$ used in any of the SFA problems, for example by using shorter samples for the same state space, which also increases the sample error. Note that in either case, an increase in the sample error reflects the fact that the SFA problems are further from the limits described by \cref{eq:tauSFAleft,eq:LFSFAleft}. Moreover, the main way in which this change becomes apparent is that the most intermediate SFA outputs, such as those in the middle row of \cref{fig:SimPlotEig}, become less regular (not shown). One way to understand this is that the correlation values of these outputs are much closer together than for the fast or slow outputs (see \cref{tab:CorrValstauSFA}), meaning that these outputs are the first to start mixing when the input $\vec{x}(t)$ is subject to noise. Moreover, as explained in the previous paragraph, for $\tau$SFA the clustering of correlation values around $0$ is more prominent for larger $\tau$, which means that the number of intermediate outputs affected by noise increases with $\tau$. A similar effect also happens in LFSFA when the parameter $\gamma$ is set very close to $1$, since this assigns more weight to larger values of $\tau$ (not shown).

\section{Discussion}
\label{sec:Disc}

\subsection{Key contributions}
This paper provides an in-depth exploration of the relationship between SFA and SR. In \cref{sec:SR}, the theory of SR matrices is reviewed, with a particular focus on their corresponding eigenvalues and eigenvectors. After first exploring the case where the underlying Markov chain is reversible, generalizations are made to the non-reversible setting. To do this, the following two steps are necessary: (i)~the time reversed SR is introduced, known in the literature as PR and denoted in this paper as $\mat{M}_{\textup{rev}}$, (ii) the quantity $\mat{M}_{\textup{add}}$ is defined, which is an additive mixture of $\mat{M}$ and $\mat{M}_{\textup{rev}}$. It is worth reiterating that $\mat{M}_{\textup{add}}$ is, to the authors' knowledge, a previously unexplored quantity in the literature. Moreover, after the study by \citet{Keck2024}, which involves $\mat{M}(\mat{P}_{\textup{add}})$, this paper is the second to consider SR matrices that combine forward and backward temporal statistics, referred to generally as IR in the current text. The authors hope that the concepts and theorems explored in \cref{sec:SR} could be of general use in the SR literature. In \cref{sec:SFA}, the theory of SFA problems is explored, which has already received a lot of attention in the literature. Several variants, formulations, and types of SFA are introduced that have been defined at various stages since the original publication by \citet{Wiskott1998b}. In \Crefrange{sec:SFAdef}{sec:SFAgentaumulti}, each of these problems is analyzed in terms of its objective and constraints, with attention paid to how both of these impact the types of solutions that can occur. In \cref{sec:Type1sol,sec:Type2sol}, solutions are formulated in terms of generalized and regular eigenvalue problems, which provides a suitable level at which they can be compared to quantities from the Markov chain/SR setting. In \cref{sec:SFAfree,sec:SFAspatial}, the SFA solutions are additionally studied in two domains that are particularly relevant for the comparison to SR, namely, the cases of free responses and spatial environments. In total, 18 SFA problems are presented in a way that is both intuitive yet comprehensive, thus making the concepts accessible to a wider academic audience. \cref{sec:LimRes} presents the main results of the paper, namely that for a Markovian one-hot trajectory each version of SFA is equivalent in the limit $T\to\infty$ to an eigenvalue problem involving a matrix associated to the underlying Markov chain. In particular, SFA is related to $\mat{L}_{\textup{dir}}$, $\tau$SFA to $(\mat{P}^\tau)_{\textup{add}}$, and LFSFA to $\mat{M}_{\textup{add}}$, and in each case the delta/correlation values and SFA weights correspond to the eigenvalues and eigenvectors of these matrices, respectively. In \cref{sec:Exp}, the theoretical results are explored using simulations of a uniform random walk in an open field gridworld environment. These simulations involve the Type 2 left normalized formulations of $\tau$SFA and LFSFA, which are the most suitable for exploring the relation to transition matrices and SR, respectively, since the analytical correspondence is closest in these cases. The simulations are analyzed both in terms of the main matrices, i.e.\ their column structure, as well as the corresponding eigenvectors and eigenvalues. Moreover, all visualizations are made as heatmaps over the gridworld's state space and show a close agreement with other studies that have considered SR for this choice of environment and policy \citep{Stachenfeld2014,Stachenfeld2017,Stoewer2022}.


\subsection{Relation to prior work}

\subsubsection{Conceptual relations}
In order to accurately frame the relation of the current paper to prior research, it is helpful to first give a historical account of the origins of both SR and SFA.

SR was originally developed by \citet{Dayan1993} as an elementary representation for RL tasks. Formally, the representation is a statistical measure of the transitions that are observed of an RL agent across multiple time scales under a given policy. Moreover, SR is a particularly natural choice of representation for finite MDPs since it (i) is a core component of the value function, (ii) can be learnt in different ways depending on the type of RL task being studied \citep{Russek2017}, and (iii) can alternatively be defined for state action pairs \citep{Ducarouge2017}.\footnote{Note that the method of \textit{predictable feature analysis} (PFA), which is related to SFA, has also been applied to inputs involving actions \citep{Richthofer2018}.} More recently, SR has become a popular concept in the fields of computational neuroscience, cognitive science, and artificial intelligence \citep[for a review, see][]{Carvalho2024b}. A central line of work underlying this surge in interest is that of \citet{Stachenfeld2014,Stachenfeld2017}, which were the first studies to observe the place- and grid-like structure of the columns and eigenvectors of $\mat{M}$, respectively, in toy RL tasks.

SFA was originally developed by \citet{Wiskott1998b} as an unsupervised learning method for doing dimensionality reduction on time series data. Originally conceived as a model of visual processing, SFA is based on the idea that in biological systems the features that contain the most meaningful information are often those that vary most slowly over time, which is sometimes referred to as the \textit{slowness principle} \citep{Franzius2007a}. One strength of SFA is that it is particularly amenable to analytical study, as demonstrated by a number of publications since its creation \citep{Blaschke2006,Creutzig2008,Sprekeler2008,Sprekeler2009,Sprekeler2011,Sprekeler2014,Wiskott2003,Richthofer2020}. Another strength is that it can flexibly deal with any time series $\vec{x}(t)$ or function space $\mathcal{F}$, which is one of the reasons it has received applications in a variety of technical domains \citep[for reviews, see][]{Escalante2012,Song2024} as well as in computational neuroscience \citep{Berkes2005,Franzius2007a,Franzius2008,Legenstein2010,Schoenfeld2015}. One particularly relevant application to the current paper is that of \citet{Franzius2007a}, which was discussed in \cref{sec:SFAspatial}, and which was the first study to demonstrate that SFA can learn both place- and grid-like representations of spatial environments.

At the most general level, this paper shows that SFA and SR can be used as alternative methods for achieving equivalent state representations of a finite MDP. On one hand, this is an insightful finding, given that SFA and SR stem from different areas of machine learning research and are based on different underlying computational principles. On the other hand, the work of \citet{Franzius2007a} already indicates that SFA is capable of producing place- and grid-like representations of an agent's environment. Thus, an important question is how the representations considered in \citep{Franzius2007a} compare to those generated by SFA, or equivalently SR, in the current paper. One similarity already pointed out in \cref{sec:ExpSim} is that the SFA outputs of the current model visually resemble discretized versions of the optimal outputs studied in \citep{Franzius2007a}. However, beyond this visual correspondence there are two major differences that the authors wish to emphasize.

Firstly, a key distinction lies in the type of input data and function space considered. In \citep{Franzius2007a}, SFA is trained on naturalistic image sequences that mimic the visual stream of a rat. Each image in this sequence has a dimensionality of $38,400$ and is fed into an hSFA network, which realizes a highly expressive non-linear function space $\mathcal{F}$. The experiments performed in this paper show that, depending on the type of movement statistics used, the slow features learnt by the hSFA network form grid-like representations of the agent's location or head direction, and in the former case the application of an additional ICA step produces place-like representations of the agent's location. Moreover, although the model is not a complete account of the self-organization of place- and grid-cell activity in animals, it nonetheless suggests that slowness may be involved at some level of information processing in these systems. This stands in stark contrast to SR, which is primarily a representation based on an agent's transition statistics and does not learn anything per se. Instead, what the studies of \citet{Stachenfeld2014,Stachenfeld2017} demonstrate is that (i) place- and grid-like representations are useful for maximizing reward in spatial RL tasks, and (ii) the representations produced by SR distort in response to a number of experimental variables in a way that aligns with observations of place- and grid-cell activity in animals. Note that neither of these insights indicate SR as a learning mechanism of these representations, which is a perspective recently explored by \citet{George2024}. In particular, the author argues that since SR assumes an underlying Markovian one-hot trajectory, it essentially receives a perfect place-like representation as input, meaning that the spatial layout of the state space is perfectly known by the agent to begin with. Note that the same point applies to the application of SFA studied in \cref{sec:Results}, which also assumes an underlying Markovian one-hot trajectory. \citet{George2024} further argues that many of the spatial properties of SR can simply be attributed to the underlying Markov chain. This point is consistent with the simulations of \cref{sec:ExpSim}, since place-like representations occur not only in the columns of $\widehat{\mat{\Sigma}}^{-1}\widehat{\mat{\Psi}}_\gamma$ but also of $\widehat{\mat{\Sigma}}^{-1}\widehat{\mat{\Omega}}_\tau$, which correspond to $\mat{M}$ and $\mat{P}^\tau$, respectively, and the grid-like representations studied are CSFA solutions related to the eigenvectors of $\widehat{\mat{\Sigma}}^{-1}\widehat{\mat{\Omega}}_1$, which corresponds to $\mat{P}$.

Secondly, the relation between place- and grid-like representations in \citep{Franzius2007a} differs significantly to that in the current paper. In the former case, grid-like representations are the outputs of hSFA, which are transformed into place-like representations using ICA. In particular, the ICA transformation takes $N$ slow features as input to produce $N$ place-fields. Note that due to the decorrelation constraint of SFA, the overlap between the resulting place-fields is minimized. Moreover, the place-field size depends on $N$, with less (more) slow features leading to coarse (fine) place-fields. By comparison, in the current paper, as well as in \citep{Stachenfeld2014,Stachenfeld2017}, place-like representations are observed in the columns of the matrices being considered to begin with, which reflect the movement statistics of the agent, while grid-like representations are observed in the corresponding eigenvectors. In contrast to \citep{Franzius2007a}, the place-field size is controlled by $\gamma$, and there is one place-field for each state in $\mathcal{S}$, meaning that there is significant spatial overlap in these fields. Moreover, it is worth noting that in \citet{Stachenfeld2017} the authors develop the hypothesis that grid-cell activity forms a low-dimensional decomposition of place-cell activity, based on the fact that the grid-fields (eigenvectors) and place-fields (columns) of their model in some sense have this relation. However, this hypothesis makes little sense in the model of \citet{Franzius2007a}, since $N$ grid-fields are associated to an equal number of place-fields.

\subsubsection{Technical relations}
It is worth noting that although the experiments of \cref{sec:Exp} use a uniform policy and open field environment for simplicity, the limit results of \cref{sec:LimRes} apply generally to any environment and policy for which the associated Markov chain is ergodic. For finite MDPs, ergodicity holds whenever the state space is connected and there is some probability of random exploration \citep{Seabrook2023}, which is the case for all the experiments performed in \citep{Stachenfeld2014,Stachenfeld2017}. This point is meaningful, since it indicates that in the large data limit the relation between LFSFA and SR generalizes to all experimental settings considered in these papers, such as policies involving directional preference or goal directedness, as well as more general environmental topology/geometry. Therefore, it is reasonable to infer that LFSFA has a similar level of agreement with place- and grid-cell activity in animals as demonstrated for SR in \citep{Stachenfeld2014,Stachenfeld2017}. One important property about the performance of SFA for general policies is that for non-uniform exploration of a spatial environment the slow features typically have amplitudes that are inversely related to the visitation rates of each location \citep[see the discussion of][]{Franzius2007a}. Therefore, the main results of this paper suggest that by analogy the right eigenvectors of $\mat{L}_{\textup{dir}}$, $(\mat{P}^\tau)_{\textup{add}}$, and $\mat{M}_{\textup{add}}$ should generally have amplitudes that are inversely related to the stationary probabilities of each state. Note that for transition matrices of reversible Markov chains, this is explored using a toy model in \citet[][, Section 4.2]{Seabrook2023}.

Moreover, it is worth reiterating that in each SFA problem of this paper the matrix containing the temporal statistics of $\vec{x}(t)$ is guaranteed to be diagonalizable with real eigenvalues and a full set of real eigenvectors, and that this holds because of the symmetrization present in SFA. For Markovian one-hot trajectories, this symmetrization is most naturally formulated in terms of additive reversibilization. For the uniform random walk considered in \cref{sec:Exp}, reversibilization can be dropped from all eigenvalue problems in column B of \cref{tab:AllEigenvalueProblems}, since reversibility is automatically satisfied. However, in the more general case of a non-reversible ergodic Markov chain, reversibilization remains valid in each of the limits, and it is precisely because of this that $(\mat{P}^\tau)_{\textup{add}}$ and $\mat{M}_{\textup{add}}$ are guaranteed to have real eigenvalues and a full set of real eigenvectors (see \cref{thm:P+Madd}). The fact that symmetrization/reversibilization is an explicit property of SFA problems marks one important difference to the SR setting, where there is generally no such symmetrization/reversibilization. Instead, as explained in \cref{sec:SReigrev}, many SR models introduce reversibility as an additional implicit assumption by formulating the Markov chain as a random walk on an undirected graph. Moreover, one of the most commonly used approximation methods for SR is temporal difference (TD) learning. Even for an underlying Markov chain that is reversible there is no guarantee that a TD approximation $\widehat{\mat{M}}$ has a real-valued eigendecomposition due to approximation errors, and unlike SFA there is no inherent symmetrization/reversibilization in this method. One way to avoid this issue is to compute $(\widehat{\mat{M}})_{\textup{add}}$, which does have a real-valued eigendecomposition (see \cref{thm:P+Madd}). Alternatively, the eigenvectors of $\mat{M}$ can be approximated independently using the gradient descent method of \citet{Stachenfeld2014,Stachenfeld2017}, which is defined in such a way that it always produces real-valued vectors. It is interesting to note that in \citep{Stachenfeld2017} this gradient descent method is motivated conceptually using the notion of slowness. However, it is worth emphasizing three ways in which this method is distinct from SFA: (i) the SFA constraints are absent from the method, and in particular it is unclear why the learnt eigenvectors are different and not simply all equal to the constant vector, (ii) the method optimizes the full basis of eigenvectors rather than each one individually, which in principle could lead to mixing of independent eigenvectors, and (iii) a reversible Markov chain is assumed as input, but for a non-reversible chain it is unclear how the resulting real-valued vectors would be related to the true eigenvectors of $\mat{M}$ which are generally complex-valued.

Another way in which the current paper is connected to prior research concerns the limit results explored in \cref{sec:LimRes}. In particular, for $\tau$SFA the limits are conceptually related to something known as the \textit{maximum likelihood estimation} (MLE) of a transition matrix, which appears often in the literature on Markov state models \citep[see][, in particular chapter 4]{Bowman2014}. This typically involves a Markovian one-hot trajectory $\vec{x}(t)$ as input, from which the $\tau$-step transition matrix is approximated by
\begin{equation}
(\widehat{\mat{P}^\tau})_{ij}=\frac{\#_\tau(s_i\to s_j)}{\sum_j \#_\tau(s_i\to s_j)}\label{eq:MLEPtau}
\end{equation}
where $\#_\tau(s_i\to s_j)$ counts the number of times state $s_j$ is occupied $\tau$ time steps after state $s_i$ is occupied and the denominator marginalizes over $s_j$ in order to count the number of time $s_i$ is occupied. Note that this approximation is \textit{asymptotically unbiased}, meaning that it is exact in the infinite time limit, i.e.\
\begin{align}
\lim_{T\to\infty}(\widehat{\mat{P}^\tau})_{ij}&=\frac{\pi_i(\mat{P}^\tau)_{ij}}{\pi_i}\label{eq:MLEPtau_lim}\\
&=(\mat{P}^\tau)_{ij}
\end{align}
Computationally, this approximation is closely related to the limits of $\tau$SFA considered in this paper, and in particular the left normalized formulations due to the normalizing denominators in \cref{eq:MLEPtau,eq:MLEPtau_lim}. One key difference is that the standard definition of the MLE only considers the forward transitions and approximates $\mat{P}^\tau$, while $\tau$SFA equally combines the forward and backward temporal statistics and leads to matrices involving $(\mat{P}^\tau)_{\textup{add}}$ in the case of a Markovian one-hot trajectory. However, a small number of studies use a version of the MLE that additively combines the forward and backward statistics, in analogy to $\tau$SFA \citep{Bowman2009,Buchete2008,Krivov2008,Muff2009}. Moreover, although the MLE typically assumes a non-centered time series $\vec{x}(t)$, the centered case was considered by \citet{Anderson1989}.

\subsection{Future perspectives}
\label{sec:DiscFuture}

This paper demonstrates a computationally novel way to generate SR, namely using SFA. This is an important finding since minimizing slowness, or equivalently maximizing temporal correlation, is a biologically inspired objective \citep{Wiskott1998b,Wiskott2002}. Thus, one potential future development of the current work would be to explore slowness as a principle for designing biologically plausible learning mechanisms of SR. Note that this is an active and ongoing line of research \citep[for a review, see][]{Carvalho2024b}, with recent studies indicating that SR can be learnt using spiking \citep{Bono2023,George2023}, rate-based \citep{Keck2024}, and recurrent neural networks \citep{Fang2023}, as well as latent variable models \citep{Vertes2019}. Note also that in \citep{Lee2022}, a biological interpretation of the TD learning rule for SR was provided using insights from hippocampal research. Considering SFA-based methods might offer an interesting contribution to this family of SR learning mechanisms.

While SR in its classical definition is restricted to finite MDPs, SFA can essentially be performed on any time series and can flexibly deal with different choices of function spaces. Therefore, SFA is comparatively more adaptable than SR to general RL tasks beyond finite MDPs. This underscores a key limitation of the current paper, namely, that the input to SFA is restricted to a Markovian one-hot trajectory. While this choice enables the main results of the paper to be derived, it also limits the scope of the connection established between SFA and SR. It is worth noting that some generalizations of SR beyond finite MDPs have been made, which are typically referred to as \textit{successor features} \citep[for a review, see][]{Carvalho2024b}. Therefore, another future line of work could investigate how, if at all, the application of SFA to a general Markovian time series $\vec{x}(t)$ using non-linear function spaces is related to these methods.

In both \citep{Franzius2007a} and \citep{Stachenfeld2014,Stachenfeld2017}, comparisons are made between the grid-fields generated by the models and grid-cells in the entorhinal cortex. One major disagreement between the two is that the firing maps of grid-cells in the entorhinal cortex exhibit hexagonal symmetry, while the grid-fields generated by either SFA or SR have a symmetry that depends on the shape of the environment's boundary \citep{Stachenfeld2017}. As an example, consider the slow features visualized in \cref{sec:ExpSim} or in \citep{Franzius2007a}, both of which have rectangular symmetry due to the environment being rectangular. One way to understand this property is that slow features defined across an environment, or equivalently the eigenvectors of the associated random walk, resemble eigenfunctions of a Laplace operator, i.e.\ Fourier modes, whose symmetry properties are determined by the boundaries of the system. While some studies indicate that the symmetry of grid-cells in the entorhinal cortex can adapt to match the geometry of environmental boundaries \citep{Krupic2015}, an interesting question is whether SFA or SR can be modified in such a way that they naturally produce hexagonal grid-fields. The authors envisage two directions along which this question could be tackled. Firstly, in the context of SFA, one potential way to reduce the contribution of spatial boundaries in the optimization problem would be to define the objective and constraints locally instead of globally over time, i.e.\ based on averaging within some time window rather than across the entire time series. Secondly, one important observation made in other grid-field models is that in cases where rectangular symmetry is preferred, the inclusion of an additional constraint that forces representations to be non-negative can switch this preference to hexagonal symmetry \citep{Dordek2016,Sorscher2019,Sorscher2023}. Therefore, future research could either explore the integration of such a constraint into SFA, or consider alternative matrix decompositions of SR that produce non-negative representations, and in either case a meaningful question would be whether such adaptations lead to similar representations that instead have hexagonal symmetry. If so, the resulting representations would be a better model of grid-cells found in the brain not only because of the symmetry, but also because non-negative representations are generally better at modelling the firing activity of neurons, which by definition cannot be negative. Note also that in the case of SFA, non-negativity can only be introduced in the Type 2 setting, since the zero mean constraint present in Type~1 SFA prohibits strictly non-negative outputs.


\section{Acknowledgements}
The authors would like to sincerely thank Tobias Glasmachers for his valuable insight on the evaluation of some of the limits in \cref{sec:LimRes}, as well as Merlin Schüler and Janis Keck for several useful discussions regarding some of the general concepts in the paper. We also wish to acknowledge that large language models have been used as a tool for brainstorming and internet searching.

\begin{appendices}

\section{Linear Algebra}
\label{app:EVall}
\subsection{Eigenvalue problems}
\label{app:EVprob}

Given a real square matrix $\mat{A}\in\mathbb{R}^{N\times N}$, an eigenvector is any vector $\vec{w}\in\mathbb{C}^{N}$ that when multiplied by $\mat{A}$ produces a new vector that is simply $\vec{w}$ scaled by some constant $\lambda\in\mathbb{C}$ \citep{Banerjee2014}. If such a vector is a column vector, this can be expressed as
\begin{equation}
\label{eq:EigVecR}
\mat{A}\vec{w}=\lambda\vec{w}
\end{equation}
and $\vec{w}$ is known as a \textit{right eigenvector} of $\mat{A}$. If instead the vector is a row vector, then the corresponding equation is
\begin{equation}
\label{eq:EigVecL}
\vec{w}^T\mat{A}=\lambda\vec{w}^T
\end{equation}
and $\vec{w}$ is known as a \textit{left eigenvector} of $\mat{A}$. In \cref{eq:EigVecR,eq:EigVecL}, the scaling factor $\lambda$ is known as the \textit{eigenvalue} associated to the eigenvector, and the problem of finding solutions to these equations is referred to as an \textit{eigenvalue problem}. For a given matrix $\mat{A}$, there can be at most $N$ linearly independent right eigenvectors. If a full set of such eigenvectors exists, then \cref{eq:EigVecR} can be written for all these eigenvectors simultaneously as
\begin{equation}
\label{eq:EigVecsR}
\mat{A}\mat{W}=\mat{W}\mat{\Lambda}
\end{equation}
where $\mat{W}\in\mathbb{C}^{N\times N}$ contains the vectors as columns and $\mat{\Lambda}\in\mathbb{C}^{N\times N}$ is a diagonal matrix containing the corresponding eigenvalues. Since the columns of $\mat{W}$ are linearly independent, the inverse $\mat{W}^{-1}$ is well-defined and multiplying \cref{eq:EigVecsR} in various ways with this matrix provides insights on the relationship between $\mat{A}$ and $\mat{W}$. Firstly, multiplying \cref{eq:EigVecsR} from the left by $\mat{W}^{-1}$ gives
\begin{equation}
\label{eq:EigBasisDiag}
\mat{W}^{-1}\mat{A}\mat{W}=\mat{\Lambda}
\end{equation}
which says that the matrix $\mat{A}$ is a diagonal transformation when expressed in the basis $\mat{W}$, or in technical terms that $\mat{A}$ is \textit{diagonalizable}. Secondly, multiplying \cref{eq:EigBasisDiag} from the right by $\mat{W}^{-1}$ gives
\begin{equation}
\label{eq:EigVecsL}
\mat{W}^{-1}\mat{A}=\mat{\Lambda}\mat{W}^{-1}
\end{equation}
which says that the rows of $\mat{W}^{-1}$ are a basis of left eigenvectors of $\mat{A}$. Lastly, multiplying \cref{eq:EigVecsR} from the right by $\mat{W}^{-1}$ gives
\begin{equation}
\label{eq:EigDecomp}
\mat{A}=\mat{W}\mat{\Lambda}\mat{W}^{-1}
\end{equation}
\cref{eq:EigDecomp} expresses the matrix $\mat{A}$ purely in terms of its eigenvalue and eigenvector matrices, and for this reason is often referred to as the \textit{eigendecomposition} of $\mat{A}$ \citep{Banerjee2014}.

Not all matrices are diagonalizable. However, \textit{the spectral theorem} states that this property holds for all symmetric real matrices and describes a number of properties belonging to the eigenvalues and eigenvectors. In particular, if $\mat{A}$ is a real symmetric matrix, then the eigenvalues are real and the eigenvectors can be chosen to be both real and orthonormal \citep{Banerjee2014}. If the eigenvectors are chosen in this way, i.e.\ $\mat{W}\in\mathbb{R}^{N\times N}$ and $\mat{W}^{-1}=\mat{W}^T$, then \cref{eq:EigDecomp} simplifies to
\begin{equation}
\label{eq:EigDecompSym}
\mat{A}=\mat{W}\mat{\Lambda}\mat{W}^T
\end{equation}
and the left and right eigenvectors are the same \citep{Seabrook2023}.

If a symmetric matrix has only positive eigenvalues, then it is called \textit{positive-definite} \citep{Banerjee2014}. For such a matrix, the inverse can be written as
\begin{align}
\mat{A}^{-1}&=(\mat{W}\mat{\Lambda}\mat{W}^T)^{-1}\\
&=\mat{W}\mat{\Lambda}^{-1}\mat{W}^T\label{eq:InvEigDecompPD}
\end{align}
where $\mat{\Lambda}^{-1}$ contains the inverses of the diagonals of $\mat{\Lambda}$. Comparing \cref{eq:InvEigDecompPD} to \cref{eq:EigDecompSym}, one sees that $\mat{A}^{-1}$ has the same eigenvectors as $\mat{A}$ and its eigenvalues are simply the reciprocals of the eigenvalues of $\mat{A}$, and that therefore $\mat{A}^{-1}$ is also positive definite.

If the eigenvalues of $\mat{A}$ are instead non-negative, so that $\lambda=0$ is allowed, the matrix is called \textit{positive semi-definite} and $\mat{A}^{-1}$ is not well-defined.

\subsection{Square root and inverse square root of a positive definite matrix}
\label{app:SqrtPSD}

Let $\mat{B}\in\mathbb{R}^{N\times N}$ be a symmetric positive definite matrix. If $\mat{U}$ is an orthonormal set of eigenvectors for this matrix and $\mat{D}$ is a diagonal matrix containing the corresponding real and positive eigenvalues, then $\mat{B}=\mat{UDU}^T$ as in \cref{eq:EigDecompSym}. The square root of $\mat{B}$ can be defined as some matrix $\mat{C}$ that satisfies the following equation:
\begin{equation}
\label{eq:matroot}
\mat{B}=\mat{C}\mat{C}=\mat{C}^2
\end{equation}
Solutions to this equation can be constructed using the eigenvector and eigenvalue matrices $\mat{U}$ and $\mat{D}$, respectively \citep{Banerjee2014}. In particular, consider the matrix $\mat{B}^{\frac{1}{2}}=\mat{U}\mat{D}^{\frac{1}{2}}\mat{U}^T$ where $\mat{D}^{\frac{1}{2}}$ contains the square roots of the eigenvalues of $\mat{B}$. This matrix clearly satisfies \cref{eq:matroot}:
\begin{align}
\mat{B}^{\frac{1}{2}}\mat{B}^{\frac{1}{2}}&=\mat{U}\mat{D}^{\frac{1}{2}}\underbrace{\mat{U}^T\mat{U}}_{=\mat{\mathbbm{1}}}\mat{D}^{\frac{1}{2}}\mat{U}^T\label{eq:matroot_pos_first}\\
&=\mat{U}\underbrace{\mat{D}^{\frac{1}{2}}\mat{D}^{\frac{1}{2}}}_{\mat{D}}\mat{U}^T\\
&=\mat{UDU}^T\\
&=\mat{B}\label{eq:matroot_pos_last}
\end{align}
Two similarities of $\mat{B}$ and $\mat{B}^{\frac{1}{2}}$ are worth mentioning. Firstly, they are simultaneously diagonalizable by the basis $\mat{U}$, and the eigenvalues of $\mat{B}^{\frac{1}{2}}$ are the square roots of the eigenvalues of $\mat{B}$. Secondly, like $\mat{B}$, the matrix $\mat{B}^{\frac{1}{2}}$ is symmetric:
\begin{align}
(\mat{B}^{\frac{1}{2}})^T&=(\mat{U}\mat{D}^{\frac{1}{2}}\mat{U}^T)^T\label{eq:matroot_sym_first}\\
&=\mat{U}\mat{D}^{\frac{1}{2}}\mat{U}^T\\
&=\mat{B}^{\frac{1}{2}}\label{eq:matroot_sym_last}
\end{align}
One difference is that $\mat{B}^{\frac{1}{2}}$ is not necessarily positive definite. To see this, note that each eigenvalue of $\mat{B}$ has both a positive and negative square root, meaning that there are in fact $2^n$ choices for the matrix $\mat{D}^{\frac{1}{2}}$ where $n$ is the number of distinct eigenvalues. The unique choice for which $\mat{B}^{\frac{1}{2}}$ is positive definite is when only the positive roots are chosen, in which case $\mat{B}^{\frac{1}{2}}$ is referred to as the \textit{positive square root} of $\mat{B}$. Throughout this paper it is assumed that the square root of a positive diagonal matrix, such as $\mat{D}^{\frac{1}{2}}$, always uses the positive roots only.

Since $\mat{B}$ is positive definite, it has an inverse $\mat{B}^{-1}=\mat{U}\mat{D}^{-1}\mat{U}^T$ that is also positive definite. Therefore, like $\mat{B}$, the matrix $\mat{B}^{-1}$ has a positive square root $(\mat{B}^{-1})^{\frac{1}{2}}=\mat{U}\mat{D}^{-\frac{1}{2}}\mat{U}^T$ that satisfies \cref{eq:matroot} but for $\mat{B}^{-1}$:
\begin{align}
(\mat{B}^{-1})^{\frac{1}{2}}(\mat{B}^{-1})^{\frac{1}{2}}&=\mat{U}\mat{D}^{-\frac{1}{2}}\underbrace{\mat{U}^T\mat{U}}_{=\mat{\mathbbm{1}}}\mat{D}^{-\frac{1}{2}}\mat{U}^T\label{eq:matrootinv_pos_first}\\
&=\mat{U}\underbrace{\mat{D}^{-\frac{1}{2}}\mat{D}^{-\frac{1}{2}}}_{\mat{D}^{-1}}\mat{U}^T\\
&=\mat{U}\mat{D}^{-1}\mat{U}^T\\
&=\mat{B}^{-1}\label{eq:matrootinv_pos_last}
\end{align}
Note that $(\mat{B}^{-1})^{\frac{1}{2}}$ is also the inverse of $\mat{B}^{\frac{1}{2}}$ and can therefore be thought of as the inverse square root of $\mat{B}$. For this reason $\mat{B}^{-\frac{1}{2}}$ is used as a shorthand to denote this matrix. Moreover, following the same steps as in \Crefrange{eq:matroot_sym_first}{eq:matroot_sym_last} it can be shown that $\mat{B}^{-\frac{1}{2}}$ is symmetric, and like with $\mat{B}^{\frac{1}{2}}$ it is assumed that $\mat{D}^{-\frac{1}{2}}$ contains only the positive inverse square roots, so that $\mat{B}^{-\frac{1}{2}}$ is positive definite.

An alternative and less commonly used way to define the square root of $\mat{B}$ is some matrix $\mat{C}$ that satisfies \citep{Banerjee2014}:
\begin{equation}
\label{eq:matrootalt}
\mat{B}=\mat{C}\mat{C}^T
\end{equation}
\cref{eq:matrootalt} generalizes \cref{eq:matroot}, since for the special case where $\mat{C}$ is symmetric the two equations are the same. Therefore, one possible solution to \cref{eq:matrootalt} is the positive square root introduced above. However, multiplying the positive square root from the right by some orthogonal matrix $\mat{Q}^T$ produces a matrix $\mat{B}^{\frac{1}{2}}=\mat{U}\mat{D}^{\frac{1}{2}}\mat{U}^T\mat{Q}^T$ that also satisfies \cref{eq:matrootalt}:
\begin{align}
\mat{B}^{\frac{1}{2}}(\mat{B}^{\frac{1}{2}})^T&=(\mat{U}\mat{D}^{\frac{1}{2}}\mat{U}^T\mat{Q}^T)(\mat{U}\mat{D}^{\frac{1}{2}}\mat{U}^T\mat{Q}^T)^T\\
&=\mat{U}\mat{D}^{\frac{1}{2}}\mat{U}^T\underbrace{\mat{Q}^T\mat{Q}}_{=\mat{\mathbbm{1}}}\mat{U}\mat{D}^{\frac{1}{2}}\mat{U}^T\\
&=\mat{U}\mat{D}^{\frac{1}{2}}\mat{U}^T\mat{U}\mat{D}^{\frac{1}{2}}\mat{U}^T\\
&=\mat{B}\label{eq:matrootalt_orthog}
\end{align}
where the last step in \cref{eq:matrootalt_orthog} can be made using the same arguments as in \Crefrange{eq:matroot_pos_first}{eq:matroot_pos_last}. Thus, this definition of $\mat{B}^{\frac{1}{2}}$ defines a family of solutions to \cref{eq:matrootalt} that are parameterized by $\mat{Q}$, with $\mat{Q}=\mat{\mathbbm{1}}$ corresponding to the positive square root.

For any choice of $\mat{Q}$, the corresponding inverse square root can be found by inverting $\mat{B}^{\frac{1}{2}}$
\begin{align}
\mat{B}^{-\frac{1}{2}}&=(\mat{B}^{\frac{1}{2}})^{-1}\\
&=(\mat{U}\mat{D}^{\frac{1}{2}}\mat{U}^T\mat{Q}^T)^{-1}\\
&=\mat{Q}\mat{U}\mat{D}^{-\frac{1}{2}}\mat{U}^T
\end{align}
which also simplifies to the positive square root for $\mat{Q}=\mat{\mathbbm{1}}$. As before, the matrix $\mat{B}^{-\frac{1}{2}}$ can be thought of as a square root of $\mat{B}^{-1}$, and indeed this matrix satisfies the analogue of \cref{eq:matrootalt_orthog} for $\mat{B}^{-1}$, but with the difference that the order of transposing is swapped:
\begin{align}
(\mat{B}^{-\frac{1}{2}})^T\mat{B}^{-\frac{1}{2}}&=(\mat{Q}\mat{U}\mat{D}^{-\frac{1}{2}}\mat{U}^T)^T(\mat{Q}\mat{U}\mat{D}^{-\frac{1}{2}}\mat{U}^T)^T\\
&=\mat{U}\mat{D}^{-\frac{1}{2}}\mat{U}^T\underbrace{\mat{Q}^T\mat{Q}}_{=\mat{\mathbbm{1}}}\mat{U}\mat{D}^{-\frac{1}{2}}\mat{U}^T\\
&=\mat{U}\mat{D}^{-\frac{1}{2}}\mat{U}^T\mat{U}\mat{D}^{-\frac{1}{2}}\mat{U}^T\\
&=\mat{B}^{-1}\label{eq:matrootaltinv_orthog}
\end{align}
where the last step in \cref{eq:matrootaltinv_orthog} can be made using the same arguments as in \Crefrange{eq:matrootinv_pos_first}{eq:matrootinv_pos_last}.

In this paper, the square root and inverse square root of a positive definite matrix $\mat{B}$ are interpreted in the more general sense involving $\mat{Q}$, while maintaining the assumption that $\mat{D}^{\pm\frac{1}{2}}$ contains only the positive square roots of the eigenvalues of $\mat{B}$. It is worth noting that for this definition there is no guarantee that $\mat{B}^{\pm\frac{1}{2}}$ is symmetric for $\mat{Q}\neq\mat{\mathbbm{1}}$.

\subsection{Generalized eigenvalue problems}
\label{app:GenEVprob}

The eigenvalue problem described in \cref{eq:EigVecR,eq:EigVecL} can be extended by considering an ordered pair of matrices $(\mat{A},\mat{B})$ and a vector $\vec{w}$ such that
\begin{equation}
\label{eq:GenEigVecR}
\mat{A}\vec{w}=\lambda\mat{B}\vec{w}
\end{equation}
or
\begin{equation}
\label{eq:GenEigVecL}
\vec{w}^T\mat{A}=\lambda\vec{w}^T\mat{B}
\end{equation}
Clearly, for $\mat{B}=\mat{\mathbbm{1}}$ these problems reduce to the ones described in \cref{eq:EigVecR,eq:EigVecL}, respectively. For this reason, they are known as \textit{generalized eigenvalue problems}, with the vector $\vec{w}$ being a \textit{generalized right/left eigenvector} of $(\mat{A},\mat{B})$ and $\lambda$ being the corresponding \textit{generalized eigenvalue} \citep{Banerjee2014}. Moreover, if $(\mat{A},\mat{B})$ have $N$ linearly independent generalized right eigenvectors, \cref{eq:GenEigVecR} can be expressed as
\begin{equation}
\label{eq:GenEigVecsR}
\mat{AW}=\mat{BW\Lambda}
\end{equation}
which is the generalized analogue of \cref{eq:EigVecsR}.

Generalized eigenvalue problems are often encountered in applications of linear algebra. They can either be solved directly using various numerical techniques, or transformed into regular eigenvalue problems, which are often computationally easier to solve. A special case that has received a lot of attention in the literature, and also the one most relevant to this paper, is where $\mat{A}$ is symmetric and $\mat{B}$ is positive definite. Converting this problem to a regular eigenvalue problem has the benefit that a number of insights regarding symmetric matrices can be utilized. One way to make this conversion uses the fact that if $\mat{B}$ is positive definite then it has a square root and inverse square root given by $\mat{B}^{\frac{1}{2}}=\mat{U}\mat{D}^{\frac{1}{2}}\mat{U}^T\mat{Q}^T$ and $\mat{B}^{-\frac{1}{2}}=\mat{Q}\mat{U}\mat{D}^{-\frac{1}{2}}\mat{U}^T$, respectively (see Appendix \ref{app:SqrtPSD}). Using these matrices, together with the fact that $\mat{\mathbbm{1}}=\mat{B}^{\frac{1}{2}}\mat{B}^{-\frac{1}{2}}=(\mat{B}^{\frac{1}{2}}\mat{B}^{-\frac{1}{2}})^T=(\mat{B}^{-\frac{1}{2}})^T(\mat{B}^{\frac{1}{2}})^T$, it is possible to convert \cref{eq:GenEigVecsR} into a regular eigenvalue problem as follows \citep{Banerjee2014}:
\begin{align}
& &\mat{A}\underbrace{(\mat{B}^{-\frac{1}{2}})^T(\mat{B}^{\frac{1}{2}})^T}_{=\mat{\mathbbm{1}}}\mat{W}&=\mat{B}\mat{W\Lambda}&\text{(insertion of identity)}\label{eq:GenEVSymSN1}\\
&\Longleftrightarrow &\mat{B}^{-\frac{1}{2}}\mat{A}(\mat{B}^{-\frac{1}{2}})^T(\mat{B}^{\frac{1}{2}})^T\mat{W}&=\mat{B}^{-\frac{1}{2}}\mat{B}\mat{W\Lambda}&\text{(multiplication from left by $\mat{B}^{-\frac{1}{2}}$)}\\
& & &=(\mat{B}^{\frac{1}{2}})^T\mat{W\Lambda}&\text{(because $\mat{B}=\mat{B}^{\frac{1}{2}}(\mat{B}^{\frac{1}{2}})^T$)}\\
&\Longleftrightarrow &\mat{B}^{-\frac{1}{2}}\mat{A}(\mat{B}^{-\frac{1}{2}})^T\widetilde{\mat{W}}&=\widetilde{\mat{W}}\mat{\Lambda}&\text{($\widetilde{\mat{W}}=(\mat{B}^{\frac{1}{2}})^T\mat{W}$)}\label{eq:GenEVSymSN2}
\end{align}
\cref{eq:GenEVSymSN2} describes a regular eigenvalue problem of the matrix $\mat{B}^{-\frac{1}{2}}\mat{A}(\mat{B}^{-\frac{1}{2}})^T$, which is guaranteed to be symmetric since $(\mat{B}^{-\frac{1}{2}}\mat{A}(\mat{B}^{-\frac{1}{2}})^T)^T=\mat{B}^{-\frac{1}{2}}\mat{A}^T(\mat{B}^{-\frac{1}{2}})^T=\mat{B}^{-\frac{1}{2}}\mat{A}(\mat{B}^{-\frac{1}{2}})^T$. For this reason, the conversion in \Crefrange{eq:GenEVSymSN1}{eq:GenEVSymSN2} is referred to as \textit{symmetric normalization} in this paper, and the symmetry of the resulting matrix provides two key insights about the starting generalized eigenvalue problem. Firstly, the eigenvalues and eigenvectors in \cref{eq:GenEVSymSN2} are guaranteed to be real by virtue of the spectral theorem. Since the eigenvalues are not modified by the normalization process this means the same is true of the generalized eigenvalues, and since $\widetilde{\mat{W}}=(\mat{B}^{\frac{1}{2}})^T\mat{W}$ (with $(\mat{B}^{\frac{1}{2}})^T$ also being real) the same is true of the generalized eigevectors $\mat{W}$. Secondly, the spectral theorem also says that the eigenvectors in \cref{eq:GenEVSymSN2} can be chosen to be orthonormal w.r.t\ the standard Euclidean inner product, in which case \ $\widetilde{\mat{W}}^T\widetilde{\mat{W}}=\mat{\mathbbm{1}}$. Using this fact, and transforming the basis $\widetilde{\mat{W}}$ back into $\mat{W}$ gives:
\begin{align}
\widetilde{\mat{W}}^T\widetilde{\mat{W}}&=\mat{W}^T(\mat{B}^{\frac{1}{2}})(\mat{B}^{\frac{1}{2}})^T\mat{W}\\
&=\mat{W}^T\mat{B}\mat{W}\\
&=\mat{\mathbbm{1}}\label{eq:GenEVOrthog}
\end{align}
Assuming that $\vec{w}_i$ and $\vec{w}_j$ denote columns of $\mat{W}$, then the rightmost part of \cref{eq:GenEVOrthog} can alternatively be written in an indexed form as
\begin{equation}
\label{eq:GenEVOrthogInd}
\mat{w}_i^T\mat{B}\mat{w}_j=(\mat{w}_i , \mat{w}_j)_{\mat{B}}=\delta_{ij}
\end{equation}
where $\delta_{ij}$ is the delta Kronecker symbol. \cref{eq:GenEVOrthogInd} is particularly informative because it says that the generalized eigenvectors $\mat{W}$ of the starting problem are orthogonal w.r.t.\ the weighted inner product $(\cdot , \cdot)_{\mat{B}}$.

The procedure described above is not the only way to convert \cref{eq:GenEigVecsR} into a regular eigenvalue problem when $\mat{A}$ is symmetric and $\mat{B}$ is positive definite. An alternative method makes use of the inverse $\mat{B}^{-1}$, which is guaranteed to exist since by assumption $\mat{B}$ has no eigenvalues of zero. Multiplying \cref{eq:GenEigVecsR} from the left by $\mat{B}^{-1}$ gives \citep{Ghojogh2023}
\begin{align}
\label{eq:GenEVSymLN}
\mat{B}^{-1}\mat{A}\mat{W}&=\mat{W\Lambda}
\end{align}
which is referred to as \textit{left normalization} in this paper. Note that unlike $\mat{A}$ and $\mat{B}$, the matrix $\mat{B}^{-1}\mat{A}$ on the left-hand side of \cref{eq:GenEVSymLN} is not necessarily symmetric. Nonetheless, since neither $\mat{\Lambda}$ nor $\mat{W}$ are changed by this normalization, the eigenvalues and eigenvectors of $\mat{B}^{-1}\mat{A}$ are the same as the generalized counterparts in \cref{eq:GenEigVecsR}. Thus, $\mat{B}^{-1}\mat{A}$ has real eigenvalues and the eigenvectors can be chosen to be real and orthogonal in the sense decribed by \cref{eq:GenEVOrthogInd}.

Note that both the normalization methods presented above only work because $\mat{B}$ is positive definite by assumption. Another setting that has received a lot of interest in the literature \citep{Zoltowski1987,Hochstenbach2019,Hochstenbach2023,Hochstenbach2024,Ghojogh2023}, and which is also relevant to the current paper, is where $\mat{A}$ is again symmetric and $\mat{B}$ is positive semi-definite. In this case, symmetric and left normalization cannot be applied directly since they involve computing $\mat{B}^{-\frac{1}{2}}$ and $\mat{B}^{-1}$, respectively, neither of which are well-defined when $\mat{B}$ has one or more eigenvalues of zero. One way to generalize the normalization methods to this case is through a numerical hack in which $\mat{B}$ is perturbed along the diagonals by a small amount $\epsilon$, i.e.\
\begin{equation}
\mat{B}\to\mat{B}_\epsilon=\mat{B}+\epsilon\mat{\mathbbm{1}}\label{eq:QuickDirty}
\end{equation}
which is referred to by \citet{Ghojogh2023} as the \textit{quick and dirty solution} in the context of left normalization. In order to understand the matrix $\mat{B}_\epsilon$, note that if $\mat{B}$ is positive semi-definite then it is diagonalizable and that like all matrices it commutes with the identity matrix. Therefore, $\mat{B}$ and $\mat{\mathbbm{1}}$ share an eigenbasis and the perturbation simply increases all eigenvalues of $\mat{B}$ by $\epsilon$, so that any zero eigenvalues become positive and $\mat{B}_\epsilon$ is positive definite. Therefore, replacing $\mat{B}$ by $\mat{B}_\epsilon$ in \cref{eq:GenEigVecsR} produces a generalized eigenvalue problem that can be solved with the same basis $\mat{W}$ and which can be converted to a regular eigenvalue problem using either symmetric or left normalization. It is worth noting that matrices having eigenvalues numerically close to zero, known as \textit{ill-conditioned matrices}, can be practically just as problematic as having eigenvalues equal to zero, since normalization involves computing either $\mat{B}^{-\frac{1}{2}}$ or $\mat{B}^{-1}$, both of which can have very large round-off errors if $\mat{B}$ is ill-conditioned. Therefore, the parameter $\epsilon$ in \cref{eq:QuickDirty} needs to be large enough to avoid this issue. On the other hand, adding too much noise runs the risk erasing important information in $\mat{B}$ that describes whichever system is being modelled. For these reasons, when applying this trick it is important to take care in selecting a suitable value of $\epsilon$.

\section{SR}
\subsection{Definition of SR}
\label{app:SRsum}
\textbf{\cref{thm:SRinv+Neumann}} \textit{The matrix $\mat{M}$ is well-defined and can alternatively be expressed as
\begin{equation}
\mat{M}=\sum_{k=0}^{\infty}\gamma^k \mat{P}^k
\end{equation}
which is known as a Neumann series.}
\begin{proof}
In order to show that $\mat{M}$ is well-defined, note that $\mat{P}$ is a transition matrix and therefore each of its eigenvalues $\lambda$ is bounded by $|\lambda|\leq 1$ \citep{Seabrook2023}. Together with the fact that $\gamma\in(0,1)$, this means that the matrix $\gamma\mat{P}$ has eigenvalues bounded by $|\gamma\lambda|<1$, and therefore cannot have an eigenvalue of $1$. By extension, this means that the matrix $\mat{\mathbbm{1}} - \gamma \mat{P}$ cannot have an eigenvalue of zero, and therefore must be invertible.

As a consequence of the argument above, the matrix $\gamma\mat{P}$ has a spectral radius $\rho(\gamma\mat{P})<1$. This is true if and only if the Neumann series
\begin{equation}
\sum_{k=0}^{\infty}\gamma^k \mat{P}^k=\mat{\mathbbm{1}}+\gamma\mat{P}+\gamma^2\mat{P}^2+\gamma^3\mat{P}^3+\cdots
\end{equation}
converges and is equal to $(\mat{\mathbbm{1}} - \gamma \mat{P})^{-1}$ \citep[see][p.\ 618]{Meyer2000}.
\end{proof}

\subsection{Eigenvalues and Eigenvectors of SR - general case}
\label{app:SREVgen}
\noindent\textbf{\cref{thm:eigSR}} \textit{For a general Markov chain, if $\vec{w}$ is an eigenvector of $\mat{P}$ with eigenvalue $\lambda\in\mathbb{C}$, then it is also an eigenvector of $\mat{P}^k$ and $\mat{M}$ with eigenvalues $\lambda^k$ and $\frac{1}{1-\gamma\lambda}$, respectively, where $k>1$.}
\begin{proof}
Let $\vec{w}$ be a right eigenvector of $\mat{P}$ with eigenvalue $\lambda$, i.e.\
\begin{equation}
\mat{P}\vec{w}=\lambda\vec{w}\label{eq:Peigvec}
\end{equation}
Then multiplying $\vec{w}$ from the left by $\mat{P}^k$ gives
\begin{align}
\mat{P}^k\vec{w}&=\underbrace{\mat{P}\cdot \mat{P} \cdots \mat{P}}_{k\text{ times}}\vec{w}\\
&=\underbrace{\mat{P}\cdot \mat{P} \cdots \mat{P}}_{k-1\text{ times}}\mat{P}\vec{w}\\
\overset{(\ref{eq:Peigvec})}&{=}\underbrace{\mat{P}\cdot \mat{P} \cdots \mat{P}}_{k-1\text{ times}}\lambda\vec{w}
\end{align}
and repeating the same steps a further $k-1$ times leads to
\begin{equation}
\mat{P}^k\vec{w}=\lambda^k\vec{w}\label{eq:Pkeigvec}
\end{equation}
Moreover, the action of $\mat{M}$ on $\vec{w}$ is given by
\begin{align}
\mat{M}\vec{w}\overset{(\ref{eq:SRmat})}&{=}\bigg(\sum_{k=0}^{\infty}\gamma^k \mat{P}^k\bigg)\vec{w}\\
&=\sum_{k=0}^{\infty}\gamma^k \mat{P}^k\vec{w}\\
\overset{(\ref{eq:Pkeigvec})}&{=}\sum_{k=0}^{\infty}\gamma^k \lambda^k\vec{w}\\
&=\bigg(\sum_{k=0}^{\infty}\gamma^k \lambda^k\bigg)\vec{w}\\
&=\frac{1}{1-\gamma\lambda}\vec{w}\label{eq:eigSR1}
\end{align}
In the last line, the geometric series is expressed in closed form. This is allowed because $\mat{P}$ is a transition matrix, which means that $|\lambda|\leq1$, and therefore $|\gamma \lambda|<1$. Note that the eigenvalues of $\mat{M}$ are related to those of $\mat{P}$ in such a way that the ordering from largest to smallest is preserved.
\end{proof}

\subsection{Eigenvalues and Eigenvectors of SR - reversible Markov chains}
\label{app:SREVrev}
\noindent \textbf{\cref{thm:Pk+Mrev}} \textit{For a reversible Markov chain with stationary distribution $\vec{\pi}$, the $k$-step transition matrix $\mat{P}^k$ and SR matrix $\mat{M}$ satisfy the following properties:
\begin{enumerate}
\item $\mat{\Pi}\mat{P}^k$ and $\mat{\Pi}\mat{M}$ are symmetric,
\item $\mat{\Pi}^{\frac{1}{2}}\mat{P}^k\mat{\Pi}^{-\frac{1}{2}}$ and $\mat{\Pi}^{\frac{1}{2}}\mat{M}\mat{\Pi}^{-\frac{1}{2}}$ are symmetric,
\item $\mat{P}^k$ and $\mat{M}$ are simultaneously diagonalizable with real eigenvalues and eigenvectors,
\item The left and right eigenvectors of $\mat{P}^k$ and $\mat{M}$ can be chosen to be orthogonal w.r.t\ to the weighted inner products $\langle \cdot, \cdot\rangle_{\mat{\Pi}^{-1}}$ and $\langle \cdot, \cdot\rangle_{\mat{\Pi}}$, respectively,
\end{enumerate}
where $\mat{\Pi}=\textup{diag}(\vec{\pi})$ and $k\geq 1$.}
\begin{proof}
First consider the symmetry of $\mat{\Pi P}^k$. For $k=1$ this is guaranteed because the starting Markov chain is reversible (see \cref{thm:Prev}), i.e.\
\begin{equation}
(\mat{\Pi P} )^T=\mat{P}^T\mat{\Pi}=\mat{\Pi P}\label{eq:DBmat}
\end{equation}
For $k>1$ the symmetry of $\mat{\Pi}\mat{P}^k$ can be established by first considering the following:
\begin{align}
(\mat{\Pi P}^k)^T&=(\mat{P}^k)^T\mat{\Pi}\\
&=(\underbrace{\mat{P}\cdot \mat{P} \cdots \mat{P}}_{k\text{ times}})^T\mat{\Pi}\\
&=\underbrace{\mat{P}^T\cdot \mat{P}^T \cdots \mat{P}^T }_{k\text{ times}}\mat{\Pi}\label{eq:Pkrev_third}\\
&=\underbrace{\mat{P}^T\cdot \mat{P}^T \cdots \mat{P}^T}_{k-1\text{ times}}\mat{P}^T\mat{\Pi}\label{eq:Pkrev_fourth}\\
\overset{(\ref{eq:DBmat})}&{=}\underbrace{\mat{P}^T\cdot \mat{P}^T \cdots \mat{P}^T}_{k-1\text{ times}}\mat{\Pi P}\label{eq:Pkrev_fifth}
\end{align}
Then, repeating the steps from \cref{eq:Pkrev_third} to \cref{eq:Pkrev_fifth} a further $k-1$ times leads to
\begin{equation}
(\mat{\Pi P}^k)^T=\mat{\Pi P}^k\label{eq:Pkrev_sixth}
\end{equation}
meaning that $\mat{\Pi P}^k$ is symmetric. Note that the symmetry of $\mat{\Pi P}^k$ implies that the $k$-step Markov chain is reversible \citep{Seabrook2023}. Therefore, by \cref{thm:Prev}, $\mat{\Pi}^{\frac{1}{2}}\mat{P}^k\mat{\Pi}^{-\frac{1}{2}}$ is symmetric, $\mat{P}^k$ has real eigenvalues and eigenvectors, and the eigenvectors of $\mat{P}^k$ satisfy the orthogonality relations described.

By \cref{thm:eigSR}, $\mat{P}^k$ and $\mat{M}$ are simultaneously diagonalizable. Therefore, since the eigenvalues and eigenvectors of $\mat{P}^k$ are real, so too are those of $\mat{M}$, and by the same reasoning the eigenvectors of $\mat{M}$ satisfy the orthogonality relations described. Therefore, it only remains to consider the symmetry of $\mat{\Pi M}$ and $\mat{\Pi}^{\frac{1}{2}}\mat{M}\mat{\Pi}^{-\frac{1}{2}}$.

The matrix $\mat{\Pi M}$ is given by:
\begin{align}
\mat{\Pi M}\overset{(\ref{eq:SRmat})}&{=}\mat{\Pi}\sum_{k=0}^{\infty}\gamma^k \mat{P}^k\\
&=\sum_{k=0}^{\infty}\gamma^k \mat{\Pi P}^k\label{eq:SRflow}
\end{align}
Note that each matrix $\mat{\Pi P}^k$ in \cref{eq:SRflow} is non-negative. Such matrices are subject to a well-known result formulated by Frobenius, which says that their spectral radius is less than or equal to the largest row sum \citep{Minc1988,Baboulakis2020}. For $\mat{\Pi P}^k$, the row sums are simply the stationary probabilities, i.e.\
\begin{equation}
\sum_j  (\mat{\Pi P}^k)_{ij}=\sum_j \pi_i(\mat{P}^k)_{ij}=\pi_i\underbrace{\sum_j (\mat{P}^k)_{ij}}_{=1}=\pi_i
\end{equation}
meaning that the spectral radius of $\mat{\Pi P}^k$ is no greater than the largest stationary probability, and is therefore strictly less than $1$. Thus, since $\gamma$ is also less than $1$, the Neumann series in \cref{eq:SRflow} is guaranteed to converge \citep[see][p.\ 618]{Meyer2000}, and since \cref{eq:Pkrev_sixth} implies that each term in the sum is symmetric then so too is $\mat{\Pi}\mat{M}$.

The matrix $\mat{\Pi}^{\frac{1}{2}}\mat{M}\mat{\Pi}^{-\frac{1}{2}}$ is given by:
\begin{align}
\mat{\Pi}^{\frac{1}{2}}\mat{M}\mat{\Pi}^{-\frac{1}{2}}\overset{(\ref{eq:SRmat})}&{=}\mat{\Pi}^{\frac{1}{2}}\bigg(\sum_{k=0}^{\infty}\gamma^k \mat{P}^k\bigg)\mat{\Pi}^{-\frac{1}{2}}\\
&=\sum_{k=0}^{\infty}\gamma^k\mat{\Pi}^{\frac{1}{2}}\mat{P}^k\mat{\Pi}^{-\frac{1}{2}}\label{eq:Msymtrans}
\end{align}
Note that each term $\mat{\Pi}^{\frac{1}{2}}\mat{P}^k\mat{\Pi}^{-\frac{1}{2}}$ is a similarity transformation on $\mat{P}^k$ and therefore has the same eigenvalues and a spectral radius of $1$ \citep{Seabrook2023}. Thus, the sum in \cref{eq:Msymtrans} converges, and since each term $\mat{\Pi}^{\frac{1}{2}}\mat{P}^k\mat{\Pi}^{-\frac{1}{2}}$ is symmetric then so too is $\mat{\Pi}^{\frac{1}{2}}\mat{M}\mat{\Pi}^{-\frac{1}{2}}$.


\end{proof}

\subsection{Eigenvalues and Eigenvectors of SR - additive reversibilization}
\label{app:SREVadd}
\noindent \textbf{\cref{thm:P+Madd}} \textit{For an ergodic, not necessarily reversible Markov chain with stationary distribution $\vec{\pi}$, the matrices $(\mat{P}^k)_{\textup{add}}$ and $\mat{M}_{\textup{add}}$ satisfy the following properties:
\begin{itemize}
\item $\mat{\Pi}(\mat{P}^k)_{\textup{add}}$ and $\mat{\Pi}\mat{M}_{\textup{add}}$ are symmetric,
\item $\mat{\Pi}^{\frac{1}{2}}(\mat{P}^k)_{\textup{add}}\mat{\Pi}^{-\frac{1}{2}}$ and $\mat{\Pi}^{\frac{1}{2}}\mat{M}_{\textup{add}}\mat{\Pi}^{-\frac{1}{2}}$ are symmetric,
\item $(\mat{P}^k)_{\textup{add}}$ and $\mat{M}_{\textup{add}}$ are diagonalizable with real eigenvalues and eigenvectors,
\item The left and right eigenvectors of $(\mat{P}^k)_{\textup{add}}$ and $\mat{M}_{\textup{add}}$ can be chosen to be orthogonal w.r.t\ to the weighted inner products $\langle \cdot, \cdot\rangle_{\mat{\Pi}^{-1}}$ and $\langle \cdot, \cdot\rangle_{\mat{\Pi}}$, respectively,
\end{itemize}
where $\mat{\Pi}=\textup{diag}(\vec{\pi})$ and $k\geq 1$.}

\begin{proof}
Note that $\mat{P}^k$ describes the $k$-step Markov chain and $(\mat{P}^k)_{\textup{add}}$ the corresponding additive reversibilization. Hence, $(\mat{P}^k)_{\textup{add}}$ describes a reversible Markov chain and thus by \cref{thm:Pk+Mrev} it satisfies all properties listed. Therefore it is sufficient to consider only the points related to $\mat{M}_{\textup{add}}$.

Multiplying $\mat{M}_{\textup{add}}$ from the left by $\mat{\Pi}$ gives:
\begin{align}
\mat{\Pi}\mat{M}_{\textup{add}}\overset{(\ref{eq:Madd2})}&{=}\mat{\Pi}\bigg(\sum_{k=0}^{\infty}\gamma^k  (\mat{P}^k)_{\text{add}}\bigg)\\
&=\sum_{k=0}^{\infty}\gamma^k \mat{\Pi}(\mat{P}^k)_{\text{add}}\label{eq:SRaddflow}
\end{align}
By an equivalent argument as in the previous proof, the sum in \cref{eq:SRaddflow} converges, and since each term $\mat{\Pi}(\mat{P}^k)_{\text{add}}$ inside the sum is symmetric then so too is $\mat{\Pi}\mat{M}_{\textup{add}}$.

The similarity transformation $\mat{\Pi}^{\frac{1}{2}}(\mat{M})_{\textup{add}}\mat{\Pi}^{-\frac{1}{2}}$ is given by:
\begin{align}
\mat{\Pi}^{\frac{1}{2}}\mat{M}_{\textup{add}}\mat{\Pi}^{-\frac{1}{2}}\overset{(\ref{eq:Madd2})}&{=}\mat{\Pi}^{\frac{1}{2}}\bigg(\sum_{k=0}^{\infty}\gamma^k  (\mat{P}^k)_{\text{add}}\bigg)\mat{\Pi}^{-\frac{1}{2}}\\
&=\sum_{k=0}^{\infty}\gamma^k \mat{\Pi}^{\frac{1}{2}}(\mat{P}^k)_{\text{add}}\mat{\Pi}^{-\frac{1}{2}}\label{eq:SRaddsymtrans}
\end{align}
By an equivalent argument as in the previous proof, the sum in \cref{eq:SRaddsymtrans} converges, and since each term $\mat{\Pi}^{\frac{1}{2}}(\mat{P}^k)_{\text{add}}\mat{\Pi}^{-\frac{1}{2}}$ is symmetric then so too is $\mat{\Pi}^{\frac{1}{2}}\mat{M}_{\textup{add}}\mat{\Pi}^{-\frac{1}{2}}$.

Since $\mat{\Pi}^{\frac{1}{2}}\mat{M}_{\textup{add}}\mat{\Pi}^{-\frac{1}{2}}$ is symmetric, it is diagonalizable with real eigenvalues and eigenvectors, and it is always possible to choose the eigenvectors to be orthogonal w.r.t\ the Euclidean inner product \citep{Banerjee2014}. In this case,
\begin{equation}
\mat{\Pi}^{\frac{1}{2}}\mat{M}_{\textup{add}}\mat{\Pi}^{-\frac{1}{2}}=\mat{U}\mat{D}\mat{U}^T\label{eq:PtausymEig}
\end{equation}
where $\mat{D}$ is a diagonal matrix containing the eigenvalues and $\mat{U}$ is an orthogonal matrix containing the eigenvectors. Rearranging this equation gives:
\begin{align}
\mat{M}_{\textup{add}}&=\mat{\Pi}^{-\frac{1}{2}}\mat{U}\mat{D}\mat{U}^T\mat{\Pi}^{\frac{1}{2}}\\
&=(\mat{\Pi}^{-\frac{1}{2}}\mat{U})\mat{D}(\mat{\Pi}^{-\frac{1}{2}}\mat{U})^{-1}\\
&=\widetilde{\mat{U}}\mat{D}\widetilde{\mat{U}}^{-1}\label{eq:PtauEig}
\end{align}
\cref{eq:PtauEig} says that $\mat{M}_{\textup{add}}$ is diagonalizable, where $\mat{D}$ are the eigenvalues and $\widetilde{\mat{U}}^{-1}$ and $\widetilde{\mat{U}}$ are a biorthogonal system of left and right eigenvectors, respectively \citep{Seabrook2023}. In comparison to \cref{eq:PtausymEig} where the matrix on the left-hand side is symmetric and the left and right eigenvectors are the same, in \cref{eq:PtauEig} the matrix is generally non-symmetric, which is why there are distinct left and right eigenvectors. Note that since all matrices of \cref{eq:PtauEig} are real-valued, the eigenvalues and eigenvectors of $\mat{M}_{\textup{add}}$ are real. Note also that while $\mat{M}$ and $\mat{P}^k$ are guaranteed to be simultaneously diagonalizable (see \cref{thm:eigSR}), this generally does not hold for $\mat{M}_{\textup{add}}$ and $(\mat{P}^k)_{\textup{add}}$.

In contrast to $\mat{\Pi}^{\frac{1}{2}}\mat{M}_{\textup{add}}\mat{\Pi}^{-\frac{1}{2}}$, there is no guarantee that $\mat{M}_{\textup{add}}$ is symmetric or that the eigenvector matrices in \cref{eq:PtauEig} are orthogonal w.r.t.\ to the Euclidean inner product. Instead, the matrix $\widetilde{\mat{U}}^{-1}$, which contains the left eigenvectors, satisfies
\begin{align}
(\widetilde{\mat{U}}^{-1})^T\mat{\Pi}^{-1}(\widetilde{\mat{U}}^{-1})&=(\mat{\Pi}^{-\frac{1}{2}}\mat{U})^T\mat{\Pi}(\mat{\Pi}^{-\frac{1}{2}}\mat{U})\\
&=\mat{U}^T\underbrace{\mat{\Pi}^{\frac{1}{2}}\mat{\Pi}^{-1}\mat{\Pi}^{\frac{1}{2}}}_{=\mat{\mathbbm{1}}}\mat{U}\\
&=\mat{U}^T\mat{U}\\
&=\mat{\mathbbm{1}}
\end{align}
which is equivalent to the left eigenvectors being orthogonal w.r.t\ $\langle \cdot, \cdot\rangle_{\mat{\Pi}^{-1}}$. Similarly, the matrix $\widetilde{\mat{U}}$, which contains the right eigenvectors, satisfies
\begin{align}
\widetilde{\mat{U}}^T\mat{\Pi}\widetilde{\mat{U}}&=(\mat{\Pi}^{-\frac{1}{2}}\mat{U})^T\mat{\Pi}(\mat{\Pi}^{-\frac{1}{2}}\mat{U})\\
&=\mat{U}^T\underbrace{\mat{\Pi}^{-\frac{1}{2}}\mat{\Pi}\mat{\Pi}^{-\frac{1}{2}}}_{=\mat{\mathbbm{1}}}\mat{U}\\
&=\mat{U}^T\mat{U}\\
&=\mat{\mathbbm{1}}
\end{align}
which is equivalent to the right eigenvectors being orthogonal w.r.t\ $\langle \cdot, \cdot\rangle_{\mat{\Pi}}$.
\end{proof}

\section{SFA}
\subsection{Addition of noise}
\label{app:LinSFAnoise}
This section explores how the covariance matrix $\mat{\Sigma}$ and second moment matrix $\widehat{\mat{\Sigma}}$ associated to a time series $\vec{x}(t)$ are affected by the addition of centered Gaussian noise with variance $\sigma^2$ along all axes, i.e. $\vec{x}_\sigma(t)=\vec{x}(t)+\vec{\xi}(t)$ where $\vec{\xi}(t)\sim\mathcal{N}(\vec{0},\sigma^2\mat{\mathbbm{1}})$. First consider the following decomposition of the starting covariance matrix
\begin{align}
\mat{\Sigma}&=\langle \vec{c}(t)\vec{c}(t)^T\rangle_t \label{eq:CovMatdecomp_first}\\
&=\langle (\vec{x}(t)-\langle \vec{x}(t)\rangle_t)(\vec{x}(t)-\langle \vec{x}(t)\rangle_t)^T\rangle_t\\
&=\langle\vec{x}(t)\vec{x}(t)^T\rangle_t - \langle \vec{x}(t)\rangle_t\langle\vec{x}(t)^T\rangle_t-\langle \vec{x}(t)\rangle_t\langle\vec{x}(t)^T\rangle_t+\langle \vec{x}(t)\rangle_t\langle\vec{x}(t)^T\rangle_t\\
&=\langle\vec{x}(t)\vec{x}(t)^T\rangle_t - \langle \vec{x}(t)\rangle_t\langle\vec{x}(t)^T\rangle_t \label{eq:CovMatdecomp_last}
\end{align}
where the first term is the 2nd moment matrix of $\vec{x}(t)$. For the noisy signal $\vec{x}_\sigma(t)$ with covariance matrix $\mat{\Sigma}_\sigma$, \cref{eq:CovMatdecomp_last} reads:
\begin{equation}
\mat{\Sigma}_\sigma=\langle\vec{x}_\sigma(t)\vec{x}_\sigma(t)^T\rangle_t-\langle \vec{x}_\sigma(t)\rangle_t\langle\vec{x}_\sigma(t)^T\rangle_t\label{eq:CovMatdecompnoise}
\end{equation}
In the limit of large samples, it is possible to relate $\mat{\Sigma}_\sigma$ to $\mat{\Sigma}$. To see this, consider each term on the right-hand side of \cref{eq:CovMatdecompnoise} in this limit. The first term can be expanded as follows:
\begin{align}
\langle\vec{x}_\sigma(t)\vec{x}_\sigma(t)^T\rangle_t&=\langle(\vec{x}(t)+\vec{\xi}(t))(\vec{x}(t)+\vec{\xi}(t))^T\rangle_t\label{eq:noisecov1}\\
&=\langle \vec{x}(t)\vec{x}(t)^T\rangle_t + \langle \vec{x}(t)\vec{\xi}(t)^T\rangle_t +\langle \vec{\xi}(t)\vec{x}(t)^T\rangle_t +\langle \vec{\xi}(t)\vec{\xi}(t)^T\rangle_t\label{eq:noisecov2}
\end{align}
For $T\to\infty$, the last term in \cref{eq:noisecov2} is equal to $\sigma^2\mat{\mathbbm{1}}$ because $\vec{\xi}(t)$ is sampled from a multivariate Gaussian with variance $\sigma^2$, and the second and third terms are equal to zero since $\vec{x}(t)$ and $\vec{\xi}(t)$ are statistically independent and $\vec{\xi}(t)$ has zero mean:
\begin{align}
\lim_{T\to\infty}\langle \vec{x}(t)\vec{\xi}(t)^T\rangle_t &=\lim_{T\to\infty}\langle \vec{x}(t)\rangle_t\langle\vec{\xi}(t)^T\rangle_t=0\\
\lim_{T\to\infty}\langle \vec{\xi}(t)\vec{x}(t)^T\rangle_t &=\lim_{T\to\infty}\langle\vec{\xi}(t)\rangle_t\langle \vec{x}(t)^T\rangle_t=0
\end{align}
Therefore:
\begin{equation}
\lim_{T\to\infty}\langle\vec{x}_\sigma(t)\vec{x}_\sigma(t)^T\rangle_t=\langle \vec{x}(t)\vec{x}(t)^T\rangle_t+\sigma^2\mat{\mathbbm{1}}\label{eq:noisecovlimit1}
\end{equation}
The second term in \cref{eq:CovMatdecompnoise} can be expanded as follows:
\begin{align}
\langle \vec{x}_\sigma(t)\rangle_t\langle\vec{x}_\sigma(t)^T\rangle_t&=(\langle\vec{x}(t)\rangle_t+\langle\vec{\xi}(t)\rangle_t)(\langle\vec{x}(t)\rangle_t+\langle\vec{\xi}(t)\rangle_t)^T\\
&=\langle \vec{x}(t)\rangle_t\langle\vec{x}(t)^T\rangle_t+ \langle \vec{x}(t)\rangle_t\langle\vec{\xi}(t)^T\rangle_t+ \langle \vec{\xi}(t)\rangle_t\langle\vec{x}(t)^T\rangle_t+ \langle \vec{\xi}(t)\rangle_t\langle\vec{\xi}(t)^T\rangle_t\label{eq:noisecov3}
\end{align}
In the limit $T\to \infty$, all terms in \cref{eq:noisecov3} except the first are zero since $\vec{\xi}(t)$ has zero mean. Therefore:
\begin{equation}
\lim_{T\to\infty}\langle \vec{x}_\sigma(t)\rangle_t\langle\vec{x}_\sigma(t)^T\rangle_t=\langle \vec{x}(t)\rangle_t\langle\vec{x}(t)^T\rangle_t\label{eq:noisecovlimit2}
\end{equation}
Thus, combining \cref{eq:noisecovlimit1,eq:noisecovlimit2} leads to the following limit on $\mat{\Sigma}_\sigma$:
\begin{align}
\lim_{T\to\infty}\mat{\Sigma}_\sigma&=\langle \vec{x}(t)\vec{x}(t)^T\rangle_t+\sigma^2\mat{\mathbbm{1}}-\langle \vec{x}(t)\rangle_t\langle\vec{x}(t)^T\rangle_t\\
&=\mat{\Sigma}+\sigma^2\mat{\mathbbm{1}}\label{eq:noisecovlimit3}
\end{align}
\cref{eq:noisecovlimit3} says that in the large data limit the signal $\vec{x}_\sigma(t)$ has the same covariance matrix as $\vec{x}(t)$ except for a positive increase along the diagonal by $\sigma^2$. In this limit, the addition of noise leaves the eigenvectors invariant and simply increases the eigenvalues by $\sigma^2$. Thus, any eigenvalues of zero become positive, meaning that $\mat{\Sigma}_\sigma$ is positive definite and is therefore invertible.

Note that the relation between $\mat{\Sigma}_\sigma$ to $\mat{\Sigma}$ in \cref{eq:noisecovlimit3} is analogous to the one between $\mat{B}_\epsilon$ and $\mat{B}$ in \cref{eq:QuickDirty}, meaning that in the large data limit the addition of noise is equivalent to the numerical hack presented in Appendix \ref{app:GenEVprob}. Therefore, in analogy to this numerical hack, the addition of noise needs to balance two goals. On the one hand, $\sigma$ needs to be large enough that all eigenvalues equal to or close to zero are sufficiently increased and the matrices $\mat{\Sigma}^{-\frac{1}{2}}$ and $\mat{\Sigma}^{-1}$ can be approximately computed. On the other hand, $\sigma$ should not be so large that the noise masks the starting signal $\vec{x}(t)$. Practically, these two goals are best balanced by performing parameter tuning on $\sigma$ \citep{Konen2009,Tabeart2020}.

The addition of noise has an equivalent impact on the second moment matrix of $\vec{x}(t)$, which is given by:
\begin{equation}
\widehat{\mat{\Sigma}}=\langle \vec{x}(t)\vec{x}(t)^T\rangle_t
\end{equation}
When $\vec{x}(t)$ is perturbed by noise this matrix becomes
\begin{equation}
\widehat{\mat{\Sigma}}_\sigma=\langle \vec{x}_\sigma(t)\vec{x}_\sigma(t)^T\rangle_t
\end{equation}
which in the limit $T\to\infty$ is described by \cref{eq:noisecovlimit1}. Therefore, in this limit the second moment matrix of $\vec{x}(t)$ is perturbed in the same way as the covariance matrix, meaning that all details outlined in the previous paragraph apply also to $\widehat{\mat{\Sigma}}$.

It is worth emphasizing that the effects of noise explored in this section are only exact in the limit $T\to\infty$. For large finite values of $T$ that are typically considered in SFA, there is inevitably always some degree of sampling variability, meaning that the additional diagonal entries in $\mat{\Sigma}_\sigma$ and $\widehat{\mat{\Sigma}}_\sigma$ are subject to errors, and there can also be additional off-diagonal terms that correspond to random correlations between $\vec{x}(t)$ and $\vec{\xi}(t)$.

\subsection{Whitening and symmetric normalization in Type 1 SFA Problems}
\label{app:SymNormWhite}

In statistics, \textit{whitening} refers to any linear transformation that converts a set of starting variables into a new set of uncorrelated variables, each with unit variance. This transformation ensures that the new variables have a covariance matrix equal to the identity matrix \citep{Kessy2018}. Typically, whitening assumes the input data has zero mean, or equivalently that a prior centering step has been performed, meaning that it more naturally fits into the Type 1 SFA formalism.

For a time series $\vec{x}(t)$, whitening transformations can be expressed as $\vec{z}(t)=\mat{H}\vec{c}(t)$ where $\vec{c}(t)$ is the centered signal and $\mat{H}$ is some matrix such that
\begin{align}
\langle \vec{z}(t)\vec{z}(t)^T\rangle_t&=\mat{H}\langle \vec{c}(t)\vec{c}(t)^T\rangle_t\mat{H}^T\\
&=\mat{H}\mat{\Sigma}\mat{H}^T\label{eq:whitening1}\\
&=\mat{\mathbbm{1}}\label{eq:whitening2}
\end{align}
Such a transformation is only possible if both $\mat{\Sigma}$ and $\mat{H}$ are invertible. To see this, note that $\text{det}(\mat{\mathbbm{1}})=1$, which means that $\text{det}(\mat{H}\mat{\Sigma}\mat{H}^T)=\text{det}(\mat{H})\text{det}(\mat{\Sigma})\text{det}(\mat{H}^T)=1$ and therefore both $\mat{\Sigma}$ and $\mat{H}$ must have non-zero determinant. Because of this, in situations where $\mat{\Sigma}$ has one or more eigenvalues of zero, whitening cannot be performed directly and it is necessary to either add noise or use the pseudoinverse (see \cref{sec:Type1Norm}).

Multiplying \cref{eq:whitening1,eq:whitening2} by $\mat{H}^T$ from the left then gives
\begin{equation}
(\mat{H}^T\mat{H}\mat{\Sigma})\mat{H}^T=\mat{H}^T
\end{equation}
which in turn means that
\begin{equation}
\mat{H}^T\mat{H}\mat{\Sigma}=\mat{\mathbbm{1}}
\end{equation}
and therefore
\begin{equation}
\mat{H}^T\mat{H}=\mat{\Sigma}^{-1}\label{eq:whitening3}
\end{equation}
Since \cref{eq:whitening3} has the same form as \cref{eq:matrootaltinv_orthog}, the whitening matrix $\mat{H}$ satisfies the requirement for being an inverse square root of $\mat{\Sigma}$, i.e.\ $\mat{H}=\mat{\Sigma}^{-\frac{1}{2}}=\mat{Q}\mat{U}\mat{D}^{-\frac{1}{2}}\mat{U}^T$, where $\mat{Q}$ is an orthogonal matrix, $\mat{U}$ contains the eigenvectors of $\mat{\Sigma}$, and $\mat{D}^{-\frac{1}{2}}$ contains the inverse positive square roots of the eigenvalues of $\mat{\Sigma}$ (see Appendix \ref{app:SqrtPSD}).

Note that since $\mat{Q}$ is included in the definition of $\mat{\Sigma}^{-\frac{1}{2}}$ above, whitening transformations are only defined up to a rotation, and because of this there exist different choices for whitening a given dataset \citep{Kessy2018}. The two most common choices in the context of SFA are \textit{ZCA}- and \textit{PCA}-whitening, with the former corresponding to $\mat{Q}=\mat{\mathbbm{1}}$ and $\mat{\Sigma}^{-\frac{1}{2}}=\mat{U}\mat{D}^{-\frac{1}{2}}\mat{U}^T$ and the latter to $\mat{Q}=\mat{U}^T$ and $\mat{\Sigma}^{-\frac{1}{2}}=\mat{D}^{-\frac{1}{2}}\mat{U}^T$. In words, these transformations can be described sequentially from right to left in the following two/three steps: (i) rotate the coordinate system into the eigenbasis $\mat{U}^T$ of $\mat{\Sigma}$, (ii) scale the components along each axis by $D_{ii}^{-\frac{1}{2}}=\lambda_i^{-\frac{1}{2}}$, where $\lambda_i$ is the eigenvalue associated to each eigenvector, (iii) in the case of ZCA, rotate back into the original coordinate system using the transformation $\mat{U}$. It is worth noting that there are two ways in which ZCA is special in comparison to other whitening methods. Firstly, the corresponding matrix is guaranteed to be symmetric since $(\mat{U}\mat{D}^{-\frac{1}{2}}\mat{U}^T)^T=\mat{U}\mat{D}^{-\frac{1}{2}}\mat{U}^T$, while for $\mat{Q}^T\neq\mat{\mathbbm{1}}$ this is generally not the case. Secondly, due to step (iii) in ZCA, this version of whitening is the one that preserves the original orientation of data points as much as possible.

The use of whitening as a preprocessing step in Type 1 SFA problems modifies the underlying generalized eigenvalue problems in two key ways. Firstly, since $\vec{z}(t)$ has a unit covariance matrix, the constraints are automatically fulfilled and the problems reduce to regular eigenvalue problems. Secondly, the remaining matrix in each problem obtains an additional pre-multiplication by $\mat{\Sigma}^{-\frac{1}{2}}$ and post-multiplication by $(\mat{\Sigma}^{-\frac{1}{2}})^T$. To see this, note that the whitened signal $\vec{z}(t)$ and its time derivative $\dot{\vec{z}}(t)$ are given by $\vec{z}(t)=\mat{\Sigma}^{-\frac{1}{2}}\vec{c}(t)$ and $\dot{\vec{z}}(t)=\mat{\Sigma}^{-\frac{1}{2}}\dot{\vec{c}}(t)$. Therefore, the covariance matrix of $\dot{\vec{z}}(t)$ can be expressed as
\begin{align}
\langle \dot{\vec{z}}(t)\dot{\vec{z}}(t)^T\rangle_t&=\langle \dot{\vec{z}}(t)\dot{\vec{z}}(t)^T\rangle_t\\
&=\mat{\Sigma}^{-\frac{1}{2}}\langle \dot{\vec{c}}(t)\dot{\vec{c}}(t)^T\rangle_t(\mat{\Sigma}^{-\frac{1}{2}})^T\\
&=\mat{\Sigma}^{-\frac{1}{2}}\dot{\mat{\Sigma}}(\mat{\Sigma}^{-\frac{1}{2}})^T
\end{align}
the STC matrix with time-lag $\tau$ of $\vec{z}(t)$ as
\begin{align}
\frac{1}{2}\big(\langle\vec{z}(t)\vec{z}(t+\tau)^T\rangle_t+\langle\vec{z}(t+\tau)\vec{z}(t)^T\rangle_t\big)&= \frac{1}{2}\big(\mat{\Sigma}^{-\frac{1}{2}}\langle\vec{c}(t)\vec{c}(t+\tau)^T\rangle_t(\mat{\Sigma}^{-\frac{1}{2}})^T+\mat{\Sigma}^{-\frac{1}{2}}\langle\vec{c}(t+\tau)\vec{c}(t)^T\rangle_t(\mat{\Sigma}^{-\frac{1}{2}})^T\big)\\
&=\mat{\Sigma}^{-\frac{1}{2}}\bigg(\frac{1}{2}\big(\langle\vec{c}(t)\vec{c}(t+\tau)^T\rangle_t+\langle\vec{c}(t+\tau)\vec{c}(t)^T\rangle_t\big)\bigg)(\mat{\Sigma}^{-\frac{1}{2}})^T\\
&=\mat{\Sigma}^{-\frac{1}{2}}\mat{\Omega}_\tau(\mat{\Sigma}^{-\frac{1}{2}})^T\label{eq:STCtauwhite}
\end{align}
and the LF matrix of $\vec{z}(t)$ as
\begin{align}
\sum_{\tau=0}^{\tau_{\text{max}}}\frac{\gamma^\tau}{2}\big(\langle\vec{z}(t)\vec{z}(t+\tau)^T\rangle_t+\langle\vec{z}(t+\tau)\vec{z}(t)^T\rangle_t\big)\overset{(\ref{eq:STCtauwhite})}&{=}\sum_{\tau=0}^{\tau_{\text{max}}}\gamma^\tau\mat{\Sigma}^{-\frac{1}{2}}\mat{\Omega}_\tau(\mat{\Sigma}^{-\frac{1}{2}})^T\\
&=\mat{\Sigma}^{-\frac{1}{2}}\bigg(\sum_{\tau=0}^{\tau_{\text{max}}}\gamma^\tau\mat{\Omega}_\tau\bigg)(\mat{\Sigma}^{-\frac{1}{2}})^T\\
&=\mat{\Sigma}^{-\frac{1}{2}}\mat{\Psi}_\gamma(\mat{\Sigma}^{-\frac{1}{2}})^T
\end{align} 
Note that the regular eigenvalue problems involving these matrices are precisely those listed in \cref{tab:AllEigenvalueProblems} that correspond to Type 1 symmetric normalized SFA, $\tau$SFA, or LFSFA, respectively. Thus, the use of whitening as a preprocessing step in Type 1 SFA Problems is equivalent doing symmetric normalization.

\section{Limit Theorems}
\label{app:Lim}

\subsection{Type 1 SFA Problems}
\label{app:LimType1}

\textbf{\cref{thm:Type1LimConstraint}} \textit{For an ergodic Markovian one-hot trajectory $\vec{x}(t)$, the matrix $\mat{\Sigma}$ associated to the Type 1 constraints obeys the following limit:
\begin{align}
\lim_{T\to\infty}\mat{\Sigma}&=\mat{\Pi}-\vec{\pi \pi}^T
\end{align}
where $\vec{\pi}$ is the unique stationary distribution of the underlying Markov chain and $\mat{\Pi}=\textup{diag}(\vec{\pi})$. Moreover, the limiting matrix $\mat{\Pi}-\vec{\pi \pi}^T$ is guaranteed to have a single eigenvalue of $0$.}

\begin{proof}
The covariance matrix $\mat{\Sigma}$ of $\vec{x}(t)$ can be written as (see Equations \ref{eq:CovMatdecomp_first}-\ref{eq:CovMatdecomp_last})
\begin{equation}
\mat{\Sigma}=\langle\vec{x}(t)\vec{x}(t)^T\rangle_t - \langle \vec{x}(t)\rangle_t\langle\vec{x}(t)^T\rangle_t
\end{equation}
or in elementwise notation as
\begin{equation}
\Sigma_{ij}=\langle x_i(t)x_j(t)\rangle_t - \langle x_i(t)\rangle_t\langle x_j(t)\rangle_t\label{eq:CovMatelement}
\end{equation}
In the second term of \cref{eq:CovMatelement}, $\langle x_i(t)\rangle_t$ is the average value of $x_i(t)$. Since $x_i(t)=1$ when state $s_i$ is occupied and $0$ otherwise, this average is equivalent to the proportion of times that $s_i$ is occupied in the entire trajectory. Since the Markov chain is ergodic by assumption, this proportion converges to a unique stationary probability $\pi_i$ as $T\to\infty$ \citep{Seabrook2023}. Therefore, in this limit the second term of \cref{eq:CovMatelement} is $\pi_i\pi_j$. For the first term of \cref{eq:CovMatelement}, note that due to the properties of the one-hot encoding the product $x_i(t)x_j(t)$ is non-zero only if $i=j$, in which case it is equal to $x_i^2(t)$. Since $x_i^2(t)$ is equivalent to $x_i(t)$ for a one-hot encoding, its average is also equal to the stationary probability $\pi_i$. Therefore
\begin{equation}
\lim_{T\to\infty}\Sigma_{ij}=\left\{
	\begin{array}{ll}
		\pi_i-\pi_i^2 & \mbox{if } i=j\\
		-\pi_i\pi_j & \mbox{otherwise }
	\end{array}
\right.
\end{equation}
or in matrix notation
\begin{equation}
\lim_{T\to\infty}\mat{\Sigma}=\mat{\Pi}-\vec{\pi \pi}^T\label{eq:LimCov}
\end{equation}
Note that the constant vector of ones is an eigenvector of this matrix with eigenvalue $0$, i.e.\
\begin{align}
(\mat{\Pi}-\vec{\pi \pi}^T)\vec{1}&=\underbrace{\mat{\Pi}\vec{1}}_{\vec{\pi}}-\vec{\pi}\underbrace{\vec{\pi}^T\vec{1}}_{=1}\\
&=\vec{\pi}-\vec{\pi}\\
&=0
\end{align}
Moreover, if $\vec{v}$ is any eigenvector of $\mat{\Pi}-\vec{\pi \pi}^T$ with eigenvalue $0$, i.e.\ $(\mat{\Pi}-\vec{\pi \pi}^T)\vec{v}=0$, then
\begin{align}
\mat{\Pi}\vec{v}&=\vec{\pi \pi}^T\vec{v}
\end{align}
If $c=\vec{\pi}^T\vec{v}$, then this becomes
\begin{align}
\mat{\Pi}\vec{v}&=c\vec{\pi}
\end{align}
which in component notation reads
\begin{align}
\pi_iv_i=c\pi_i
\end{align}
Since the chain is ergodic by assumption, $\pi_i>0\;\forall i$ and therefore each side of the above equation can be divided by $\pi_i$, leading to
\begin{align}
v_i=c
\end{align}
meaning that each component of $\vec{v}$ is the same and therefore $\vec{v}$ is a constant vector. Thus, $\vec{v}$ is some multiple of $\vec{1}$ which means that the eigenspace associated to $\lambda=0$ is one-dimensional.
\end{proof}

\noindent\textbf{\cref{thm:Type1LimObjective}} \textit{For an ergodic Markovian one-hot trajectory $\vec{x}(t)$, the matrices $\dot{\mat{\Sigma}}$, $\mat{\Omega}_\tau$, and $\mat{\Psi}_\gamma$ associated to the Type 1 objectives obey the following limits:
\begin{align}
\lim_{T\to\infty}\dot{\mat{\Sigma}}&=2\mat{L}_{\textup{dir}}+\vec{\pi \pi}^T\\
\lim_{T\to\infty}\mat{\Omega}_\tau&=\mat{\Pi}(\mat{P}^\tau)_{\textup{add}}-\vec{\pi \pi}^T\\
\lim_{T\to\infty}\mat{\Psi}_\gamma&=\mat{\Pi}\mat{M}_{\textup{add}}-\alpha\vec{\pi}\vec{\pi}^T
\end{align}
where $\mat{L}_{\textup{dir}}$ is is the combinatorial directed Laplacian matrix associated to the underlying Markov chain, $\mat{P}$ is the transition matrix, $\mat{M}$ is the SR matrix with discount factor $\gamma$, and $\alpha=\frac{1-\gamma^{\tau_{\textup{max}}+1}}{1-\gamma}\in[1,\frac{1}{1-\gamma})$ is a scalar constant.}

\begin{proof}
Consider first the STC matrix at time-lag $\tau$, which is given by
\begin{align}
\mat{\Omega}_\tau&=\frac{1}{2}\big(\langle\vec{c}(t)\vec{c}(t+\tau)^T\rangle_t+\langle\vec{c}(t+\tau)\vec{c}(t)^T\rangle_t\big)
\end{align}
or in elementwise notation
\begin{align}
(\mat{\Omega}_\tau)_{ij}&=\frac{1}{2}\big(\langle c_i(t)c_j(t+\tau)\rangle_t+\langle c_i(t+\tau) c_j(t)\rangle_t\big)\label{eq:STCtauelement}
\end{align}
The first term on the right-hand side of \cref{eq:STCtauelement} can be rearranged to give:
\begin{align}
\langle c_i(t)c_j(t+\tau)\rangle_t&=\langle (x_i(t)-\langle x_i(t)\rangle_t)(x_j(t+\tau)-\langle x_j(t)\rangle_t)\rangle_t\\
&=\langle x_i(t)x_j(t+\tau)\rangle_t - \langle x_i(t)\rangle_t\langle x_j(t)\rangle_t-\langle x_i(t)\rangle_t\langle x_j(t+\tau)\rangle_t+\langle x_i(t)\rangle_t\langle x_j(t)\rangle_t\label{eq:STCtauelementdecomp}
\end{align}
In the first term of \cref{eq:STCtauelementdecomp}, the product $x_i(t)x_j(t+\tau)$ is equal to $1$ if state $s_i$ is occupied at time $t$ and state $s_j$ is occupied $\tau$ steps later, and is zero otherwise. Therefore, this first term counts the number of times this transition occurs, and is equivalent to the frequency with which this $\tau$-step transition happens. For an ergodic Markov chain this converges to the flow probability from $s_i$ to $s_j$ of the $\tau$-step Markov chain, i.e.\ $\pi_i(\mat{P}^\tau)_{ij}$, as $T\to\infty$ \citep{Seabrook2023}. The other terms in \cref{eq:STCtauelementdecomp} all converge to $\pi_i\pi_j$ as $T\to\infty$, meaning that:
\begin{align}
\lim_{T\to\infty}\langle c_i(t)c_j(t+\tau)\rangle_t&=\pi_i(\mat{P}^\tau)_{ij}-\pi_i\pi_j-\pi_i\pi_j+\pi_i\pi_j\\
&=\pi_i(\mat{P}^\tau)_{ij}-\pi_i\pi_j
\end{align}
By an equivalent argument, it is possible to show a similar result for the second term on the right-hand side of \cref{eq:STCtauelement}:
\begin{align}
\lim_{T\to\infty}\langle c_i(t+\tau) c_j(t)\rangle_t&=\pi_j(\mat{P}^\tau)_{ji}-\pi_i\pi_j
\end{align}
Therefore, in this limit \cref{eq:STCtauelement} becomes
\begin{align}
\lim_{T\to\infty}(\mat{\Omega}_\tau)_{ij}&=\frac{1}{2}\big(\pi_i(\mat{P}^\tau)_{ij}-\pi_i\pi_j+\pi_j(\mat{P}^\tau)_{ji}-\pi_i\pi_j\big)\\
&=\frac{\pi_i(\mat{P}^\tau)_{ij}+\pi_j(\mat{P}^\tau)_{ji}}{2}-\pi_i\pi_j
\end{align}
or in matrix notation
\begin{align}
\lim_{T\to\infty}\mat{\Omega}_\tau&=\frac{\mat{\Pi P}^\tau+(\mat{\Pi P}^\tau)^T}{2}-\vec{\pi}\vec{\pi}^T\label{eq:STClimit_first}\\
&=\frac{\mat{\Pi P}^\tau+(\mat{P}^\tau)^T\mat{\Pi}^T}{2}-\vec{\pi}\vec{\pi}^T\\
&=\mat{\Pi}\bigg(\frac{\mat{P}^\tau+\mat{\Pi}^{-1}(\mat{P}^\tau)^T\mat{\Pi}}{2}\bigg)-\vec{\pi}\vec{\pi}^T\\
&=\mat{\Pi}(\mat{P}^\tau)_{\text{add}}-\vec{\pi}\vec{\pi}^T\label{eq:STClimit_last}
\end{align}
where the subscript in \cref{eq:STClimit_last} denotes the process of additive reversibilization (see \cref{sec:SReig}).

The LF matrix is related to $\mat{\Omega}_\tau$ by:
\begin{equation}
\mat{\Psi}_\gamma=\sum_{\tau=0}^{\tau_{\text{max}}}\gamma^\tau\mat{\Omega}_\tau\label{eq:LFmat2}
\end{equation}
Therefore, in the limit $T\to\infty$ this matrix evaluates to
\begin{align}
\lim_{T\to\infty}\mat{\Psi}_\gamma&=\lim_{T\to\infty}\bigg( \sum_{\tau=0}^{\tau_{\text{max}}}\gamma^\tau\mat{\Omega}_\tau\bigg)\label{eq:LFlimit_first}\\
&=\sum_{\tau=0}^{\tau_{\text{max}}}\gamma^\tau\bigg( \lim_{T\to\infty}\mat{\Omega}_\tau\bigg)\\
\overset{(\ref{eq:STClimit_last})}&{=}\sum_{\tau=0}^{\tau_{\text{max}}}\gamma^\tau \bigg(\mat{\Pi}(\mat{P}^\tau)_{\text{add}}-\vec{\pi}\vec{\pi}^T \bigg)\\
&=\mat{\Pi}\bigg(\sum_{\tau=0}^{\tau_{\text{max}}}\gamma^\tau(\mat{P}^\tau)_{\text{add}}\bigg)-\vec{\pi}\vec{\pi}^T\sum_{\tau=0}^{\tau_{\text{max}}}\gamma^\tau\label{eq:LFlimit_fourth}\\
\overset{(\ref{eq:Madd2})}&{=}\mat{\Pi}\mat{M}_{\text{add}}-\vec{\pi}\vec{\pi}^T\bigg(\underbrace{\frac{1-\gamma^{\tau_{\text{max}}+1}}{1-\gamma}}_{=\alpha}\bigg)\\
&=\mat{\Pi}\mat{M}_{\text{add}}-\alpha\vec{\pi}\vec{\pi}^T\label{eq:LFlimit_last}
\end{align}
Note that the only difference between $\mat{M}_{\textup{add}}$ in the equations above and the definition in \cref{eq:Madd2} is that the former has a finite horizon while the latter has an infinite horizon (see \cref{sec:SRtime}). Moreover, the second term in \cref{eq:LFlimit_fourth} is evaluated using the formula for a geometric series with a finite number of terms, and is equal to a constant $\alpha$ that depends on $\gamma$ and $\tau_{\text{max}}$. For the miminum value $\tau_{\text{max}}=1$ this constant is $1$ and in the limit $\tau_{\text{max}}\to\infty$ it is $\frac{1}{1-\gamma}$. Therefore, $\alpha\in[1,\frac{1}{1-\gamma})$.

The matrix $\dot{\mat{\Sigma}}$ is related to $\mat{\Omega}_1$ and $\mat{\Sigma}$ by (see Equations \ref{eq:ACmatfirst}-\ref{eq:ACmatlast}):
\begin{equation}
\dot{\mat{\Sigma}}=2(\mat{\Sigma}-\mat{\Omega}_1)
\end{equation}
Therefore, in the limit $T\to\infty$ this matrix evaluates to
\begin{align}
\lim_{T\to\infty}\dot{\mat{\Sigma}}&=\lim_{T\to\infty}2(\mat{\Sigma}-\mat{\Omega}_1)\\
\overset{(\ref{eq:STClimit_last},\ref{eq:LimCov})}&{=}2\mat{\Pi}-2\mat{\Pi}\mat{P}_{\textup{add}}+\vec{\pi}\vec{\pi}^T\\
\overset{(\ref{eq:Padd})}&{=}2\bigg(\mat{\Pi}-\frac{\mat{\Pi}\mat{P}+\mat{P}^T\mat{\Pi}}{2}\bigg)+\vec{\pi}\vec{\pi}^T\\
&=2\bigg(\underbrace{\mat{\Pi}-\frac{\mat{\Pi}\mat{P}+(\mat{\Pi}\mat{P})^T}{2}}_{=\mat{L}_{\text{dir}}}\bigg)+\vec{\pi}\vec{\pi}^T\label{eq:LdirDef}\\
&=2\mat{L}_{\text{dir}}+\vec{\pi}\vec{\pi}^T
\end{align}
where $\mat{L}_{\text{dir}}$ is the \textit{combinatorial directed Laplacian} defined by \citet{Chung2005}, which can be understood using concepts from Markov chain theory \citep{Seabrook2023}. In particular, $\mat{\Pi}\mat{P}$ contains the stationary flow probabilities of the underlying Markov chain, which are all non-negative, meaning that this matrix can be interpreted as the weight matrix of a graph $G$ connecting all pairs of states $s_i,s_j\in\mathcal{S}$. Moreover, $\mat{\Pi}\mat{P}$ is related to $\mat{P}$ through a scaling of the rows by the stationary probabilities. Therefore, in the terminology of \citet{Seabrook2023}, $G$ belongs to the random walk set of the underlying Markov chain, and so performing a random walk on $G$ produces this Markov chain. Moreover, since the chain is not assumed to be reversible $\mat{\Pi}\mat{P}$ is generally non-symmetric, in which case $G$ is a directed graph. While graph Laplacians are typically defined for undirected graphs, various methods have been proposed for the extension to the directed case \citep{Seabrook2023}. The quantity $\mat{L}_{\text{dir}}$ does this by symmetrizing the edges in $G$ to produce a new undirected graph $G'$ for which the weight matrix is $\frac{\mat{\Pi}\mat{P}+(\mat{\Pi}\mat{P})^T}{2}$ and the degree of each vertex is given by:
\begin{align}
d_i&=\frac{1}{2}\sum_j\pi_iP_{ij}+\frac{1}{2}\sum_j\pi_jP_{ji}\label{eq:DegSymGraph1}\\
&=\frac{\pi_i}{2}\underbrace{\sum_jP_{ij}}_{=1}+\frac{\pi_i}{2}\\
&=\pi_i\label{eq:DegSymGraph3}
\end{align}
Therefore, $\mat{L}_{\text{dir}}$ is like the combinatorial graph Laplacian for undirected graphs, i.e.\ $\mat{L}=\mat{D}-\mat{W}$, but defined on the symmetrized graph $G'$.
\end{proof}

\subsection{Type 2 SFA Problems}
\label{app:LimType2}

\textbf{\cref{thm:Type2LimConstraint}} \textit{For an ergodic Markovian one-hot trajectory $\vec{x}(t)$, the matrix $\widehat{\mat{\Sigma}}$ associated to the Type 2 constraints obeys the following limit:
\begin{align}
\lim_{T\to\infty}\widehat{\mat{\Sigma}}&=\mat{\Pi}
\end{align}
where $\vec{\pi}$ is the unique stationary distribution of the underlying Markov chain and $\mat{\Pi}=\textup{diag}(\vec{\pi})$. Moreover, the limiting matrix $\mat{\Pi}$ is guaranteed to have no eigenvalues of $0$.}

\begin{proof}
The 2nd moment matrix $\widehat{\mat{\Sigma}}$ of $\vec{x}(t)$ is given by
\begin{equation}
\mat{\Sigma}=\langle\vec{x}(t)\vec{x}(t)^T\rangle_t 
\end{equation}
or in elementwise notation as
\begin{equation}
\widehat{\Sigma}_{ij}=\langle x_i(t)x_j(t)\rangle_t\label{eq:2ndMomMatelement}
\end{equation}
\cref{eq:2ndMomMatelement} is simply the first term of \cref{eq:CovMatelement}, and therefore applying the same logic as in the previous proof leads to:
\begin{equation}
\lim_{T\to\infty}\widehat{\Sigma}_{ij}=\left\{
	\begin{array}{ll}
		\pi_i & \mbox{if } i=j\\
		0 & \mbox{otherwise }
	\end{array}
\right.
\end{equation}
or in matrix notation
\begin{equation}
\lim_{T\to\infty}\widehat{\mat{\Sigma}}=\mat{\Pi}\label{eq:Lim2ndMom}
\end{equation}

Note that $\mat{\Pi}$ is a diagonal matrix containing the stationary probabilities of the Markov chain, which are also the eigenvalues of this matrix. Moreover, since the chain is ergodic by assumption, all stationary probabilities, or equivalently all eigenvalues, are strictly positive.
\end{proof}

\noindent \textbf{\cref{thm:Type2LimObjective}} \textit{For an ergodic Markovian one-hot trajectory $\vec{x}(t)$, the matrices $\widehat{\dot{\mat{\Sigma}}}$, $\widehat{\mat{\Omega}}_\tau$, and $\widehat{\mat{\Psi}}_\gamma$ associated to the Type 2 objectives obey the following limits:
\begin{align}
\lim_{T\to\infty}\widehat{\dot{\mat{\Sigma}}}&=2\mat{L}_{\textup{dir}}\\
\lim_{T\to\infty}\widehat{\mat{\Omega}}_\tau&=\mat{\Pi}(\mat{P}^\tau)_{\textup{add}}\\
\lim_{T\to\infty}\widehat{\mat{\Psi}}_\gamma&=\mat{\Pi}\mat{M}_{\textup{add}}
\end{align}
where $\mat{L}_{\textup{dir}}$ is the combinatorial directed Laplacian matrix associated to the underlying Markov chain, $\mat{P}$ is the transition matrix, and $\mat{M}$ is the SR matrix with discount factor $\gamma$.}

\begin{proof}
The matrix $\widehat{\mat{\Omega}}_\tau$ is the same as $\mat{\Omega}_\tau$ except that no centering transformation is assumed and $\vec{c}(t)$ is exchanged for $\vec{x}(t)$, i.e.\
\begin{align}
\widehat{\mat{\Omega}}_\tau&=\frac{1}{2}\big(\langle\vec{x}(t)\vec{x}(t+\tau)^T\rangle_t+\langle\vec{x}(t+\tau)\vec{x}(t)^T\rangle_t\big)
\end{align}
or in elementwise notation
\begin{align}
(\widehat{\mat{\Omega}}_\tau)_{ij}&=\frac{1}{2}\big(\langle x_i(t)x_j(t+\tau)\rangle_t+\langle x_i(t+\tau) x_j(t)\rangle_t\big)\label{eq:STCtauelement_nocent}
\end{align}
By the same reasoning as in the proof of \cref{thm:Type2LimObjective}, in the limit $T\to\infty$ the first and second terms on the right-hand side of \cref{eq:STCtauelement_nocent} are given by $\pi_i(\mat{P}^\tau)_{ij}$ and $\pi_j(\mat{P}^\tau)_{ji}$, respectively. Therefore
\begin{align}
\lim_{T\to\infty}(\widehat{\mat{\Omega}}_\tau)_{ij}&=\frac{\pi_i(\mat{P}^\tau)_{ij}+\pi_j(\mat{P}^\tau)_{ji}}{2}
\end{align}
or in matrix notation
\begin{align}
\lim_{T\to\infty}\widehat{\mat{\Omega}}_\tau&=\frac{\mat{\Pi P}^\tau+(\mat{\Pi P}^\tau)^T}{2}\\
&=\mat{\Pi}(\mat{P}^\tau)_{\text{add}}\label{eq:STClimit_nocent}
\end{align}
where the first line reduces to the second line following the same steps as in \Crefrange{eq:STClimit_first}{eq:STClimit_last}.

The matrix $\widehat{\mat{\Psi}}_\gamma$ is related to $\widehat{\mat{\Omega}}_\tau$ by
\begin{equation}
\widehat{\mat{\Psi}}_\gamma=\sum_{\tau=0}^{\tau_{\text{max}}}\gamma^\tau\widehat{\mat{\Omega}}_\tau
\end{equation}
and can be evaluated in the limit $T\to\infty$ by using \cref{eq:STClimit_nocent} and applying a similar set of steps as in \Crefrange{eq:LFlimit_first}{eq:LFlimit_last}, i.e.\
\begin{align}
\lim_{T\to\infty}\widehat{\mat{\Psi}}_\gamma&=\lim_{T\to\infty}\bigg( \sum_{\tau=0}^{\tau_{\text{max}}}\gamma^\tau\widehat{\mat{\Omega}}_\tau\bigg)\\
&=\sum_{\tau=0}^{\tau_{\text{max}}}\gamma^\tau\bigg( \lim_{T\to\infty}\widehat{\mat{\Omega}}_\tau\bigg)\\
&=\sum_{\tau=0}^{\tau_{\text{max}}}\gamma^\tau \bigg(\mat{\Pi}(\mat{P}^\tau)_{\text{add}} \bigg)\\
&=\mat{\Pi}\bigg(\sum_{\tau=0}^{\tau_{\text{max}}}\gamma^\tau(\mat{P}^\tau)_{\text{add}}\bigg)\\
&=\mat{\Pi}\mat{M}_{\text{add}}
\end{align}

Following the same argument as in \Crefrange{eq:ACmatfirst}{eq:ACmatlast}, it is possible to show that the matrices $\widehat{\dot{\mat{\Sigma}}}$, $\widehat{\mat{\Sigma}}$, and $\widehat{\mat{\Omega}}_1$ are related in the same way as $\dot{\mat{\Sigma}}$, $\mat{\Sigma}$, and $\mat{\Omega}_1$, i.e.\
\begin{equation}
\widehat{\dot{\mat{\Sigma}}}=2(\widehat{\mat{\Sigma}}-\widehat{\mat{\Omega}}_1)\label{eq:TimeDerivACnocent}
\end{equation}
\cref{eq:TimeDerivACnocent} evaluates in the limit $T\to\infty$ to:
\begin{align}
\lim_{T\to\infty}\widehat{\dot{\mat{\Sigma}}}&=\lim_{T\to\infty}2(\widehat{\mat{\Sigma}}-\widehat{\mat{\Omega}}_1)\\
\overset{(\ref{eq:STClimit_nocent},\ref{eq:Lim2ndMom})}&{=}2\mat{\Pi}-2\mat{\Pi}\mat{P}_{\textup{add}}\\
\overset{(\ref{eq:LdirDef})}&{=}2\mat{L}_{\text{dir}}
\end{align}
\end{proof}

\end{appendices}

\begin{refcontext}[sorting=nyt]
\printbibliography
\end{refcontext}


\end{document}